\documentclass[twoside,11pt]{article}

\usepackage{blindtext}

\usepackage[preprint]{jmlr2e}

\usepackage{xcolor} 
\usepackage{bbm} 
\usepackage{algorithm} 
\usepackage[noend]{algpseudocode} 
\usepackage{amsmath} 
\usepackage{subcaption}
\usepackage{tikz} 
\usetikzlibrary{shapes.geometric}
\usepackage{amsfonts}
\usepackage{amssymb}
\usepackage{bm}
\usepackage[noend]{algpseudocode}
\usepackage{listings}
\usepackage{lipsum}
\usepackage{wrapfig}

\DeclareMathAlphabet{\mathmybb}{U}{bbold}{m}{n}
\newcommand{\indicator}{\mathmybb{1}}

\usepackage{custom_macros}

\usepackage{lastpage}
\jmlrheading{XX}{2023}{1-\pageref{LastPage}}{03/23}{XX/XX}{XX-XXXX}{Julien Ferry, Gabriel Laberge and Ulrich Aïvodji}

\ShortHeadings{Learning Hybrid Interpretable Models: Theory, Taxonomy, and Methods}{Ferry, Laberge and Aïvodji}
\firstpageno{1}

\begin{document}

\title{Learning Hybrid Interpretable Models:\\ Theory, Taxonomy, and Methods}

\author{
    \name $\!\!$Julien Ferry \email jferry@laas.fr \\
    \addr Operations Research, Combinatorial Optimization and Constraints\\
    LAAS-CNRS, Universit\'{e} de Toulouse, CNRS\\
    Toulouse, France
    \AND
    \name Gabriel Laberge \email gabriel.laberge@polymtl.ca \\
    \addr G\'enie Informatique et G\'enie Logiciel\\
    Polytechnique Montr\'eal\\
    Montr\'eal, Canada
    \AND
    \name Ulrich Aïvodji \email ulrich.aivodji@etsmtl.ca \\
    \addr Software and Information Technology Engineering\\
    \'Ecole de Technologie Supérieure\\
    Montr\'eal, Canada}

\editor{} 

\maketitle

\begin{abstract}
A hybrid model involves the cooperation of an interpretable model and a complex black box. At inference, 
any input of the hybrid model is assigned to either its interpretable or complex component based on a 
gating mechanism. The advantages of such models over classical ones are two-fold: 1) They grant users precise 
control over the level of transparency of the system and 2) They can potentially perform better than a standalone 
black box since redirecting some of the inputs to an interpretable model implicitly acts as regularization. Still, 
despite their high potential, hybrid models remain under-studied in the interpretability/explainability literature.
In this paper, we remedy this fact by presenting a thorough investigation of such models from three perspectives: 
Theory, Taxonomy, and Methods. First, we explore the
theory behind the generalization of hybrid models from the Probably-Approximately-Correct (PAC) perspective. 
A consequence of our PAC guarantee is the existence of a \textit{sweet spot} for the optimal transparency of the system. 
When such a sweet spot is attained, a hybrid model can potentially perform better than a standalone black box. 
Secondly, we provide a general taxonomy for the different ways of training hybrid models: the \postbb$\,$ and 
\prebb$\,$ paradigms. These approaches differ in the order in which the interpretable and complex components are trained. 
We show where the state-of-the-art hybrid models Hybrid-Rule-Set and Companion-Rule-List fall in this taxonomy. 
Thirdly, we implement the two paradigms in a single method: \HybridCORELS{}, which extends the CORELS algorithm 
to hybrid modeling. By leveraging CORELS, HybridCORELS provides a certificate of optimality of its interpretable 
component and precise control over transparency. We finally show empirically that HybridCORELS is competitive 
with existing hybrid models, and performs just as well as a standalone black box (or even better) while being partly transparent.
\end{abstract}



\section{Introduction}

The ever-increasing integration of machine learning models in high-stakes decision-making contexts (e.g., healthcare, justice, finance) has fostered a 
growing demand for transparency in recent years. Current workhorses to address transparency concerns in machine learning include black-box explanation 
and transparent design techniques~\citep{guidotti2018survey}. Black-box explanation techniques aim at explaining complex machine learning models in a 
post-hoc fashion with global explanations such as \texttt{Trepan}~\citep{craven1995extracting} and \texttt{BETA}~\citep{lakkaraju2017interpretable} or local explanations such as \texttt{LIME}~\citep{ribeiro2016should} and \texttt{SHAP}~\citep{lundberg2017unified}. On the other hand, transparent design concerns 
the development of inherently interpretable models such as rule lists~\citep{DBLP:journals/ml/Rivest87,DBLP:journals/jmlr/AngelinoLASR17}, rule 
sets~\citep{rijnbeek2010finding}, decision trees~\citep{breiman2017classification}, and scoring systems~\citep{ustun2016supersparse}. 

However, both black-box explanations and transparent design face performance and trustworthiness challenges that can prevent their wide adoption. 
On the one hand, while inherently interpretable models can be more easily understood and adopted by non-domain experts, their out-of-the-box performance 
can be worst than non-transparent models. Moreover, training such models to optimality is often NP-hard due to their discrete nature. On the other hand, 
black boxes can effortlessly attain high performance but their decision mechanisms are opaque and hard to understand by both experts and non-experts. 
Also, post-hoc explanations of these complex models have been shown to be unreliable and highly manipulable by ill-intentioned entities \citep{aivodji2019fairwashing,slack2020fooling,dimanov2020you,laberge2022fooling,aivodji2021characterizing}. This conundrum between black-box or transparent 
designs is colloquially referred to as the ``accuracy-transparency trade-off'', that is, one has to choose between transparent models with lower 
performance or opaque models that perform well but whose explanations are not trustworthy. Still, this trade-off is not a quantitative measure but rather 
a part of the collective imagination of researchers in interpretable machine learning. For this reason, the accuracy-transparency trade-off has been heavily
criticized and even labeled a myth \citep{rudin2019stop}. But the question remains, does such a trade-off exist? And if it does, is there a way to 
quantitatively measure it? Or even optimize it? 

\begin{figure}[t!]
    \centering
    \begin{subfigure}[b]{0.49\textwidth}
        \centering
        \raisebox{13pt}{
            \resizebox{\textwidth}{!}{         
                \begin{tikzpicture}
    \tikzset{>=stealth};
    
    \tikzstyle{gate} = [diamond, fill=black,draw=black, 
    minimum height=1.5cm, minimum width=1cm, align=center]
    
    \tikzstyle{block} = [rectangle, fill=white!95!black,draw=black, 
    minimum height=1.5cm, minimum width=3cm, align=center, rounded corners=10]
    
    \node[gate] (gate) at (0, 0) {\textcolor{white}{Gate}};
    \node[right of=gate, xshift=0.25cm] at (gate.east) {Send to $h_s$?};
    \node[block] (simple) at (-3, -3) {Simple Model \\$h_s(\bm{x})$};
    \node[block] (complex) at (3, -3) {Opaque Model \\$h_c(\bm{x})$};
    
    
    \tikzstyle{arrow} = [->, line width=0.4mm]
    
    \draw[arrow] (0, 2) node[anchor=south] {Query $\bm{x}$} -- (gate.north);
    \draw[arrow] (gate.south) |- (-1.5, -1.5) node[anchor=south] {Yes} -| (simple.north);
    \draw[arrow] (gate.south) |- (1.5, -1.5) node[anchor=south] {No} -| (complex.north);
    
\end{tikzpicture}
            }
        }
        \caption{}
    \end{subfigure}
    \hfill
    \begin{subfigure}[b]{0.49\textwidth}
        \centering
        \includegraphics[width=\textwidth]
        {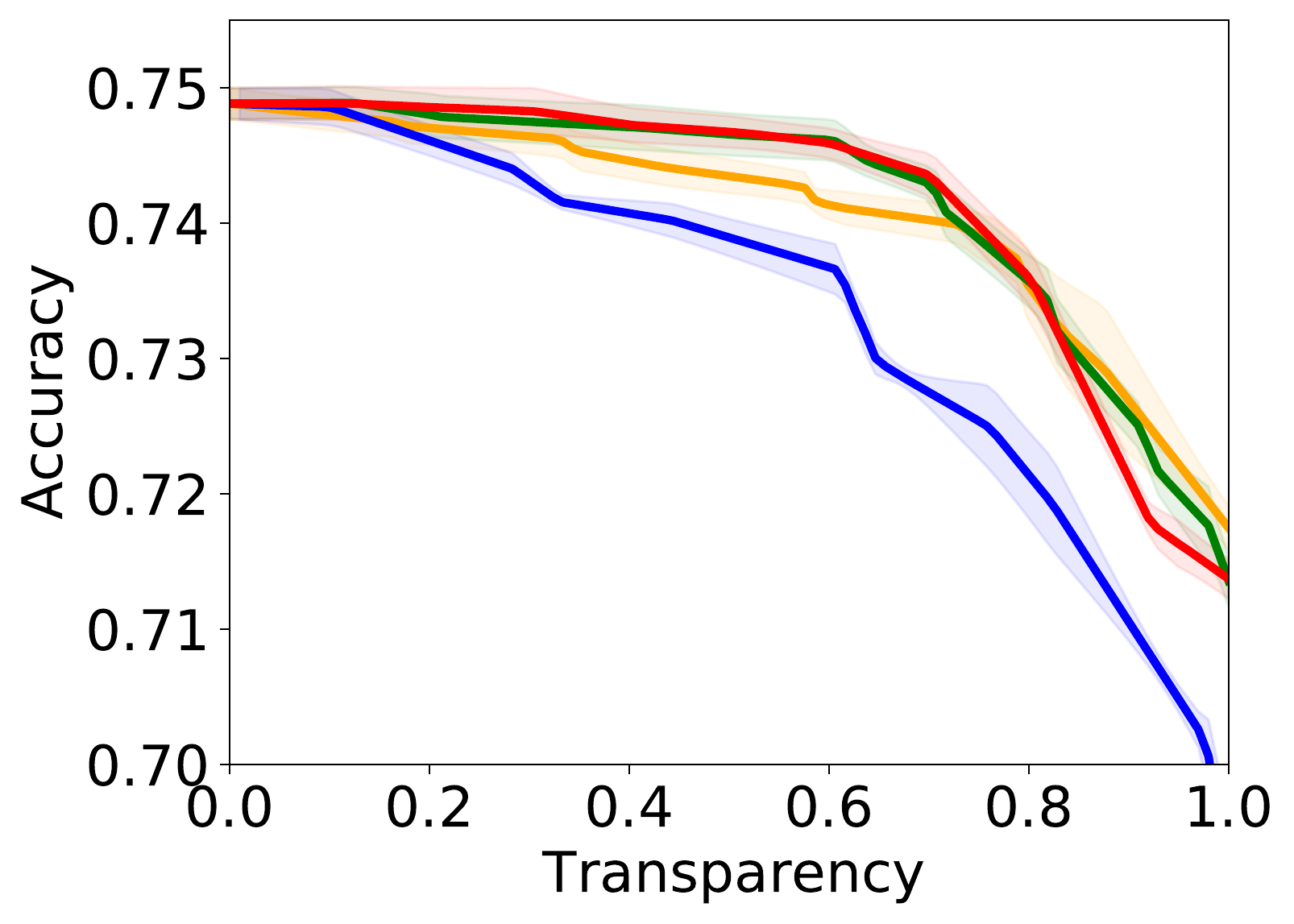}
        \caption{}
    \end{subfigure}

    \vskip 10 pt
    \includegraphics[width=0.9\textwidth]
   {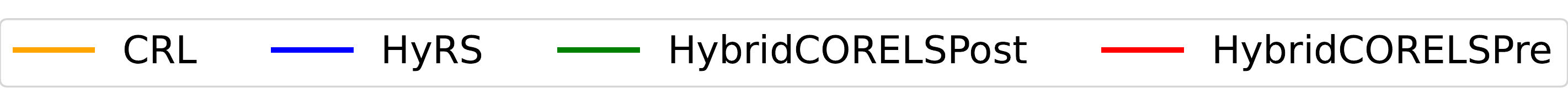}
   
    \caption{Overview of Hybrid Interpretable Modeling. (a) General schematic of a Hybrid Model where, at inference time, a gating mechanism 
    determines whether to send the instance to the interpretable component $h_s$ or to the complex one $h_c$. (b) Letting transparency be the 
    ratio of samples sent to the interpretable component $h_s$, the trade-off between accuracy and transparency can be measured and compared 
    across different Hybrid Models.}
    \label{fig:hybrid_models_intro}
\end{figure}
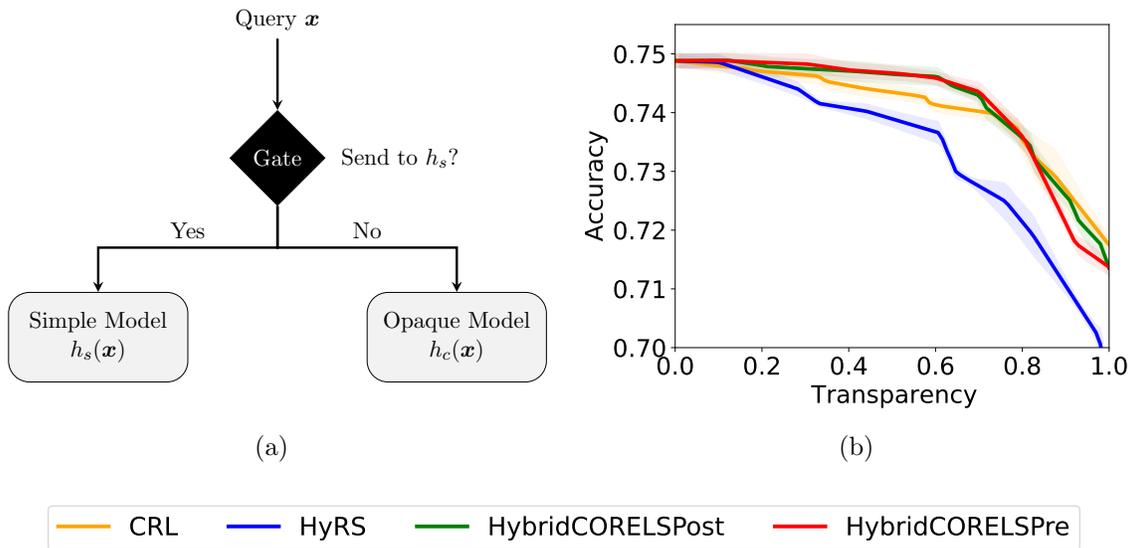

To explore such questions, we will not treat black-box and transparent designs as dichotomies. Rather, we will embrace both and explore
the continuum between the two philosophies. More specifically, we will study Hybrid Interpretable Models \citep{wang2019gaining,pan2020interpretable,wang2021hybrid}, which are systems that involve the cooperation of an interpretable model and a complex 
black box. At inference time, any input of the hybrid model is assigned to either its interpretable or complex component based on a 
gating mechanism, see Figure \ref{fig:hybrid_models_intro}~(a). The intuition behind this type of modeling is that 
not all examples in a dataset are hard to classify. 

We define the system's transparency as the ratio of samples that are sent to the interpretable
part. The higher the transparency, the more model predictions one can actually understand and possibly certify. 
However, it is possible that the interpretable component makes more errors on average meaning that the overall system
suffers a performance loss. Therefore, an integral part of hybrid modeling is to empirically explore the accuracy-transparency 
trade-off and find the best compromises, see Figure \ref{fig:hybrid_models_intro}~(b). We note that the accuracy-transparency trade-off
is no longer a myth, but actually something we measure and optimize. This is why we believe Hybrid Models are a very interesting 
research direction in the quest for interpretable machine learning.

Still, despite their high potential, hybrid models remain under-studied and under-used in the interpretability/explainability literature.
One of the reasons for this under-exploration could be that
learning interpretable models is very hard (often NP-Hard), and fitting a Hybrid Model on top can only be harder.
To address this issue, past studies have optimized such models using local search heuristics \citep{wang2019gaining,pan2020interpretable}.
Nevertheless, we show in this study that the inherent stochasticity of these local search algorithms hinders the ability of practitioners to consistently
attain a target level of transparency. Simply put, hybrid models are currently not user-friendly enough to promote widespread study and application.

Given the recent development of highly efficient libraries for training interpretable models to optimality (\textit{e.g.}, CORELS for rule-lists \citep{DBLP:journals/jmlr/AngelinoLASR17}, GOSDT for decision trees \citep{hu2019optimal})), we 
believe it is now possible to practically train Hybrid Models to optimality, even when adding a hard constraint on transparency level.

To sensitize the community to the immediate potential of hybrid models and to encourage additional research, we offer a fundamental 
investigation of such models from three perspectives: Theory, Taxonomy, and Methods. 
From the theory point of view, we explore Probably-Approximately-Correct (PAC) generalization guarantees of hybrid models. 
A consequence of our PAC guarantee is the existence of a \textit{sweet spot} for the optimal transparency of the system. 
When such a sweet spot is attained, a hybrid model can potentially perform better than a standalone black box. 
Secondly, we provide a general taxonomy for the different ways of training hybrid models: the \postbb$\,$ and \prebb$\,$ paradigms. 
These approaches differ in the order in which the interpretable and complex components are trained. 
We show where state-of-the-art hybrid models fall in this taxonomy. Thirdly, we implement the two paradigms in a single method: 
\HybridCORELS{}, which extends the library \corels{} to hybrid modeling. By leveraging \corels{}, \HybridCORELS{} provides 
a certificate of optimality of its interpretable component and precise control over transparency.
We finally show empirically that \HybridCORELS{} is competitive with existing hybrid models, 
and performs just as well as a standalone black box (or even better) while being partly transparent.
To resume, our contributions are as follows:

\begin{itemize}
    \item We theoretically study hybrid models under the PAC-Learning framework and derive generalization bounds. We show that such bounds depend on the amount of data classified by each part of the hybrid model and that an optimal transparency value exists.
    \item We introduce a taxonomy of hybrid models' learning methods. This taxonomy identifies two main families: the \prebb{} paradigm and the \postbb{} paradigm. We instantiate the proposed \prebb{} paradigm with a generic framework, using a key notion of \emph{black-box specialization via re-weighting}.
    \item We review state-of-the-art methods for learning hybrid models, and show that they all fall into the \postbb{} category.
    \item We modify a state-of-the-art algorithm for learning optimal sparse rule lists, named CORELS. More precisely, we propose a method for learning hybrid models with the \postbb{} paradigm. Our method, called \HybridCORELSPost{}\footnotemark is the first to provide optimality guarantees and explicit control of the model \emph{transparency}.
    \item We propose another modified version of CORELS for learning hybrid models with the \prebb{} paradigm. This method, named \HybridCORELSPre{}\footnotemark[\value{footnote}], is the first one using the proposed framework for the \prebb{} paradigm. Again, it provides optimality guarantees and explicit control of the model \emph{transparency}.
    \item We empirically show, using the proposed \HybridCORELSPre{} algorithm, that the \prebb{} paradigm is suitable for learning accurate hybrid models with transparency constraints.
    \item We empirically compare both \HybridCORELSPre{} and \HybridCORELSPost{} with state-of-the-art methods for learning hybrid models. We show that 
    both methods offer competitive trade-offs between \emph{accuracy} and \emph{transparency}, while also providing facilitated control over the 
    latter, and optimality guarantees.
\end{itemize}

\footnotetext{Our proposed methods are implemented within a publicly available and user-friendly Python module, named \HybridCORELS{}.}

\section{Hybrid Interpretable Models: a Theoretical Analysis}
\label{sec:theory_hybrid_models}
In this section, we formally introduce hybrid interpretable models and analyze them under 
the PAC-Learning framework. We derive generalization bounds and show that an optimal trade-off  
between accuracy and transparency (the proportion of data classified by the interpretable component) exists, leveraging the advantages of both parts of the model.

\subsection{Definitions}
Let $\X$ be the input space and let $\Hc,\Hs$ be two sets of binary classifiers $h:\X\rightarrow \{0,1\}$.
We shall impose that $|\Hs| \ll |\Hc| < \infty$ so that $\Hs$ represents a simple set of
models while $\Hc$ represents a complex set of models. Finally,
we let $\mathcal{P}$ be a set of subsets of $\X$ (for instance, $\mathcal{P}$ may be the power set 
of $\X$, or the set of linear half-spaces). The intuition behind hybrid modeling is that there may exist
a region $\Omega \in \mathcal{P}$ where a complex model $h_c\in \Hc$ is overkill and hence it could be replaced
by a simpler model $h_s\in \Hs$ on that region without significant loss in terms of classification performance.
Formally, a hybrid model is a triplet $\triplet \in \Hyb:= \tripletspace$ which instantiates a function of the form 
\begin{equation*}
    \forall \bm{x} \in \X, \quad \triplet (\bm{x}) = 
    \begin{cases}
       h_s(\bm{x}) &\quad\text{if } \bm{x}\in \Omega,\\
       h_c(\bm{x}) &\quad\text{otherwise.} \\
     \end{cases}
\end{equation*}

\begin{figure}[t!]
    \centering
    \begin{subfigure}[b]{0.49\textwidth}
        \centering
        \resizebox{4cm}{!}{
            \begin{tikzpicture}

    \fill[blue!35] (0,0) rectangle ++ (1,1);
    \fill[blue!35] (0,2) rectangle ++ (1,1);
    \fill[blue!35] (0,4) rectangle ++ (1,1);
    \fill[blue!35] (1,5) rectangle ++ (1,1);
    \fill[blue!35] (3,5) rectangle ++ (1,1);
    \fill[blue!35] (5,5) node (v1) {} rectangle ++ (1,1);
    \fill[blue!35] (5,3) rectangle ++ (1,1);
    \fill[blue!35] (5,1) rectangle ++ (1,1);
    \fill[blue!35] (4,0) rectangle ++ (1,1);
    \fill[blue!35] (2,0) rectangle ++ (1,1);
    
    \fill[red!35] (0,1) rectangle ++ (1,1);
    \fill[red!35] (0,3) rectangle ++ (1,1);
    \fill[red!35] (0,5) rectangle ++ (1,1);
    \fill[red!35] (2,5) rectangle ++ (1,1);
    \fill[red!35] (4,5) rectangle ++ (1,1);
    \fill[red!35] (5,4) rectangle ++ (1,1);
    \fill[red!35] (5,2) rectangle ++ (1,1);
    \fill[red!35] (5,0) rectangle ++ (1,1);
    \fill[red!35] (3,0) rectangle ++ (1,1);
    \fill[red!35] (1,0) rectangle ++ (1,1);
    
    \fill[blue!35] (1,1) rectangle ++ (2,2);
    \fill[blue!35] (3,3) rectangle ++ (2,2);
    \fill[red!35] (1,3) rectangle ++ (2,2);
    \fill[red!35] (3,1) rectangle ++ (2,2);

    \draw[step=1.0,black,thin] (0,0) grid (6,6);
    \draw[line width=3]  (1,5) rectangle (5,1);
\end{tikzpicture}
        }
        \caption{Example of a region $\Omega$ (shown as a thick square) where a
        complex model $h_c\in \Hc$ (with $|\Hc|=2^{36}$) is overly complex.}
    \end{subfigure}
    \hfill
    \begin{subfigure}[b]{0.49\textwidth}
        \centering
        \resizebox{4cm}{!}{
            \begin{tikzpicture}

    \fill[blue!35] (0,0) rectangle ++ (1,1);
    \fill[blue!35] (0,2) rectangle ++ (1,1);
    \fill[blue!35] (0,4) rectangle ++ (1,1);
    \fill[blue!35] (1,5) rectangle ++ (1,1);
    \fill[blue!35] (3,5) rectangle ++ (1,1);
    \fill[blue!35] (5,5) node (v1) {} rectangle ++ (1,1);
    \fill[blue!35] (5,3) rectangle ++ (1,1);
    \fill[blue!35] (5,1) rectangle ++ (1,1);
    \fill[blue!35] (4,0) rectangle ++ (1,1);
    \fill[blue!35] (2,0) rectangle ++ (1,1);
    
    \fill[red!35] (0,1) rectangle ++ (1,1);
    \fill[red!35] (0,3) rectangle ++ (1,1);
    \fill[red!35] (0,5) rectangle ++ (1,1);
    \fill[red!35] (2,5) rectangle ++ (1,1);
    \fill[red!35] (4,5) rectangle ++ (1,1);
    \fill[red!35] (5,4) rectangle ++ (1,1);
    \fill[red!35] (5,2) rectangle ++ (1,1);
    \fill[red!35] (5,0) rectangle ++ (1,1);
    \fill[red!35] (3,0) rectangle ++ (1,1);
    \fill[red!35] (1,0) rectangle ++ (1,1);
    
    \draw[step=1.0,black,thin] (0,0) grid (6,6);
    
    \fill[blue!35] (1,1) rectangle ++ (2,2);
    \fill[blue!35] (3,3) rectangle ++ (2,2);
    \fill[red!35] (1,3) rectangle ++ (2,2);
    \fill[red!35] (3,1) rectangle ++ (2,2);
    
    \draw[step=3,black,thin] (1,1) grid (5,5);
    
    \draw[line width=3]  (1,5) rectangle (5,1);
\end{tikzpicture}
        }
        \caption{The complex model $h_c$ can
        be replaced by a simpler one $h_s\in \Hs$
        (with $|H_s|=2^4$). Overall, this hybrid model space has size $|\Hyb|=2^{24}$.}
    \end{subfigure}
    \caption{Toy example with $\X=[0,1]\times[0,1]$. Here the complex models $\Hc$ are all the ways 
    to color the 36 width-1 squares. The simpler models $\Hs$ are all the ways to color the 
    4 width-2 squares in the middle.}
    \label{fig:toy_example_grid}
\end{figure}
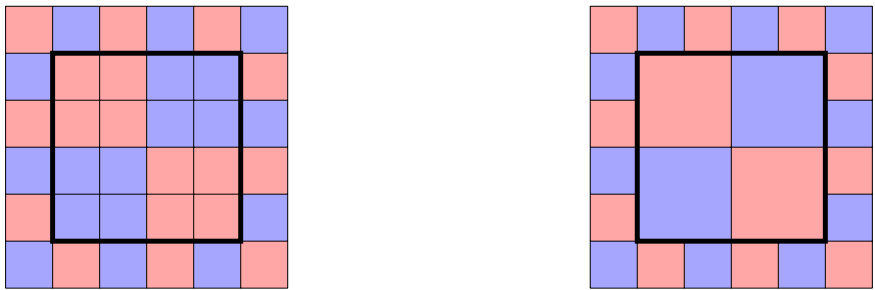

Figure \ref{fig:toy_example_grid} presents an informal argument favoring this modeling choice. We will 
additionally assume that the smaller hypothesis space $\Hs$ involves models that are \emph{interpretable by design} such as
rule lists, sparse decision trees, scoring systems, etc. This assumption will not affect the theoretical analysis,
which will just rely on $|\Hs|$ being small, but it will specify the desiderata of the hybrid model. Indeed, if
$h_s$ is interpretable, then we would like the region $\Omega$ on which it operates 
to be as big as possible without hindering performance. Letting $\D$ be a distribution over $\X\times \{0,1\}$
that represents a specific binary classification task, we want the \emph{transparency}
$C_\Omega:= \Prob_{\bm{x}\sim \mathcal{D}}[\bm{x}\in \Omega]$ to be as large as possible.

The rest of this section is structured as follows:
in \textbf{Section~\ref{subsec:pac}} we prove that finite hybrid models (\emph{i.e.,} $|\Hyb|\leq \infty$) are PAC-Learnable. 
That is if we learn a hybrid model on a finite dataset with sufficiently many examples,
then we can guarantee that the model will generalize to new unseen samples. This is an important first step
in the fundamental understanding of hybrid models. Afterward, in 
\textbf{Section~\ref{subsec:theory_sweet_spot}}, we study the impact of transparency on the tightness of the bound and
show that a ``sweet spot'' for transparency exists.

\subsection{Finite Hybrid Models are PAC-Learnable}\label{subsec:pac}
We are going to study distributions $\D$ where a 
perfect model $\tripletstar \in \Hyb$ exists:
\begin{equation}
    \poploss(\tripletstar) := 
    \Prob_{(\bm{x},y)\sim \D}[\tripletstar(\bm{x})\neq y]=0.
    \label{eq:optimal_model}
\end{equation}
Intuitively, the predictions of the optimal hybrid interpretable model $\tripletstar$ match the true 
label $y$ of every example $(\bm{x},y) \in (\X \times \{0,1\})$ drawn from distribution $\D$.
To learn such a model, we can employ the Empirical Risk Minimization (ERM) principle, which consists 
of sampling a dataset of $M$ iid examples $S:=\{(\bm{x}^{(i)}, y^{(i)})\}_{i=1}^M\sim \mathcal{D}^M$,
defining the empirical risk
\begin{equation}
    \emploss{\S}(\triplet) :=
    \sum_{i=1}^M \indicator{}[\triplet(\bm{x}^{(i)}) \neq y^{(i)}],
    \label{eq:emp_loss}
\end{equation}
and minimizing it across $\Hyb$
\begin{equation*}
    \triplet_\S := \ERM =
    \argmin_{\triplet \in \Hyb} \emploss{\S}(\triplet).
\end{equation*}
One can notice that we do not scale the empirical risk by a factor $\frac{1}{M}$ as multiplication 
by a constant factor does not affect ERM. The following theoretical result characterizes the generalization
of hybrid models learned with ERM.
\begin{theorem}
\label{theorem:generalization_bound}
Given a finite hybrid model space $(|\Hyb|< \infty$) and some
$\epsilon > 0$,
letting $C_\Omega:= 
\Prob_{\bm{x}\sim \mathcal{D}}[\bm{x}\in \Omega]$ be the transparency of $\Omega$, 
then for any distribution $\D$ where there exists a triplet $\tripletstar$
with zero generalization error (as defined in~(\ref{eq:optimal_model})), 
the following holds for a training set of size $M$:
 \begin{equation*}
      \failure \leq \sum_{\Omega\in \mathcal{P}}
        \mathcal{B}(\epsilon, C_\Omega, \Hc, \Hs, M),
 \end{equation*}
 where 
 \begin{equation*}
      \mathcal{B}(\epsilon, C_\Omega, \Hc, \Hs, M) := (1-|\Hc|)C_\Omega^M + (1-|\Hs|)C_{\overline{\Omega}}^M +
    |\Hc|(C_{\overline{\Omega}}e^{-\epsilon} + C_{\Omega})^M +
    |\Hs|(C_{\Omega}e^{-\epsilon} + C_{\overline{\Omega}})^M.
 \end{equation*}
 If we assume that the optimal subset $\Omega \equiv \Omega^\star$ is known in advance, then the bound tightens
\begin{equation}
      \failure \leq \mathcal{B}(\epsilon, C_\Omega, \Hc, \Hs, M).
    \label{eq:oracle_bound}
 \end{equation}

\begin{proof}
    The complete proof is provided in Appendix~\ref{app:proofs}.
\end{proof}
\end{theorem}

This generalization bound involves several key quantities: the amount of data $M$, the transparency $C_\Omega$ and 
its complement $C_{\overline{\Omega}}$ as well as the complexities of the hypothesis spaces 
$|\Hs|$ and $|\Hc|$. We will see in the coming subsection how these various parameters impact the tightness of the bound.

We note some of the limitations of these theoretical bounds. First, taking $C_{\Omega}= 0$, we obtain a trivial bound 
$1+|\mathcal{H}_c|e^{-\epsilon M}$. The same thing occurs when setting $C_{\Omega}=1$. Basically, the bound is
trivial unless input samples are shared between the complex and simple models. Secondly, the bound requires the
knowledge of transparency $C_\Omega:= \mathbb{P}_{\bm{x}\sim \mathcal{D}}[\bm{x}\in \Omega]$
which cannot be computed exactly in practice since the data-generating distribution $\mathcal{D}$ is unknown. The only way 
to practically estimate this quantity is to count how many data instances land in the region $\Omega$. Thirdly, the bound is
loose as its computation relies on applying the union bound repeatedly over $\mathcal{P}$, $\Hc$, and $\Hs$. Still, for 
$C_{\Omega}\in ]0,1[$, and any $\epsilon\in ]0, 1]$ the bound decreases as $M$ increases which implies that learning 
hybrid models is possible in theory.

\subsection{Fine-Tuning the Transparency}\label{subsec:theory_sweet_spot}
\begin{figure}[t!]
    \centering
    \begin{subfigure}[b]{0.24\textwidth}
        \centering
        \resizebox{2.5cm}{!}{
            \begin{tikzpicture}

    \fill[blue!35] (0,0) rectangle ++ (1,1);
    \fill[blue!35] (0,2) rectangle ++ (1,1);
    \fill[blue!35] (0,4) rectangle ++ (1,1);
    \fill[blue!35] (1,5) rectangle ++ (1,1);
    \fill[blue!35] (3,5) rectangle ++ (1,1);
    \fill[blue!35] (5,5) node (v1) {} rectangle ++ (1,1);
    \fill[blue!35] (5,3) rectangle ++ (1,1);
    \fill[blue!35] (5,1) rectangle ++ (1,1);
    \fill[blue!35] (4,0) rectangle ++ (1,1);
    \fill[blue!35] (2,0) rectangle ++ (1,1);
    
    \fill[red!35] (0,1) rectangle ++ (1,1);
    \fill[red!35] (0,3) rectangle ++ (1,1);
    \fill[red!35] (0,5) rectangle ++ (1,1);
    \fill[red!35] (2,5) rectangle ++ (1,1);
    \fill[red!35] (4,5) rectangle ++ (1,1);
    \fill[red!35] (5,4) rectangle ++ (1,1);
    \fill[red!35] (5,2) rectangle ++ (1,1);
    \fill[red!35] (5,0) rectangle ++ (1,1);
    \fill[red!35] (3,0) rectangle ++ (1,1);
    \fill[red!35] (1,0) rectangle ++ (1,1);
    
    \draw[step=1.0,black,thin] (0,0) grid (6,6);
    
    \fill[blue!35] (1,1) rectangle ++ (2,2);
    \fill[blue!35] (3,3) rectangle ++ (2,2);
    \fill[red!35] (1,3) rectangle ++ (2,2);
    \fill[red!35] (3,1) rectangle ++ (2,2);
    
    \draw[step=3,black,thin] (1,1) grid (5,5);
    
    \draw[line width=3]  (1,5) rectangle (5,1);
\end{tikzpicture}
        }
        \caption{Region $\Omega_1$}
    \end{subfigure}
    \hfill
    \begin{subfigure}[b]{0.24\textwidth}
        \centering
        \resizebox{2.5cm}{!}{
            \begin{tikzpicture}

    \fill[blue!35] (0,0) rectangle ++ (1,1);
    \fill[blue!35] (0,2) rectangle ++ (1,1);
    \fill[blue!35] (0,4) rectangle ++ (1,1);
    \fill[blue!35] (1,5) rectangle ++ (1,1);
    \fill[blue!35] (3,5) rectangle ++ (1,1);
    \fill[blue!35] (5,5) rectangle ++ (1,1);
    \fill[blue!35] (5,3) rectangle ++ (1,1);
    \fill[blue!35] (5,1) rectangle ++ (1,1);
    \fill[blue!35] (4,0) rectangle ++ (1,1);
    \fill[blue!35] (2,0) rectangle ++ (1,1);
    
    \fill[red!35] (0,1) rectangle ++ (1,1);
    \fill[red!35] (0,3) rectangle ++ (1,1);
    \fill[red!35] (0,5) rectangle ++ (1,1);
    \fill[red!35] (2,5) rectangle ++ (1,1);
    \fill[red!35] (4,5) rectangle ++ (1,1);
    \fill[red!35] (5,4) rectangle ++ (1,1);
    \fill[red!35] (5,2) rectangle ++ (1,1);
    \fill[red!35] (5,0) rectangle ++ (1,1);
    \fill[red!35] (3,0) rectangle ++ (1,1);
    \fill[red!35] (1,0) rectangle ++ (1,1);
    
    \fill[red!35] (3,1) rectangle ++ (2,2);

    \draw[step=1.0,black,thin] (0,0) grid (6,6);
    \fill[blue!35] (3,3) node (v3) {} rectangle ++ (2,2);
    \fill[blue!35] (1,1) rectangle ++ (2,2);
    \fill[red!35] (1,3) rectangle ++ (2,2);
    
    \draw[step=3,black,thin] (1,1) grid (3,5);
    
	\draw[draw=black, line width=3] (1,1) -- (3,1) -- (3,3) -- (5,3) 
		-- (5,5) -- (1,5) -- (1,1);
\end{tikzpicture}
        }
        \caption{Region $\Omega_2$}
    \end{subfigure}
    \hfill
    \begin{subfigure}[b]{0.24\textwidth}
        \centering
        \resizebox{2.5cm}{!}{
            \begin{tikzpicture}

    \fill[blue!35] (0,0) rectangle ++ (1,1);
    \fill[blue!35] (0,2) rectangle ++ (1,1);
    \fill[blue!35] (0,4) rectangle ++ (1,1);
    \fill[blue!35] (1,5) rectangle ++ (1,1);
    \fill[blue!35] (3,5) rectangle ++ (1,1);
    \fill[blue!35] (5,5) node (v1) {} rectangle ++ (1,1);
    \fill[blue!35] (5,3) rectangle ++ (1,1);
    \fill[blue!35] (5,1) rectangle ++ (1,1);
    \fill[blue!35] (4,0) rectangle ++ (1,1);
    \fill[blue!35] (2,0) rectangle ++ (1,1);
    
    \fill[red!35] (0,1) rectangle ++ (1,1);
    \fill[red!35] (0,3) rectangle ++ (1,1);
    \fill[red!35] (0,5) rectangle ++ (1,1);
    \fill[red!35] (2,5) rectangle ++ (1,1);
    \fill[red!35] (4,5) rectangle ++ (1,1);
    \fill[red!35] (5,4) rectangle ++ (1,1);
    \fill[red!35] (5,2) rectangle ++ (1,1);
    \fill[red!35] (5,0) rectangle ++ (1,1);
    \fill[red!35] (3,0) rectangle ++ (1,1);
    \fill[red!35] (1,0) rectangle ++ (1,1);
    
    \fill[blue!35] (3,3) rectangle ++ (2,2);
    \fill[red!35] (3,1) rectangle ++ (2,2);
    \draw[step=1.0,black,thin] (0,0) grid (6,6);
    \fill[blue!35] (1,1) rectangle ++ (2,2);
    \fill[red!35] (1,3) rectangle ++ (2,2);
    
    \draw[step=3,black,thin] (1,1) grid (3,5);
    
    \draw[line width=3]  (1,5) rectangle (3,1);
\end{tikzpicture}
        }
        \caption{Region $\Omega_3$}
    \end{subfigure}
    \hfill
    \begin{subfigure}[b]{0.24\textwidth}
        \centering
        \resizebox{2.5cm}{!}{
            \begin{tikzpicture}

    \fill[blue!35] (0,0) rectangle ++ (1,1);
    \fill[blue!35] (0,2) rectangle ++ (1,1);
    \fill[blue!35] (0,4) rectangle ++ (1,1);
    \fill[blue!35] (1,5) rectangle ++ (1,1);
    \fill[blue!35] (3,5) rectangle ++ (1,1);
    \fill[blue!35] (5,5) node (v1) {} rectangle ++ (1,1);
    \fill[blue!35] (5,3) rectangle ++ (1,1);
    \fill[blue!35] (5,1) rectangle ++ (1,1);
    \fill[blue!35] (4,0) rectangle ++ (1,1);
    \fill[blue!35] (2,0) rectangle ++ (1,1);
    
    \fill[red!35] (0,1) rectangle ++ (1,1);
    \fill[red!35] (0,3) rectangle ++ (1,1);
    \fill[red!35] (0,5) rectangle ++ (1,1);
    \fill[red!35] (2,5) rectangle ++ (1,1);
    \fill[red!35] (4,5) rectangle ++ (1,1);
    \fill[red!35] (5,4) rectangle ++ (1,1);
    \fill[red!35] (5,2) rectangle ++ (1,1);
    \fill[red!35] (5,0) rectangle ++ (1,1);
    \fill[red!35] (3,0) rectangle ++ (1,1);
    \fill[red!35] (1,0) rectangle ++ (1,1);
    
    \fill[red!35] (1,3) rectangle ++ (2,2);
    \fill[blue!35] (3,3) rectangle ++ (2,2);
    \fill[red!35] (3,1) rectangle ++ (2,2);
    \draw[step=1.0,black,thin] (0,0) grid (6,6);
    \fill[blue!35] (1,1) rectangle ++ (2,2);
    
    \draw[step=3,black,thin] (1,1) grid (3,5);
    
    \draw[line width=3]  (1,1) rectangle (3,3);
\end{tikzpicture}
        }
        \caption{Region $\Omega_4$}
    \end{subfigure}
    \caption{Four hybrid models that are functionally
    equivalent but have different regions $\Omega$.}
    \label{fig:model_equivalence}
\end{figure}
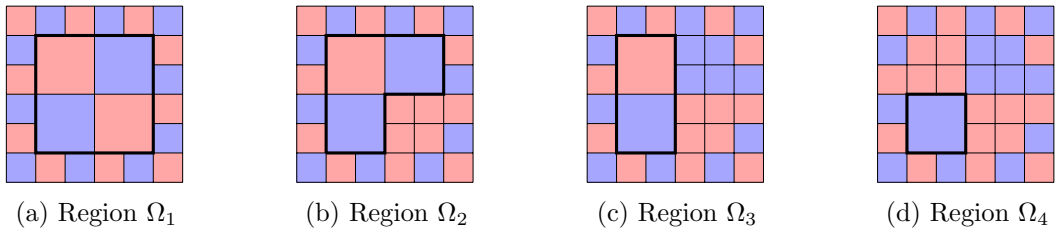

A particular property of hybrid models is that the optimal model $\tripletstar$ from Equation 
(\ref{eq:optimal_model}) need not be unique. Indeed, given the flexibility of choosing the region $\Omega$ on which 
the simple model is applied, we could have two hybrid models with the same functional output. Figure \ref{fig:model_equivalence}
presents a toy example of four hybrid models that are all functionally equivalent but with different regions $\Omega$. 

Now the hypothesis that the optimal region $\Omega^\star$ is known in advance
could be replaced with the knowledge of a set of optimal regions $\{\Omega^\star_i\}_{i=1}^R$. 
If such is the case, which region should be returned by the learning algorithm? 
Using the empirical error as a criterion would not work since any ERM fitted using these optimal 
regions would return an error of 0. We propose to leverage the theoretical bound to decide which region to 
employ. Specifically, if we fix some region $\Omega^\star_i$ for the ERM algorithm, then Equation 
(\ref{eq:oracle_bound}) provides a bound $\mathcal{B}(\epsilon, C_\Omega, \Hc, \Hs, M)$ on the probability of 
having an error that exceeds $\epsilon$ for any $\epsilon\in\,]0,1]$. Taking the area under the curve
\begin{equation*}
    \text{AUC}(C_{\Omega^\star_i}, \Hc, \Hs, M)
    := \int_0^1 \mathcal{B}(\epsilon, C_{\Omega^\star_i}, \Hc, \Hs, M) d\epsilon\quad\forall i=1,2,\ldots,R,
\end{equation*}
can be used as a measure of the tightness of the bound for all failure levels $\epsilon$. Hence, by studying
the $\text{AUC}(C_{\Omega^\star_i}, \Hc, \Hs, M)$ as a function of $\Omega^\star_i$, one can theoretically decide which 
region to use in the final model. 

In the following example, we have defined $\Hs$ as the set of all binary depth-3 decision trees (7 internal nodes 
and 8 leaves with binary outcomes) fitted on 200 binary features ($\X=\{0,1\}^{200}$). This hypothesis space has a size 
$|\Hs|=2^8\times 200\times 199^2\times198^4\approx 3.11\ee{18}$. We have defined $\Hc$ to be any hypothesis space 
that is larger than $\Hs$ by some factor $|\Hc|=N\times|\Hs|$.

\begin{figure}[t!]
    \centering
    \begin{subfigure}[b]{0.5\linewidth}
        \centering
        \includegraphics[width=\textwidth]
        {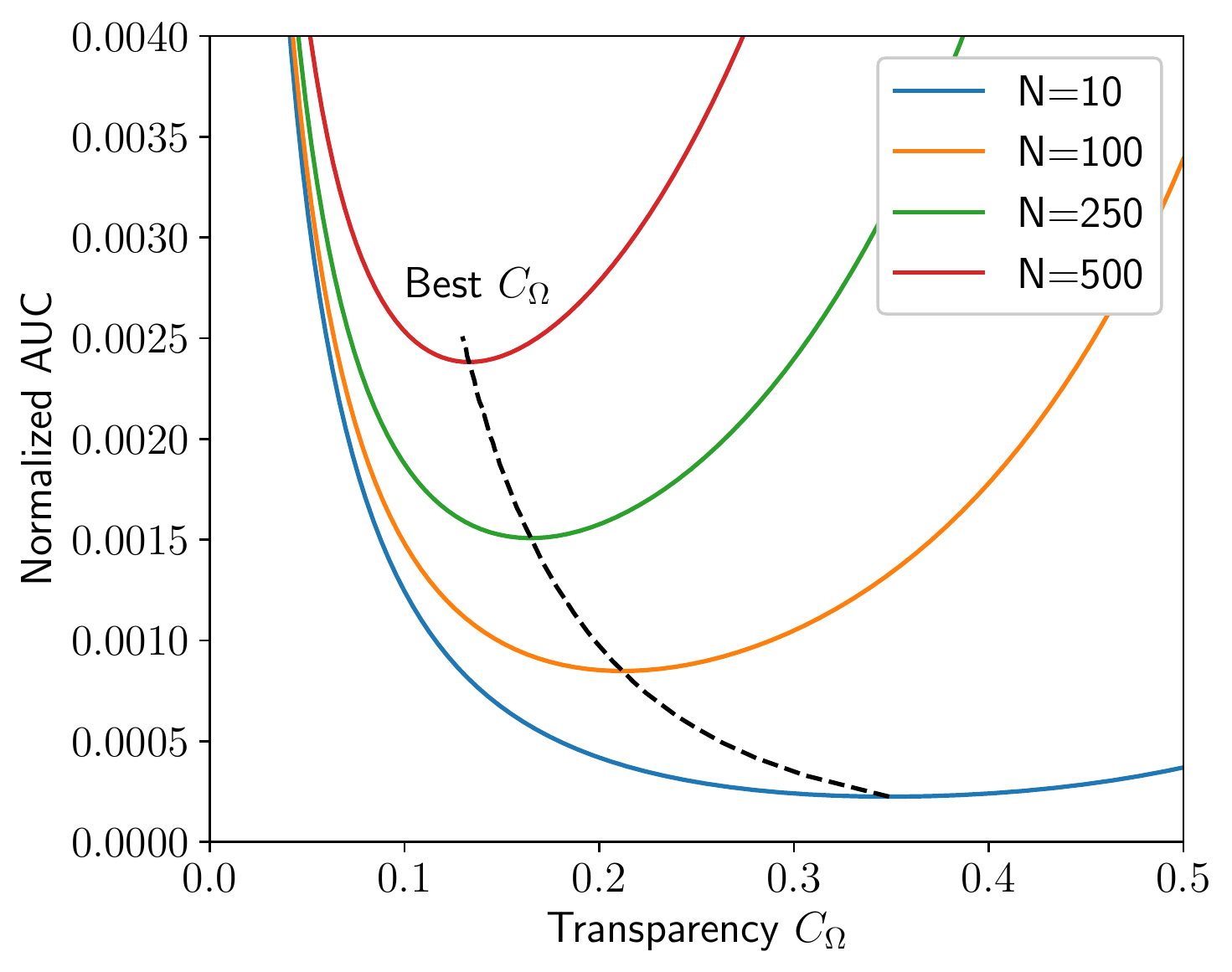}
        \caption{Varying $N$ with $M=5000$.}
    \end{subfigure}
    \hfill
    \begin{subfigure}[b]{0.484\linewidth}
        \centering
        \vspace{0.15cm}
        \includegraphics[width=\textwidth]
        {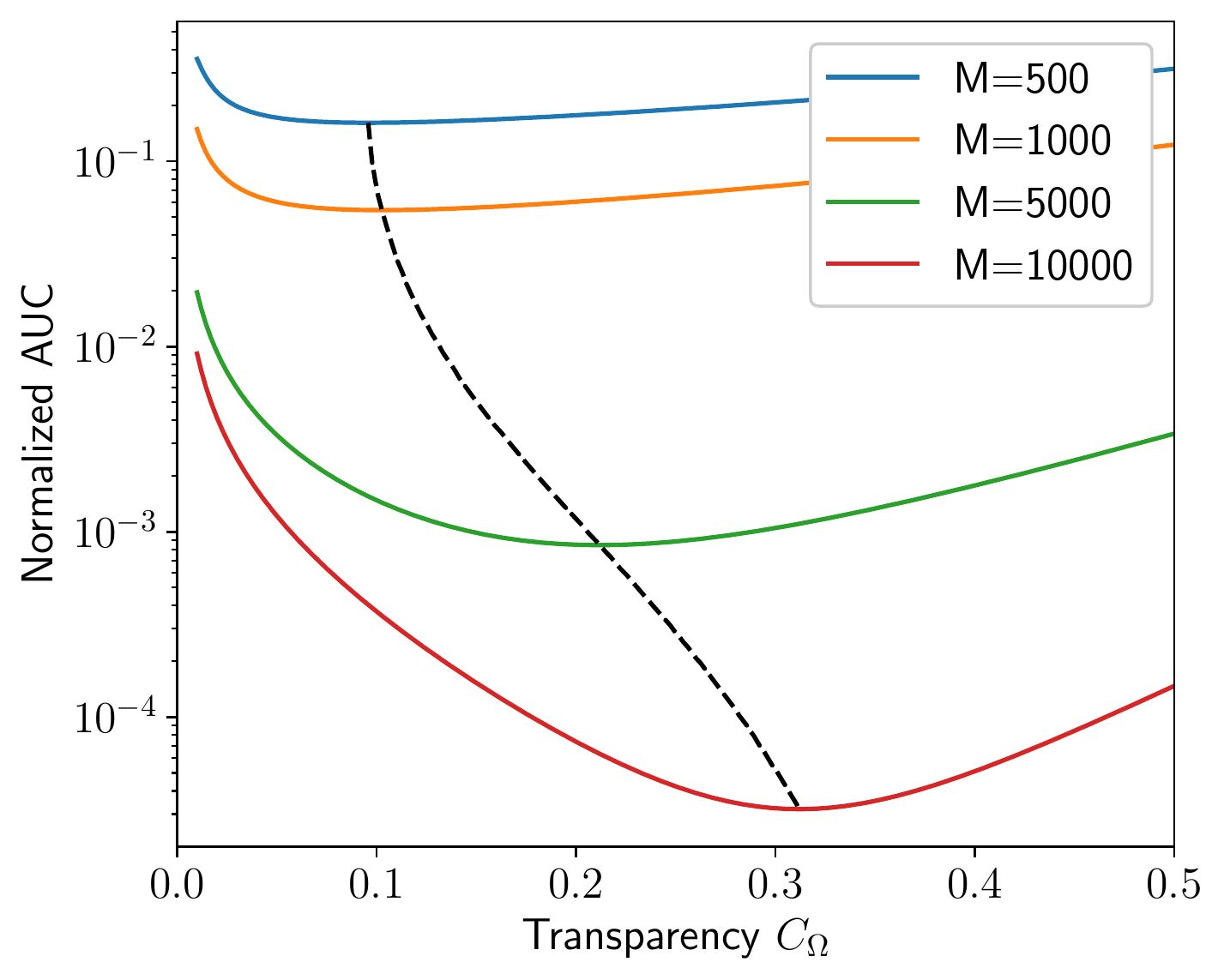}
        \caption{Varying $M$ with $N=100$.}
    \end{subfigure}
    \caption{Normalized AUC (\textit{i.e.}, AUC/$|\Hs|$) of the theoretical upper bound
    as a function of transparency $C_\Omega$. We observe a ``sweet spot'' with minimal AUC which 
    depends on $N$ (ratio of the hypothesis spaces' sizes $\frac{|\Hc|}{|\Hs|}$) and $M$ (size of the training dataset).}
    \label{fig:theo_bound}
\end{figure}

Figure \ref{fig:theo_bound} presents the AUC of the generalization bound as a function of the transparency
for this hypothetical example. We observe that, given $\Hc$, $\Hs$, and $M$, there is a ``sweet spot'' 
where the bound on error is the tightest
\begin{equation*}
    \Omega^{\star\star}(\Hc, \Hs, M)=\argmin_{i=1,2,\ldots,R}\text{AUC}(C_{\Omega^\star_i}, \Hc, \Hs, M).
\end{equation*}

Looking more specifically at Figure \ref{fig:theo_bound}~(a), increasing $N$ reduces the 
transparency that reaches the optimal AUC. This means that the more complex $\Hc$ is, the more 
input samples must be sent to train $h_c$ so it does not overfit.
Inspecting Figure \ref{fig:theo_bound}~(b), the transparency that attains minimal AUC increases as $M$ increases. 
This means that as we reach large values of $M$, we can afford to train the black box on a smaller ratio of the 
data without over-fitting.

We conclude this example by emphasizing that Figure \ref{fig:theo_bound} is mostly of theoretical interest so practitioners must take it 
with a grain of salt. More precisely, the exact values of the ``sweet spot'' for transparency are not indicative of the values one 
would obtain in real-life experiments. This is because our analysis is performed on a loose upper bound which we hope still captures 
the generalization dynamics of hybrid models. In real-life applications, the existence of an optimal transparency must be 
assessed experimentally. Still, the theory suggests the existence of such a ``sweet spot'', which in itself is an interesting result.

\subsection{Takeaways}
Although the bound makes strong assumptions that may not hold in practical applications, 
our theoretical analysis leads to fundamental 
insights into training hybrid models:
\begin{enumerate}
    \item Training hybrid interpretable models is theoretically possible given enough data.
    \item Important parameters that influence generalization are the complexities $|\Hs|$ and $|\Hc|$, 
    the transparency $C_\Omega$, and the number of data points $M$.
    \item There exists a ``sweet spot'' of the bound in terms of transparency which varies with 
    $\Hc$, $\Hs$, and $M$. Henceforth, in practical applications, we should sweep over possible values of 
    transparency. Some of the resulting hybrid models may attain better generalization.
\end{enumerate}

\section{Learning Hybrid Interpretable Models: Taxonomy and Methods}

We now introduce our proposed taxonomy of hybrid models learning frameworks.
We then show how rule-based classifiers can be used to implement hybrid interpretable models. 
Finally, we position state-of-the-art methods within the proposed taxonomy.

\subsection{Taxonomy of Hybrid Models Learning Frameworks}

A major challenge in training hybrid models is that two models must be trained instead of one. Given the proliferation of out-of-the-box 
implementations of complex model $h_c$, such as \texttt{Scikit-Learn} and \texttt{XGBoost} classifiers, it would be simpler to rely 
on them via their pre-existing \texttt{fit} and \texttt{predict} methods. Henceforth, we encourage hybrid model training procedures to 
be \textit{agnostic} to the type of black box $h_c$. This makes hybrid models a lot more versatile and user-friendly because 
any practitioner could just plug in their favorite black box implementation.

Given the technical constraint of black box agnosticism, we now leverage the previous PAC generalization bound to derive a learning objective. 
We have seen that the two important quantities to guarantee generalization are the complexity of the simple hypothesis space $\Hs$ and the transparency $C_{\Omega}\approx |S\cap \Omega|/|S|$. Since ``smaller is better'' in any learning objective, it should actually contain the complement of transparency 
$ C_{\overline{\Omega}}=(1-C_{\Omega})\approx |S\cap \overline{\Omega}|/|S|$. A general regularized learning objective would then be
\begin{equation}
    \obj(\triplet, S) = \frac{\emploss{\S}(\triplet)}{|\S|} + \lambda \cdot K_{\Hs} + \beta \cdot \frac{|\S\cap \overline{\Omega}|}{|\S|},
    \label{eq:learning_hybrid_general}
\end{equation}
where $K_{\Hs}$ is a complexity measure of $\Hs$ and $\lambda,\beta\geq0$ are regularization hyper-parameters that respectively control 
the cost of increasing the complexity of $\Hs$ (for instance considering depth), and that of increasing the black box part coverage $C_{\overline{\Omega}}$ (equivalently decreasing the interpretable part coverage $C_\Omega$, which constitutes the model's transparency).
Again, the proportion $C_\Omega$ of data classified by the simple part of the hybrid model is called \emph{transparency}. 
A hybrid model whose transparency is $0.0$ hence simply consists of a black box, while one with a $1.0$ 
transparency is an entirely interpretable model. Hybrid models usually make some trade-offs between transparency and predictive accuracy.

Equation (\ref{eq:learning_hybrid_general}) presents the learning of hybrid models in its most abstract form and we shall make it more specific shortly. 
We first present several ways to minimize the objective over the space $\Hyb=\tripletspace$ that differ on the order in
which the simple $h_s$ and the black box $h_c$ parts are trained.

\subsubsection{The \postbb{} Paradigm: Wrapping an Interpretable Model around the Complex One}

A common approach encompassing all state-of-the-art methods for learning hybrid models consists in training a black box 
first and then wrapping an interpretable model on top of it. We coin this strategy as the \postbb{} paradigm.
In this setting, the interpretable components $h_s$ and $\Omega$ can be seen as a way to simplify the model in regions where it is overkill. 
A key advantage of this paradigm is that a user owning a pre-trained black box with high performance can easily wrap an interpretable 
model on top of it to get an increase of transparency (and possibly a generalization improvement as suggested by our theoretical analysis of 
\textbf{Section~\ref{subsec:theory_sweet_spot}}). Furthermore, the interpretable part of the hybrid model is able to correct the mistakes 
made by the black box, as its predictions are known in advance. We illustrate the \postbb{} paradigm in Figure \ref{fig:two_paradigmes}~(Top).

\subsubsection{The \prebb{} Paradigm: black box Specialization by Reweighting}\label{sec:weighting_scheme}

\begin{figure}[t!]
    \centering
    \resizebox{\linewidth}{!}{
        \input{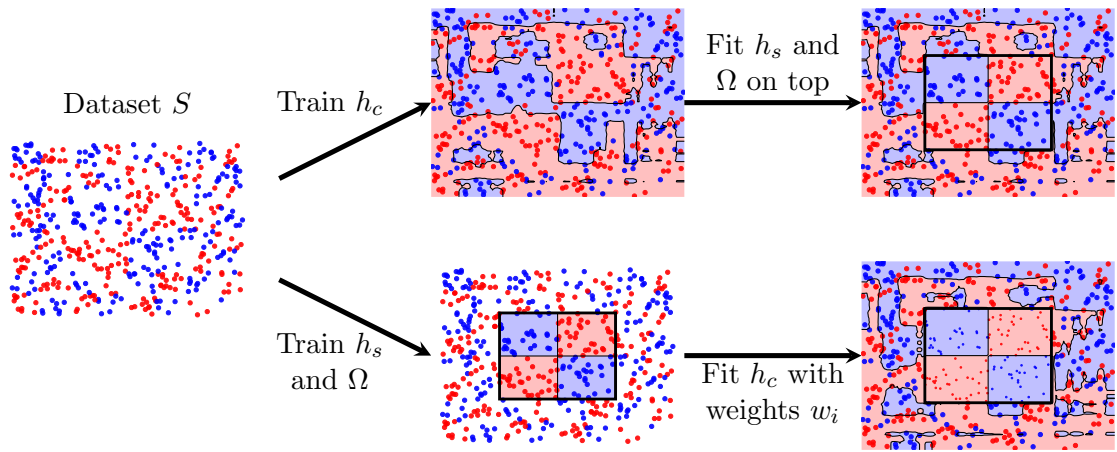}
    }
    \caption{Two paradigms for learning hybrid models. 
    (Top) In the \postbb{} paradigm, a black box is first
    trained on the whole dataset. Then, the interpretable components are fitted on top of the black box to
    simplify it in regions where it is overkill.
    (Bottom) In the \prebb{} paradigm, the interpretable part of the model is trained to identify a 
    region where the task is simple. Afterward, the black box model is fitted on the
    data with specialization weights $w_i$ to encourage high performance on instances outside of $\Omega$. 
    Here the weights are visualized as the markers' size.}
    \label{fig:two_paradigmes}
\end{figure}

Another possibility for learning hybrid models consists in first learning the interpretable part of the model 
before training a black box model on the remaining examples. This approach, which we label \prebb{}, does not currently exist in the literature.
The objective of the initial training of the interpretable part is to identify the easiest examples from the data and train a simple model on them.
Then, the black box part will only have to classify the examples not sent to the simple part ($\bm{x}\notin \Omega$).
Leveraging the black box complexity to specialize it on such part of the input space could hence lead to enhanced performances.
However, it could also cause overfitting, especially when the interpretable part transparency is high (and the black box only deals with 
a small portion of the input space/a reduced number of examples). In our proposed framework, this issue is tackled by training the 
black box on a reweighted version of the entire training set, with weights
\begin{equation}
    \forall i \in \{1, 2,\ldots,M\},\quad w_i = 
    \frac{e^{\alpha \, \indicator{}[\bm{x}^{(i)}\in \overline{\Omega}\,]}}
    {\sum_{j=1}^{M}{e^{\specializationcoefficient \, \indicator{}[\bm{x}^{(j)}\in \overline{\Omega}\,]}}}, \label{eq:weighting_scheme}
\end{equation}
that are higher for instances not classified by $h_s$. The non-uniform weights rely on a \textbf{specialization coefficient} 
$\specializationcoefficient\geq0$. The higher $\specializationcoefficient$, the more the black box focuses on the data not captured by the 
interpretable part of the model. On the other hand, low values of $\specializationcoefficient$ (\emph{e.g.,} for 
$\specializationcoefficient=0$, all examples' weights are equal) lead to a more generalist black box model. 
Since this trade-off is non-trivial, the hyperparameter $\specializationcoefficient$ will need to be 
fine-tuned in practice. Figure \ref{fig:two_paradigmes}~(Bottom) illustrates the
\prebb{} paradigm pipeline. We note that many classifiers in the \texttt{Scikit-Learn} and
\texttt{XGBoost} packages support non-uniform data weights in their training procedure. 
Hence, the \prebb{} paradigm is
also black box-agnostic.

This paradigm intrinsically comes with several drawbacks and advantages.
On the one side, the \prebb{} paradigm limits the possible collaboration between both parts of the hybrid model. 
Indeed, the interpretable part (characterized by $h_s$ and $\Omega$) is trained first, defining the data split with the black box part.
Then, there is no possibility to redefine the data split between the two parts of the hybrid model in the second phase of the learning (black box training).
Consequently, there can be no correction of one part of the model's errors by the other, as was done in the \postbb{} paradigm.
On the other side, because the data split is perfectly defined while training the black box, it is possible to adapt the black box training procedure to leverage its complexity and \emph{specialize} it on its support region $\overline{\Omega}$.

\subsubsection{Another Perspective: End-to-end Approach}

Finally, a last possible strategy consists in training both parts of a hybrid model \textit{simultaneously}. 
While this approach could theoretically provide the best performances (as it allows for a global optimality guarantee), it is also very challenging, 
as it requires encoding both the simple and black box parts of the hybrid model within a unified framework.

One key applicability advantage of both \prebb{} and \postbb{} paradigms is their black box-agnostic nature: there is no limitation over 
the type of black box used nor its training procedure. This would not hold anymore in an end-to-end paradigm, and we let such an approach as an 
interesting avenue for future works.

In the coming subsection, we discuss how rule-based models can be used for implementing the triplet $\triplet$.

\subsection{Rule-Based Modeling}

One of the important design choices of a hybrid model is the space $\mathcal{P}$ of possible subsets $\Omega$
where the interpretable model will operate. An example from previous work is to model these sets via thresholded 
linear models \citep{wang2021hybrid}. An alternative way to encode the input subsets $\Omega$ is by employing a 
rule-based model $r$ (\emph{e.g.,} a rule list or a rule set) and defining $\Omega_r$ as
\begin{equation*}
    \Omega_r := \{\bm{x}\in \X : \text{cover}(r, \bm{x})=1\},
\end{equation*}
where $\text{cover}(r, \bm{x})=1$ if $\bm{x}$ respects the condition in at least one of the rules in $r$ (we say that $\bm{x}$ is \emph{captured} by $r$). 
The advantage of using rule-based models to partition the input space is that they are interpretable by design, hence they can also serve as 
the simple hypothesis space $\Hs$. That is, we can assign a label to an input depending on which rule captures it. Hereafter is an example 
of a hybrid model involving a rule list $r$ containing two rules.
\begin{figure}[h]
    \centering
    \begin{minipage}{.5\linewidth}
    \begin{algorithmic}
    \If{$18 \leq \text{Age} \leq 22$ \textbf{and} \text{gender}=\text{male}} 
        \State \Return $y=1$
    \ElsIf{$\text{Prior-Crimes} > 3$} 
        \State \Return $y=1$
    \Else
        \State \Return $h_c(\bm{x})$
    \EndIf 
    \end{algorithmic}
    \end{minipage}
\end{figure}

Since a rule-based model encodes both the region $\Omega$ and the simple function $h_s$ on this region, we can think of rule-based hybrid 
models as a tuple $\tuple\in\Hc\times\Hs$ instead of a triplet $\triplet$. The learning objective on the training set $S$ becomes

\begin{equation}
    \obj(\tuple, S) = \frac{\emploss{S}(\tuple)}{\lvert S \rvert} + \lambda \cdot \lenrulelist[r]  + \beta \cdot \frac{|S\cap \overline{\Omega}_r|}{|S|},
    \label{eq:obj_rule}
\end{equation}
where we measure the complexity of a rule-list (rule-set) $r$ by its length $\lenrulelist[r]$.

\subsection{Rule-Based \postbb{} Hybrid Models}

Now that we have introduced several learning paradigms as well as a modeling choice for the hybrid model based on rules, we can describe two
approaches in the literature that apply the \postbb{} paradigm with rule-sets and rule-lists.

\subsubsection{Hybrid Rule-Set (HyRS)}
\label{subsec:hyrs}

\begin{wrapfigure}{r}{0.35\textwidth}
    \centering
    \vspace{-0.2cm}
    \begin{algorithmic}
    \If{$\text{cover}(r_+, \bm{x})$} 
        \State \Return $1$
    \ElsIf{$\text{cover}(r_-, \bm{x})$} 
        \State \Return $0$
    \Else
        \State \Return $h_c(\bm{x})$
    \EndIf 
    \end{algorithmic}
    \caption{Hybrid Rule-Set.}
    \label{fig:HyRS_example}
\end{wrapfigure}

This hybrid model has been introduced by \citet{wang2019gaining} and considers a rule set $r=r_+ \cup r_-$ that combines a set of positive 
rules $r_+$ and a set of negative rules $r_-$. The resulting hybrid model $\tuple$ takes the form of Figure \ref{fig:HyRS_example}.

The complexity of the interpretable model is the total number of rules $|r|$ and so the learning objective of Equation \ref{eq:obj_rule} is used.
The minimization of this combinatorial problem is tackled by a local search algorithm where neighborhoods are defined as random perturbations of 
the rule-sets $r_+$ and $r_-$.

\begin{figure}[t!]
    \centering
    \begin{subfigure}[b]{0.49\textwidth}
        \centering
        \includegraphics[width=\textwidth]
        {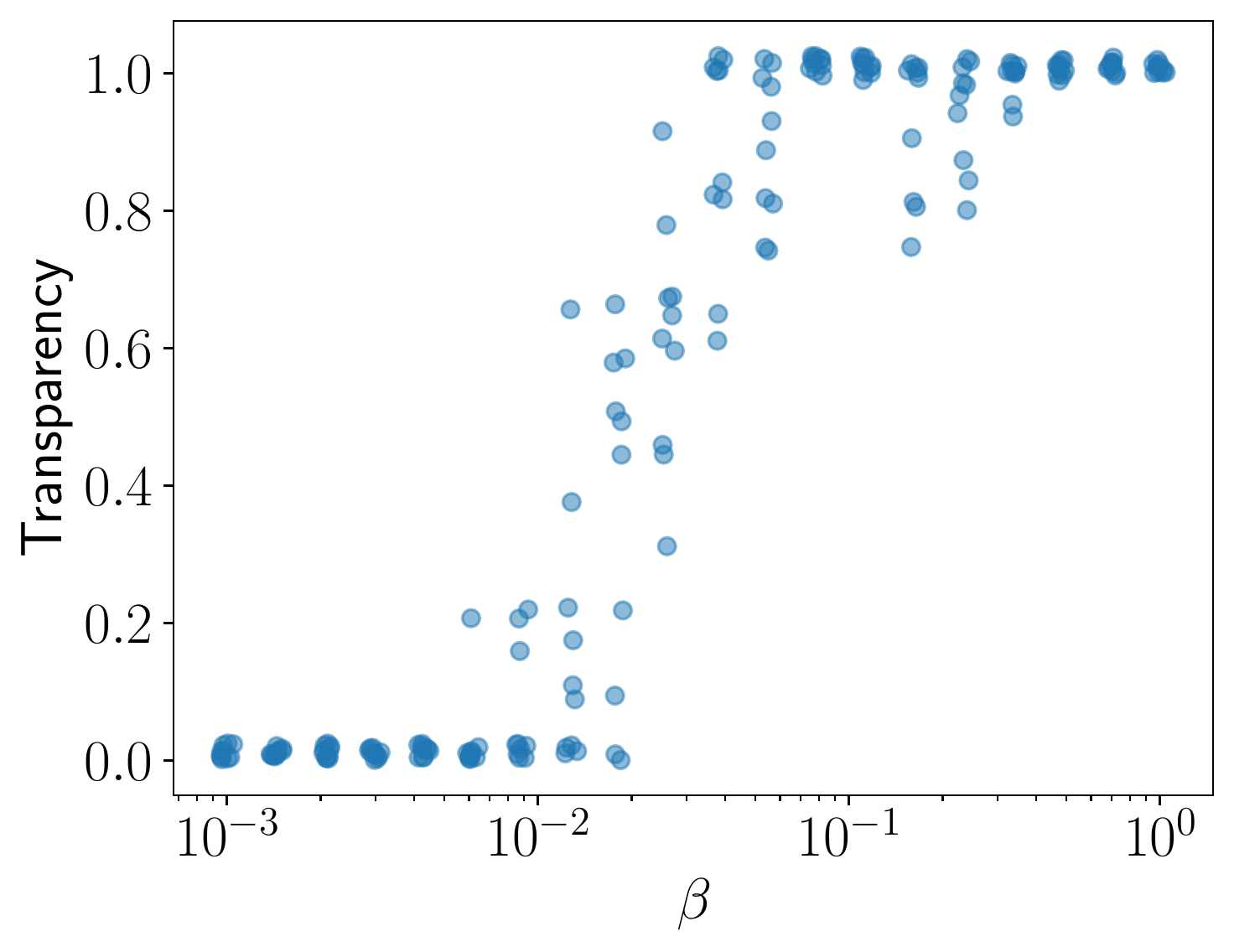}
        \caption{COMPAS dataset}
    \end{subfigure}
    \hfill
    \begin{subfigure}[b]{0.49\textwidth}
        \centering
        \includegraphics[width=\textwidth]
        {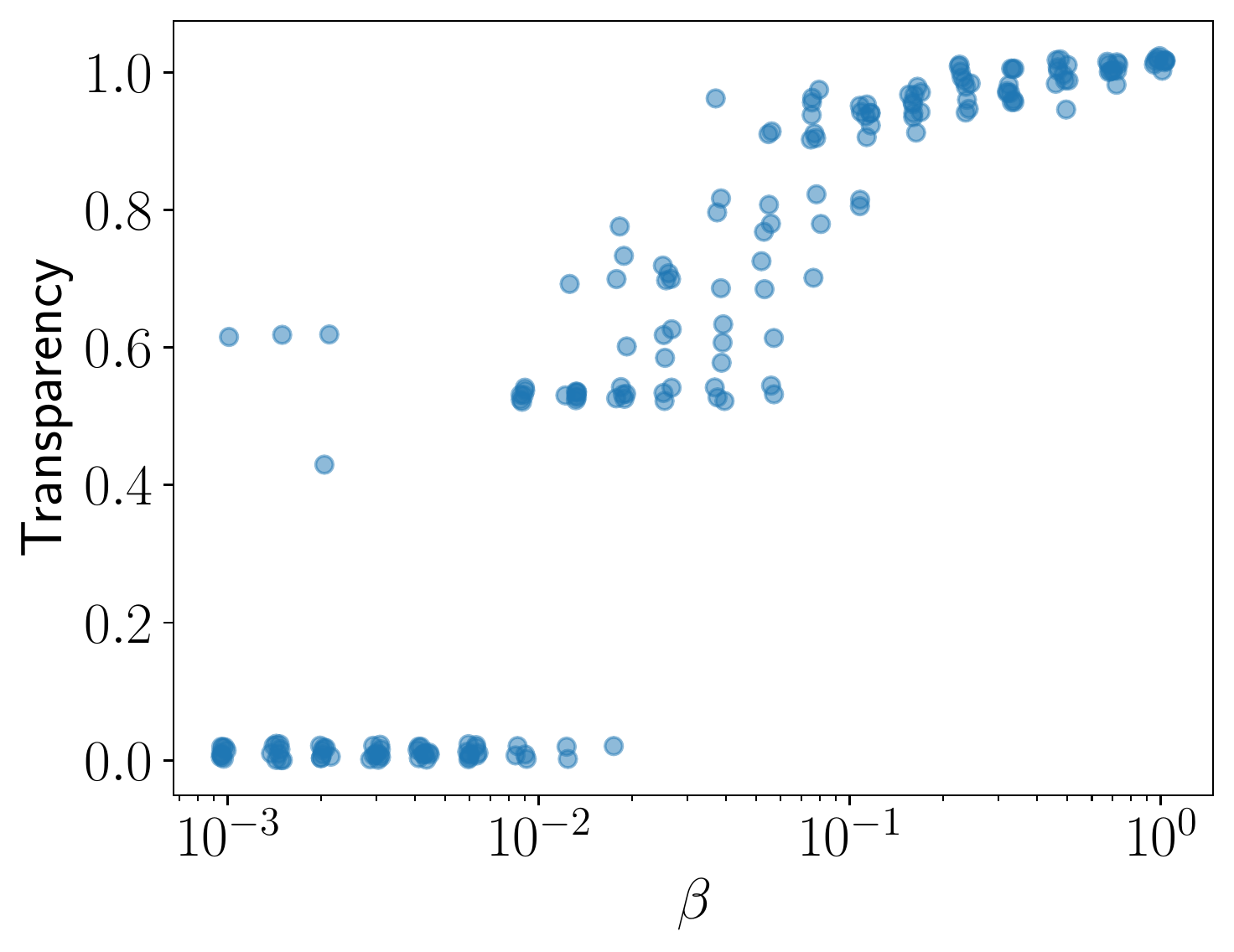}
        \caption{UCI Adult Income dataset}
    \end{subfigure}
    \caption{Instability of the transparency of HyRS for different 
    random seeds. A small jitter was applied to the points to
    remove juxtapositions.}
    \label{fig:transparency_instability}
\end{figure}

One of the drawbacks of HyRS is that the user does not have precise control over the transparency $C_\Omega$ of the
resulting hybrid model. There are two design choices in HyRS that lead to this issue. First of all, the only way to control the desired transparency is 
to increase the hyper-parameter $\beta$ which will incentivize the rule sets to cover more examples. Still, because the objective is extremely complex, 
it is not clear what $\beta$ is high enough to ensure a certain level of transparency. Secondly, since the local search algorithm employed to find 
the rules is inherently stochastic, several runs of the training procedure with the same hyperparameters can lead to very different models and, by extension, different transparencies. Figure~\ref{fig:transparency_instability} shows different reruns of HyRS on two datasets for 20 different values of 
$\beta$ that span four orders of magnitude. We see that the relation between transparency and $\beta$ is hardly monotonic because of the variance 
between reruns. Moreover, the transparency does not vary smoothly w.r.t $\beta$ as seen in the UCI Adult Income dataset, where the transparency 
jumps from $0$ to $0.5$ at around $\beta=10^{-2}$.

\subsubsection{Companion Rule-List (CRL)}\label{subsec:crl}

One year after the invention of HyRS, an alternative method called Companion-Rule-List (CRL) has been developed in order to address previous limitations~\citep{pan2020interpretable}. Notably, CLR simplifies the exploration of compromises between accuracy and transparency by 
returning multiple hybrid models with increasing transparency instead of a single model. In order to encode multiple hybrid models, CRL exploits 
the fact that, given a rule list, one can insert the black box at any level of the \textbf{else-if} statements. For instance, Figure~\ref{fig:crl_example}
presents three hybrid models $\tuple$ that are derived from the same list of three rules $r=[r_1, r_2, r_3]$.

\begin{figure*}[h]
    \begin{center}
        
    \begin{subfigure}[t]{0.3\textwidth}
        \centering
        \begin{algorithmic}
        \If{$\text{cover}(r_1, \bm{x})$} 
            \State \Return $1$
        \Else
            \State \Return $h_c(\bm{x})$
        \EndIf 
        \end{algorithmic}
    \end{subfigure}
    \hfill
    \begin{subfigure}[t]{0.3\textwidth}
        \centering
        \begin{algorithmic}
        \If{$\text{cover}(r_1, \bm{x})$} 
            \State \Return $1$
        \ElsIf{$\text{cover}(r_2, \bm{x})$} 
            \State \Return $0$
        \Else
            \State \Return $h_c(\bm{x})$
        \EndIf 
        \end{algorithmic}
    \end{subfigure}
    \hfill
    \begin{subfigure}[t]{0.3\textwidth}
        \centering
        \begin{algorithmic}
        \If{$\text{cover}(r_1, \bm{x})$} 
            \State \Return $1$
        \ElsIf{$\text{cover}(r_2, \bm{x})$} 
            \State \Return $0$
        \ElsIf{$\text{cover}(r_3, \bm{x})$} 
            \State \Return $1$
        \Else
            \State \Return $h_c(\bm{x})$
        \EndIf 
        \end{algorithmic}
    \end{subfigure}
   
    \caption{How a single rule list  $r=[r_1, r_2, r_3]$ can encode three hybrid models $\tuple$ with increasing transparency (from left to right).}
    \label{fig:crl_example}
    
    \end{center}
\end{figure*}

\begin{figure}[t!]
    \centering
    \begin{subfigure}[b]{0.49\textwidth}
        \centering
        \includegraphics[width=\textwidth]
        {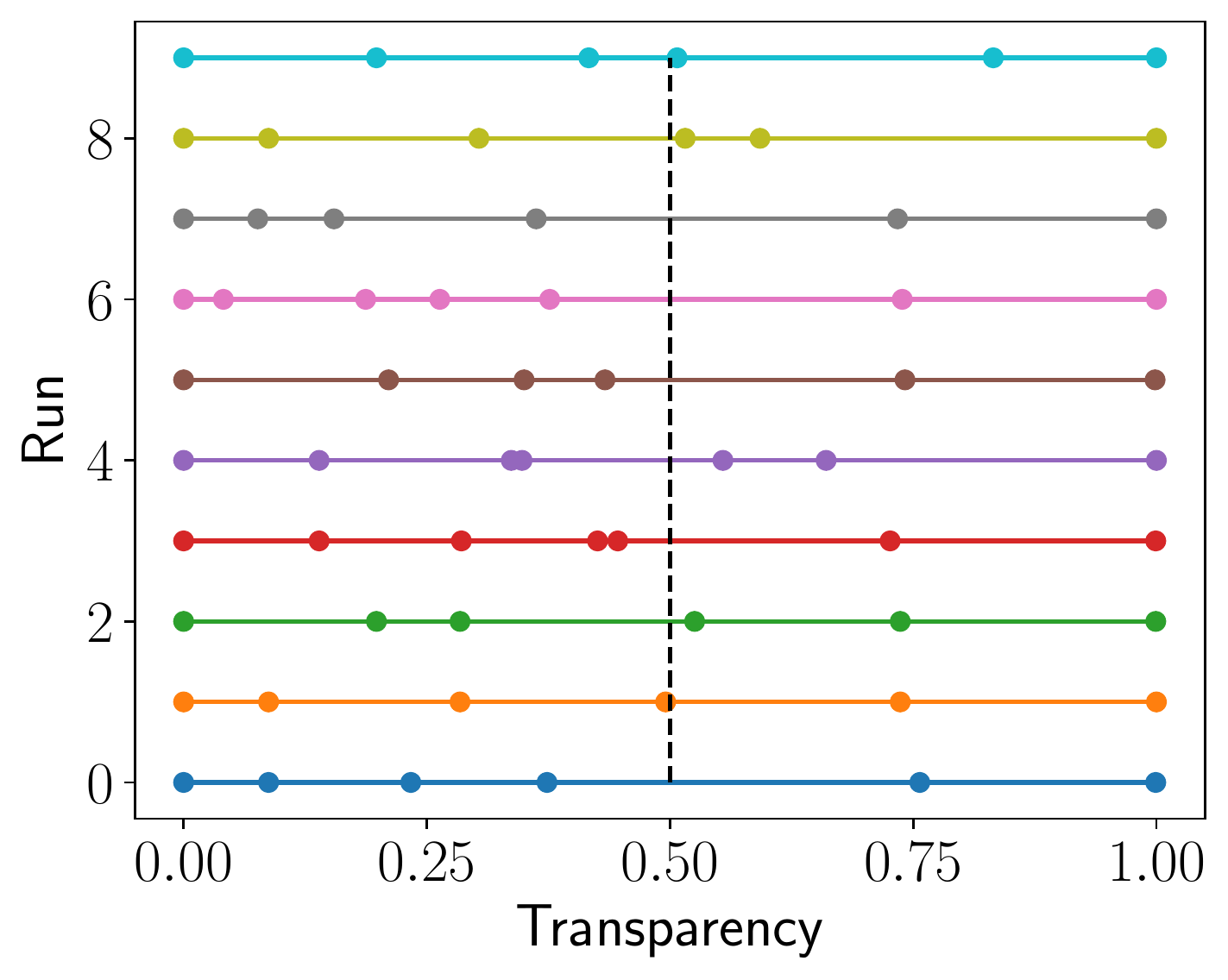}
        \caption{COMPAS dataset}
    \end{subfigure}
    \hfill
    \begin{subfigure}[b]{0.49\textwidth}
        \centering
        \includegraphics[width=\textwidth]
        {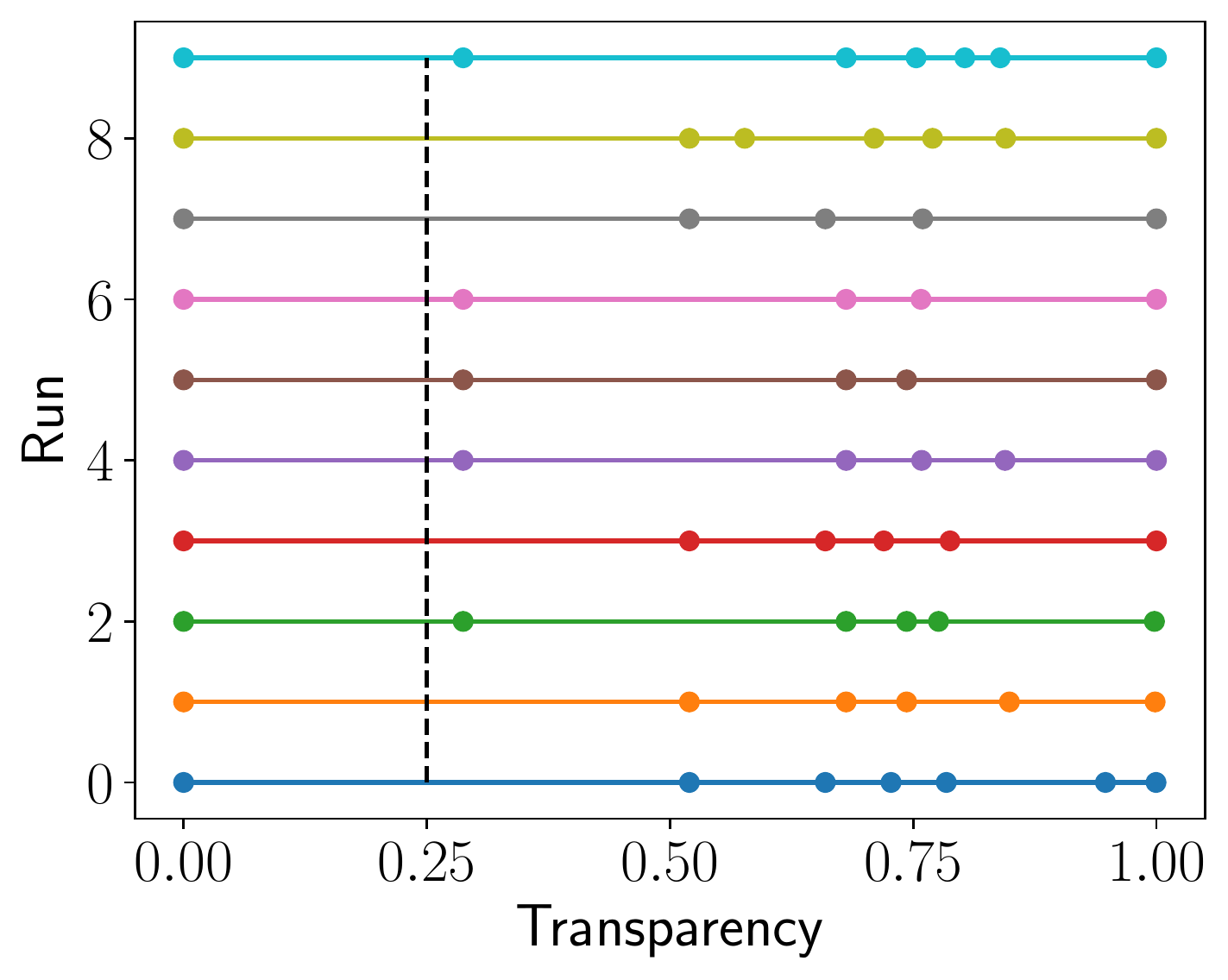}
        \caption{UCI Adult Income dataset}
    \end{subfigure}
    \caption{Instability of the transparency of CRL for different random seeds each indicated by a different color. 
    The dots represent the transparency attained by the many hybrid models returned from a single run of CRL.}
    \label{fig:transparency_instability_CRL}
\end{figure}

By returning multiple hybrid models via one call of the training function, CRL allows users to decide what hybrid model to use based on their desired
transparency. The training objective of CRL is no longer the accuracy but rather the Area-Under-the-Curve (AUC) of the accuracy-transparency 
curve of the different hybrid models. A regularisation $\lambda \cdot |r|$ is also added to the objective to avoid long rule-lists. Similarly to 
HyRS, CRL is trained with a local search algorithm where neighborhoods are defined as random perturbations of the rule-list $r$. Although 
CRL offers more possibilities for transparency, we find that the inherent stochasticity of the learning procedure still hinders the ability to 
consistently reach target transparency. Figure \ref{fig:transparency_instability_CRL} presents simple experiments conducted on the COMPAS and UCI 
Adult Income datasets where a CRL model was fitted for 10 different random seeds. We present the different levels of transparency attained by each 
run. For the COMPAS dataset, we note that if a user wishes for a transparency of at least 0.5, then on half of the runs, they would need to go up 
to about 0.75 transparency using the CRL framework (which may excessively conflict with predictive accuracy). For the UCI Adult Income dataset, 
if an end-user requires transparency of at least 0.25, then on half of the runs, they would need to go up to 0.5 transparency. These experiments 
highlight that CRL does not provide full control over the desired level of transparency of the hybrid models.

\newpage
\section{\HybridCORELS{}: Learning Optimal Hybrid Interpretable Models}


\label{sec:hybridcorels}

We now present our methods for learning optimal hybrid models. 
First, we introduce the \corels{} algorithm, which was proposed to learn optimal rule lists. 
Then, we describe the integration of the transparency requirements within our proposed methods. 
Finally, we propose \HybridCORELSPost{} (resp. \HybridCORELSPre{}), a modified version of \corels{} to 
learn optimal hybrid models following the \postbb{} (respectively, \prebb{}) framework.


\subsection{Learning Optimal Rule Lists: the \corels{} Algorithm}
\label{sec:hybridcorels_corels}

Rule lists are interpretable classifiers formed by an ordered list of if-then rules $\prefix$, followed by a default prediction $\defconseq$~\citep{DBLP:journals/ml/Rivest87}. 
The set of ordered rules preceding the default prediction is called a prefix. 
One can observe that any rule list $\rulelist = (\prefix, \defconseq)$ represents a classification function, while any prefix $\prefix$ defines a \emph{partial} classification function, defined within its support $\Omega_\prefix$ (examples matching at least one of the rules within $\prefix$) .

To learn Certifiable Optimal RulE ListS, \cite{DBLP:journals/jmlr/AngelinoLASR17} proposed \corels{}, a branch-and-bound algorithm. It represents the search space of rule lists using a prefix tree, in which each node corresponds to a prefix $\prefix$. Adding a default prediction $\defconseq$ to $\prefix$ allows the building of a rule list $\rulelist = (\prefix, \defconseq)$. 
In \corels{}' prefix tree, the children nodes of $\prefix$ correspond to prefixes formed by adding exactly one rule at the end of $\prefix$. Thus, the  $\prefix$-rooted sub-tree corresponds to all possible extensions of $\prefix$.
\corels{}' objective function for rule list $\rulelist = (\prefix, \defconseq)$ on dataset $\dataset$ is a weighted sum of classification error and sparsity:
\begin{align}
    \obj(\rulelist,\dataset) = \frac{\errs[\rulelist,\dataset]}{\lvert \dataset \rvert} + \corelsreg \cdot \lenrulelist[\prefix] \label{eq:obj_original}
\end{align} 
where $\errs[\rulelist,\dataset]$ measures the number of errors (incorrect classifications) made by $\rulelist$ on $\dataset$ (as defined in~(\ref{eq:emp_loss})), and $\lenrulelist[\prefix]$ is the length (number of rules) of rule list $\rulelist$'s prefix $\prefix$.

Let $\dataset_\prefix = S \cap \Omega_\prefix$ be the subset of $\dataset$ made of all examples of $\dataset$ captured by some prefix $\prefix$ (\emph{i.e.,} the examples classified by $\prefix$'s partial classification function). 
Just like any branch-and-bound algorithm, \corels{} uses an objective lower bound to prune the prefix tree, and eventually guide the search in a best-first search fashion. 
For each node of the prefix tree (corresponding to a prefix $\prefix$), it measures the best objective function value that may be reached by extending prefix $\prefix$. 
If this value is worse than the best solution (rule list) known so far, then the $\prefix$-rooted sub-tree can be pruned safely. 
Let $\errs[\prefix, \dataset_\prefix]$ counts the number of mistakes made by prefix $\prefix$ (measured on its support set $\dataset_\prefix$), and $\incons(\dataset)$ denote the minimum number of examples of $\dataset$ that can never be classified correctly, because they have the exact same features vector as some other examples, but with a different label (due to potential dataset inconsistencies). 
\corels{}' objective lower bound  for prefix $\prefix$ on dataset $\dataset$ is then computed as follows:
\begin{align}  
\lb(\prefix,\dataset) = \frac{\errs[\prefix, \dataset_\prefix] + \incons(\dataset \setminus \dataset_\prefix)}{\lvert \dataset \rvert} + (\lenrulelist[\prefix]+1) \cdot \corelsreg \label{eq:lb_original}
\end{align}
Intuitively, $\errs[\prefix, \dataset_\prefix] + \incons(\dataset \setminus \dataset_\prefix)$ corresponds to the minimum number of errors that any extension of $\prefix$ can make, given the errors made by $\prefix$ and the errors that can not be avoided due to data inconsistency.

\corels{} uses several efficient data structures to speed up the computation by breaking down symmetries~\citep{DBLP:journals/jmlr/AngelinoLASR17}.
For instance, a prefix permutation map ensures that only the most accurate permutation of every set of rules is kept. These data structures are still valid in our setup.
Finally, we can leverage the efficiency of the \corels{}' machinery to learn optimal hybrid models, by only modifying \corels{}' objective function and providing a valid lower bound on the new objective function.
For reference, we provide the pseudo-code of the branch-and-bound underlying \corels{} within Algorithm~\ref{alg:corels} in Appendix~\ref{appendix:pseudo_codes_corels}.
In particular, our modified algorithms will only learn prefixes (which will constitute the interpretable parts of our hybrid models), and hence will never care about the default prediction.
In sections~\ref{sec:hybridcorels_post} and~\ref{sec:hybridcorels_pre}, we show how the objective function~(\ref{eq:obj_original}) and its lower bound~(\ref{eq:lb_original}) can be modified to learn hybrid models implementing the \postbb{} and \prebb{} paradigms (respectively).

\subsection{Ensuring a User-Defined Transparency Level} 
\label{sec:hybridcorels_transparency}

State-of-the-art methods for learning hybrid models integrate transparency requirements using a regularization term, as described in sections~\ref{subsec:hyrs} and~\ref{subsec:crl}. However, this approach does not allow the user to have a precise control over the desired transparency level, and several runs with the exact same hyperparameters but different random seeds can lead to hybrid models with significantly different transparency levels.
To address this issue, we build on the flexibility of the branch and bound algorithm underlying \corels{} 
and integrate transparency as a hard constraint, stating that the learnt prefix $\prefix$ must capture at least a proportion of $\mintransp \in [0,1]$ of the examples within dataset $\dataset$:
\begin{align}
\frac{\lvert \dataset_\prefix \rvert}{\lvert \dataset \rvert} \geq \mintransp \label{eq:transparency_constraint}
\end{align}
where, as aforementioned, $\dataset_\prefix = S \cap \Omega_\prefix$ is the subset of $\dataset$ made of all examples of $\dataset$ captured by prefix $\prefix$.
Both our proposed approaches implement this hard-constraint approach. It allows for the building of hybrid models whose transparency (on the training set) is guaranteed to be at least $\mintransp$. 
To the best of our knowledge, our approach is the first to implement such direct control of the transparency level. 
Compared to state-of-the-art hybrid learning methods (which use a regularization term to encourage transparency), this approach allows for a tight control of the desired transparency, which can help build denser sets of tradeoffs between transparency and utility using $\epsilon$-constrained methods.
To enforce constraint~(\ref{eq:transparency_constraint}) using the \corels{} branch-and-bound algorithm, we simply modify the best solution update subroutine, to only perform the update operation if the candidate prefix satisfies the transparency requirement. 
This guarantees that any returned solution will satisfy~(\ref{eq:transparency_constraint}) while maintaining optimality as the exploration and bounds are not modified.

Even if constraint~(\ref{eq:transparency_constraint}) ensures the strict respect of a user-defined transparency level, we also integrate transparency using a regularization term. 
This allows to break ties: if two models exhibit the same accuracy and sparsity levels, then this regularization term will favor the one with higher transparency. 
In practice, we set the associated regularization coefficient $\regcoeffhycorels$ to a value small enough to only break ties.
Indeed, because it is already enforced through hard constraint~(\ref{eq:transparency_constraint}), we do not want transparency to trade-off with accuracy nor sparsity in the objective function (\emph{i.e.,} we will always prefer any non-zero improvement on the accuracy or sparsity term over any improvement on the transparency term).
Just like in constraint~(\ref{eq:transparency_constraint}), transparency is measured using $\frac{\lvert \dataset_\prefix \rvert}{\lvert \dataset \rvert} \in [0,1]$ (as $\dataset_\prefix \subseteq \dataset$). Thus, we penalize (un)transparency as $\frac{\lvert \dataset \setminus \dataset_\prefix \rvert}{\lvert \dataset \rvert} \in [0,1]$ in the objective function, and set $\regcoeffhycorels < \frac{1}{\lvert \dataset{} \rvert} \leq \corelsreg$ for both approaches. Finally, our objective functions~(\ref{eq:obj_post}) and~(\ref{eq:obj_pre}) both add this $(\regcoeffhycorels \cdot \frac{\lvert \dataset \setminus \dataset_\prefix \rvert}{\lvert \dataset \rvert})$ term.

\subsection{\postbb{} framework: \HybridCORELSPost{}}
\label{sec:hybridcorels_post}

We now introduce \HybridCORELSPost{}, a modified version of the \corels{} algorithm to produce optimal hybrid models using the state-of-the-art \postbb{} paradigm.
More precisely, \HybridCORELSPost{} first trains a black-box model (or takes as input a pre-trained black-box model). 
This first step is totally agnostic to the type of black-box and its training algorithm.
Then, given a minimum transparency constraint~(\ref{eq:transparency_constraint}), it builds a prefix optimizing the overall model's accuracy and sparsity. 
\paragraph{Objective} Given a black-box $h_c$'s training set predictions, \HybridCORELSPost{} builds a prefix $\prefix$ capturing at least a proportion of $\mintransp$ of the training data (transparency constraint~(\ref{eq:transparency_constraint})), and minimizing the following objective function: 
\begin{equation}
    \begin{aligned}
    \obj_\text{post}(\prefix,\dataset)
    &=\frac{\errs[\tuple, S]}{\lvert \dataset \rvert} + \corelsreg \cdot \lenrulelist[\prefix] + \regcoeffhycorels \cdot \frac{\lvert \dataset \setminus \dataset_\prefix \rvert}{|S|}\\
    &=\frac{\errs[\prefix, \dataset_\prefix]+\errs[h_c, \dataset \setminus \dataset_\prefix]}{\lvert \dataset \rvert} + \corelsreg \cdot \lenrulelist[\prefix] + \regcoeffhycorels \cdot \frac{\lvert \dataset \setminus \dataset_\prefix \rvert}{\lvert \dataset \rvert}.
    \end{aligned}
    \label{eq:obj_post}
\end{equation} 
Here, $ \frac{\errs[\prefix, \dataset_\prefix]+\errs[h_c, \dataset \setminus \dataset_\prefix]}{\lvert \dataset \rvert}$ is the exact accuracy of the overall hybrid model. Indeed, because the black-box predictions are fixed, the interpretable part is able to correct the mistakes made by the pre-trained black-box model.

\paragraph{Objective lower bound} \corels{}' original objective lower bound 
(\ref{eq:lb_original}) (leveraging both the prefix's errors and the inconsistent examples among the uncaptured ones) is still valid and tight in this setup, so we do not need to modify it. 
Indeed, the error term lower bound $\errs[\prefix, \dataset_\prefix] + \incons(\dataset \setminus \dataset_\prefix)$ is unchanged, as all remaining black-box errors $\errs[\blackbox, \dataset \setminus \dataset_\prefix]$ may potentially be corrected by extending $\prefix$, but the errors already made by prefix $\prefix$ and those related to remaining inconsistencies can not be avoided.
Then, the transparency regularization term can not be used within the objective lower bound, as this term can always reach $0.0$ by sufficiently extending prefix $\prefix$. 
Finally, the lower bound over the sparsity regularization term still holds: any extension of prefix $\prefix$ must have at least $\lenrulelist[\prefix] + 1$ rules.

Finally, \HybridCORELSPost{} is an exact method: it provably returns a prefix $\prefix$ for which $\obj_{post}(\prefix,\dataset)$~(\ref{eq:obj_post}) is the smallest among those satisfying the transparency constraint~(\ref{eq:transparency_constraint}). 
This means that, given fixed black-box predictions and desired transparency level, it produces an optimal hybrid interpretable model in terms of accuracy/sparsity.
We provide a detailed pseudo-code of \HybridCORELSPost{} in Algorithm~\ref{alg:hybridcorels_post} in the Appendix~\ref{appendix:pseudo_codes_hybridcorels_post}.

\subsection{\prebb{} framework: \HybridCORELSPre{}}
\label{sec:hybridcorels_pre}

\HybridCORELSPre{} is the first algorithm to implement our proposed \prebb{} paradigm for learning hybrid interpretable models. It first builds a prefix optimizing accuracy and sparsity, given a minimum transparency constraint~(\ref{eq:transparency_constraint}). 
Then, it trains the black-box part of the hybrid model, specializing it on the uncaptured examples, using the weighting scheme~(\ref{eq:weighting_scheme}). 
As aforementioned in \textbf{Section~\ref{sec:weighting_scheme}}, the \prebb{} paradigm intrinsically limits the possible collaboration between both parts of the hybrid model, as it is not possible for the black-box part to correct the mistakes made by the interpretable part. However, it is possible to consider the inconsistencies left to the black-box part while training the interpretable part, which implements a form of collaboration. 

\paragraph{Objective} \HybridCORELSPre{} builds a prefix $\prefix$ capturing at least $\mintransp$ of the training data (transparency constraint~(\ref{eq:transparency_constraint})), and minimizing the overall hybrid model's classification error lower bound (based on both prefix $\prefix$'s errors and the inconsistencies let to the black-box part) and sparsity: 
\begin{align}
    \obj_\text{pre}(\prefix,\dataset) = \frac{\errs[\prefix,\dataset_r] + \incons(\dataset \setminus \dataset_\prefix)}{\lvert \dataset \rvert} + \corelsreg \cdot \lenrulelist[\prefix] + \beta \cdot \frac{\lvert \dataset \setminus \dataset_r \rvert}{\lvert \dataset \rvert} \label{eq:obj_pre}
\end{align}
where the error term $\frac{\errs[\prefix,\dataset_r] + \incons(\dataset \setminus \dataset_\prefix)}{\lvert \dataset \rvert}$ corresponds to the entire hybrid model accuracy if the black-box performs perfectly (\emph{i.e.,} correctly classifies all examples except the inconsistent ones). It hence provides a tight upper-bound on the entire hybrid model accuracy. 

\paragraph{Objective lower bound} \corels{}' original objective lower bound $\lb(\prefix,\dataset)$ (\ref{eq:lb_original}) (leveraging both the prefix's errors and the inconsistent examples among the uncaptured ones) is still valid and tight in this setup, so we do not need to modify it. 
Indeed, the error term is tight: it is not possible for any extension of $\prefix$ to avoid the errors already made by $\prefix$ nor the inconsistencies within the remaining examples.
The sparsity term is also tight as any extension of $\prefix$ must have a length of at least $\lenrulelist[\prefix]+1$.
As for \HybridCORELSPost{}, the (un)transparency term can not be used within the objective lower bound, as it can always reach $0.0$.
An interesting observation is that $\lb(\prefix,\dataset) > \obj_\text{pre}(\prefix,\dataset)$ for any prefix $\prefix$ (since $\beta < \corelsreg$ as indicated in \textbf{Section~\ref{sec:hybridcorels_transparency}}). 
This means that, for any prefix $\prefix$ with sufficient transparency (\emph{i.e.,} satisfying the transparency constraint~(\ref{eq:transparency_constraint})), the algorithm will always discard any of its extensions as they increase the sparsity term and can not lower the error term (they can only equal it if they add no additional error).
In fact, extending a prefix $\prefix$ can only worsen its objective function (again, assuming that $\beta < \corelsreg$), and it will only be performed in order to meet the transparency constraint~(\ref{eq:transparency_constraint}).

Finally, \HybridCORELSPre{} is an exact method: it provably returns a prefix $\prefix$ for which $\obj_\text{pre}(\prefix,\dataset)$~(\ref{eq:obj_pre}) is the smallest among those satisfying the transparency constraint~(\ref{eq:transparency_constraint}). 
This means that, given desired transparency level, it produces an optimal prefix (interpretable part of the final hybrid model) in terms of overall hybrid model accuracy upper bound and sparsity.
If the black-box performs perfectly, then the overall model is certifiably optimal.
We provide a detailed pseudo-code of \HybridCORELSPre{} in Algorithm~\ref{alg:hybridcorels_pre} in the Appendix~\ref{appendix:pseudo_codes_hybridcorels_pre}.

We additionally introduce in the Appendix~\ref{appendix:hybridcorelsprenocollab} another possible implementation of the \prebb{} paradigm based on the \corels{} algorithm but optimizing an objective function different from that of \HybridCORELSPre{}. This new variant \HybridCORELSPreNoCollab{} learns a prefix by maximizing its accuracy on the subset $S_r$, without accounting for the task left to the black-box part. Appendix~\ref{appendix:hybridcorelsprenocollab_theory} provides a description of this algorithm and Appendix~\ref{appendix:hybridcorelsprenocollab_expes} empirically compares it with \HybridCORELSPre{}. The experiments confirm that \HybridCORELSPreNoCollab{} is not competitive with \HybridCORELSPre{} in medium to high transparency regimes, due to the lack of collaboration between both parts of the hybrid model.  





\section{Experiments}

In this section, we empirically evaluate our proposed algorithms. We first introduce our experimental setup.
Then, we use \HybridCORELSPre{} to show that the \prebb{} paradigm is suitable to learn hybrid interpretable models exhibiting interesting 
trade-offs between accuracy and transparency. We explore the parameters of this paradigm, such as the specialization coefficient, to assess 
their effect and utility. Afterwards, we compare \HybridCORELSPre{} and \HybridCORELSPost{} with two state-of-the-art methods for learning hybrid 
interpretable models: Hybrid-Rule-Set (HyRS) and Companion-Rule-List (CRL).
Our thorough experimental study demonstrates that our proposed approaches are strongly competitive with the state of the art, 
while also providing optimality guarantees and allowing tight control of the desired transparency.

\subsection{Setup}
\label{sec:expes_setup}

\paragraph{Datasets} In our experiments, we consider several datasets with various prediction tasks and sizes: 
\begin{itemize}
    \item The \textbf{COMPAS} dataset\footnote{\url{https://raw.githubusercontent.com/propublica/compas-analysis/master/compas-scores-two-years.csv}}(analyzed by~\cite{angwin2016machine}) contains 6,150 records from criminal offenders in the Broward County of Florida collected from 2013 and 2014. The corresponding binary classification task is to predict whether a person will re-offend within two years. 
    \item The \textbf{UCI Adult Income} dataset~\citep{Dua:2019} stores demographic attributes of
    48,842 individuals from the 1994 U.S. census. Its binary classification task is to predict whether or not a particular person makes more than 50K USD per year.
    \item The \textbf{ACS Employment} dataset~\citep{DBLP:conf/nips/DingHMS21} is an extension of the UCI Adult Income dataset that includes more recent Census data (2014-2018). The goal is to predict if a person is employed/unemployed based on 10 socioeconomic factors. The specific dataset contained 
    information on 203,358 constituents of the Texas state in 2018.
\end{itemize}

\paragraph{Rules mining} To ensure a fair comparison between hybrid models, we pre-mined a set of rules $\minedrules$ for each dataset. 
The various hybrid models were then restricted to select rules $r\in \minedrules$ and, therefore, any difference in performance between models is 
solely attributable to the learning algorithms and not the quality of the rules. To mine the rules, the datasets were first binarized using quantile 
for numerical features and one-hot encoding for categorical features. Afterwards, the FP-Growth algorithm \citep{han2000mining} was applied to identify 
rules of cardinality 1-2 and support of at least $1\%$. To these sets of rules, we also added the negation of each rule in the original binarized dataset. 
Finally, the 300 rules with the largest support were kept to generate $\minedrules$. We ended up with $|\minedrules|=230$ rules on COMPAS and 
$|\minedrules|=300$ on the UCI Adult Income and ACS Employment datasets.

\paragraph{Black-boxes} In all experiments we used the following \sklearn{}~\citep{scikit-learn} classifiers as black-boxes: a \texttt{RandomForestClassifier}, an \texttt{AdaBoostClassifier}, and a \texttt{GradientBoostingClassifier}. 
Such black-boxes are in line with the setup considered in the literature~\citep{wang2019gaining}. We further detail the hyper-parameters tuning of these models 
in sections~\ref{sec:expes_pre} and~\ref{sec:expes_all}. We note that the Hybrid models studied (HyRS, CRL, and HybridCORELS) are not tied to any 
specific black-box, nor to a specific implementation. Indeed they are black-box-agnostic by design.

\paragraph{Implementation details} Our algorithms \HybridCORELSPost{} and \HybridCORELSPre{} (as well as its \HybridCORELSPreNoCollab{} variant 
discussed in the Appendix~\ref{appendix:hybridcorelsprenocollab}) are integrated into a user-friendly Python module, publicly available on PyPI\footnote{\url{https://pypi.org/project/HybridCORELS}} 
and GitHub\footnote{\url{https://github.com/ferryjul/HybridCORELS}}. They build upon the original \corels{}~\citep{DBLP:journals/jmlr/AngelinoLASR17} C++ implementation\footnote{\url{https://github.com/corels/corels}} and its Python wrapper\footnote{\url{https://github.com/corels/pycorels}}. 
All experiments are run on a computing grid over a set of homogeneous nodes using Intel Platinum 8260 Cascade Lake @2.4Ghz CPU.

\paragraph{\HybridCORELS{} transparency regularization coefficient $\regcoeffhycorels$ setting} In all our experiments using 
\HybridCORELSPre{} or \HybridCORELSPost{}, we set the transparency regularization coefficient 
$\regcoeffhycorels = min(\frac{1}{2 \cdot \lvert \dataset \rvert}, \frac{\corelsreg}{2})$ to only break ties but ensure that 
no accuracy nor sparsity will be traded-off for transparency, which is already enforced as a hard constraint 
(as discussed in \textbf{Section~\ref{sec:hybridcorels_transparency}}).

\subsection{Exploring the \prebb{} Paradigm}
\label{sec:expes_pre}


\paragraph{Objective} The objective of this subsection is to assess the appropriateness of the proposed \prebb{} paradigm for learning accurate hybrid interpretable models. To this end, we use our proposed algorithm implementing this framework: \HybridCORELSPre{}, depicted in \textbf{Section~\ref{sec:hybridcorels_pre}}. More precisely, we aim to explore the effect of the \emph{specialization coefficient} on the performances of the produced models.

\paragraph{Setup} For the three datasets presented in \textbf{Section~\ref{sec:expes_setup}}, we use \HybridCORELSPre{} to produce hybrid interpretable models for several \emph{transparency} levels: low ($0.25$), medium ($0.5$), high ($0.75$, $0.85$) and very high ($0.95$). 
For the prefix building part, we optimize the hyperparameters of \HybridCORELSPre{} using grid search over the following values: $\corelsreg \in \{10^{-2}, 10^{-3}, 10^{-4}\}$, $\corelsminsupport \in \{0.01, 0.05, 0.10\}$, and the \emph{objective-guided}, \emph{lower-bound-guided}, and \emph{BFS} search policies. 
For each experiment, the prefix yielding the best (training) accuracy upper-bound (considering the prefix's errors as well as the inconsistencies left to the black-box part, as depicted in~(\ref{eq:obj_pre})) is retained.
The black-box part of the final hybrid model is then trained, and experiments are run for three different \sklearn{}~\citep{scikit-learn} black-boxes: an \texttt{AdaBoostClassifier} with default parameters, a \texttt{GradientBoostingClassifier} with default parameters and a \texttt{RandomForestClassifier} with $min\_samples\_split=10$ and $max\_depth=10$.
Each black-box is retrained using different values for the specialization coefficient $\specializationcoefficient$, ranging from $0$ (no specialization) to $10$ (highly specialized). Experiments are run for five different train/test splits, with 80\% of the data used for training and the remaining 20\% for testing.

\begin{figure*}[htbp]
    \begin{center}
        
    \begin{subfigure}{\textwidth}
        \centering
        \label{fig:results_best_prefixes_acs_employ}
         \includegraphics[width=0.49\textwidth]{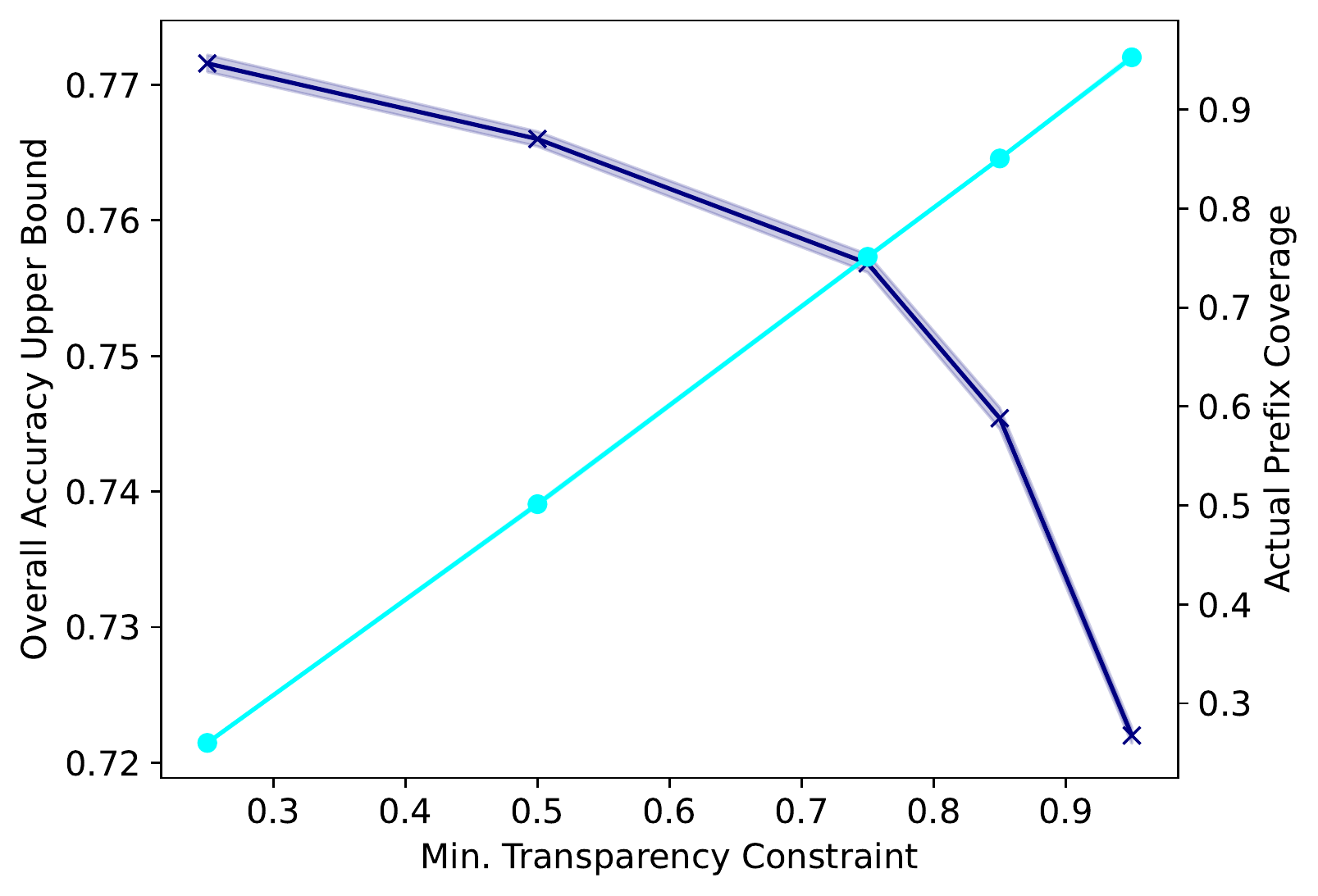}
        \includegraphics[width=0.49\textwidth]{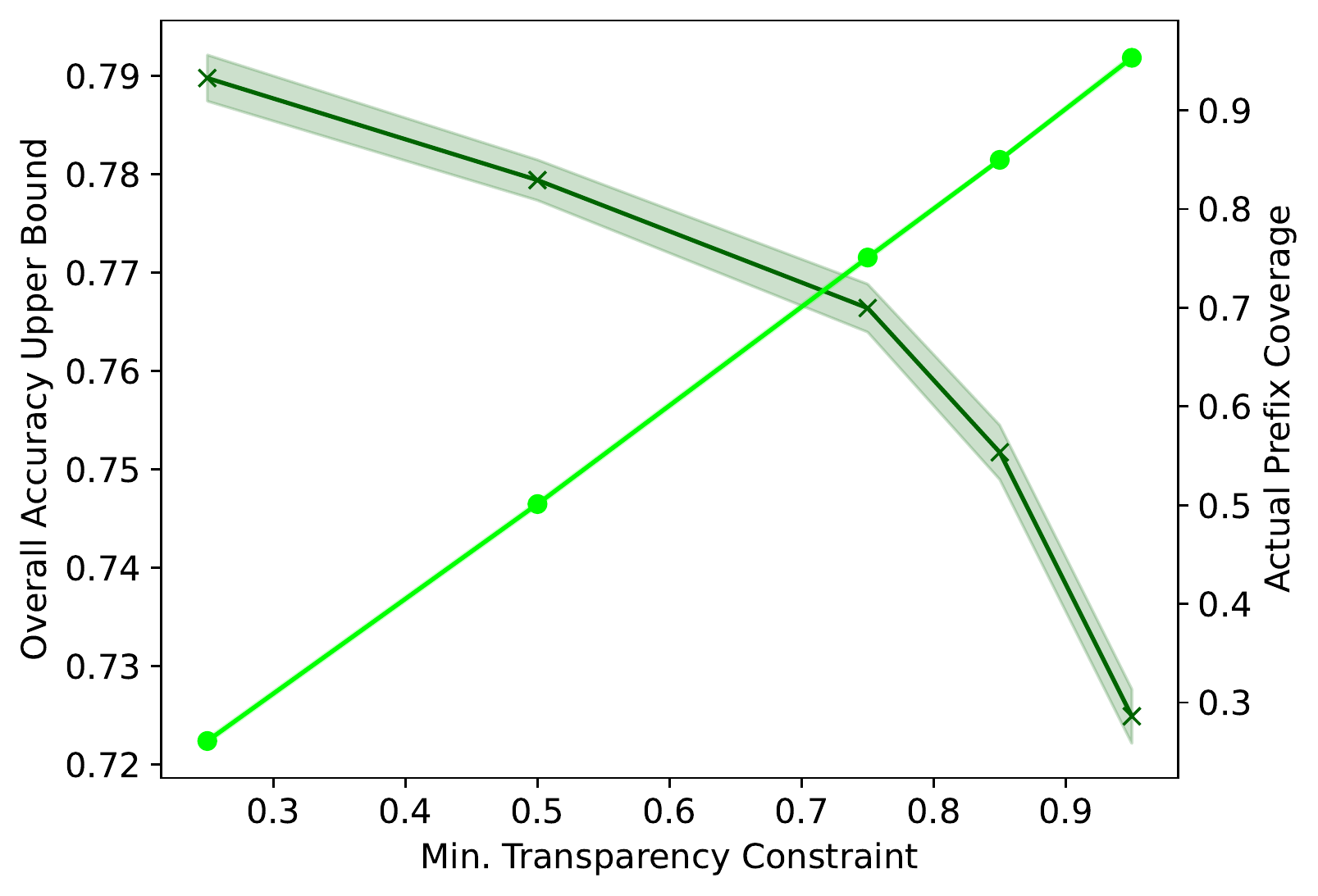}
        \caption{ACS Employment dataset.}
    \end{subfigure}
    
     \vskip 10 pt
     
    \begin{subfigure}{\textwidth}
        \centering
        \label{fig:results_best_prefixes_adult}
         \includegraphics[width=0.49\textwidth]{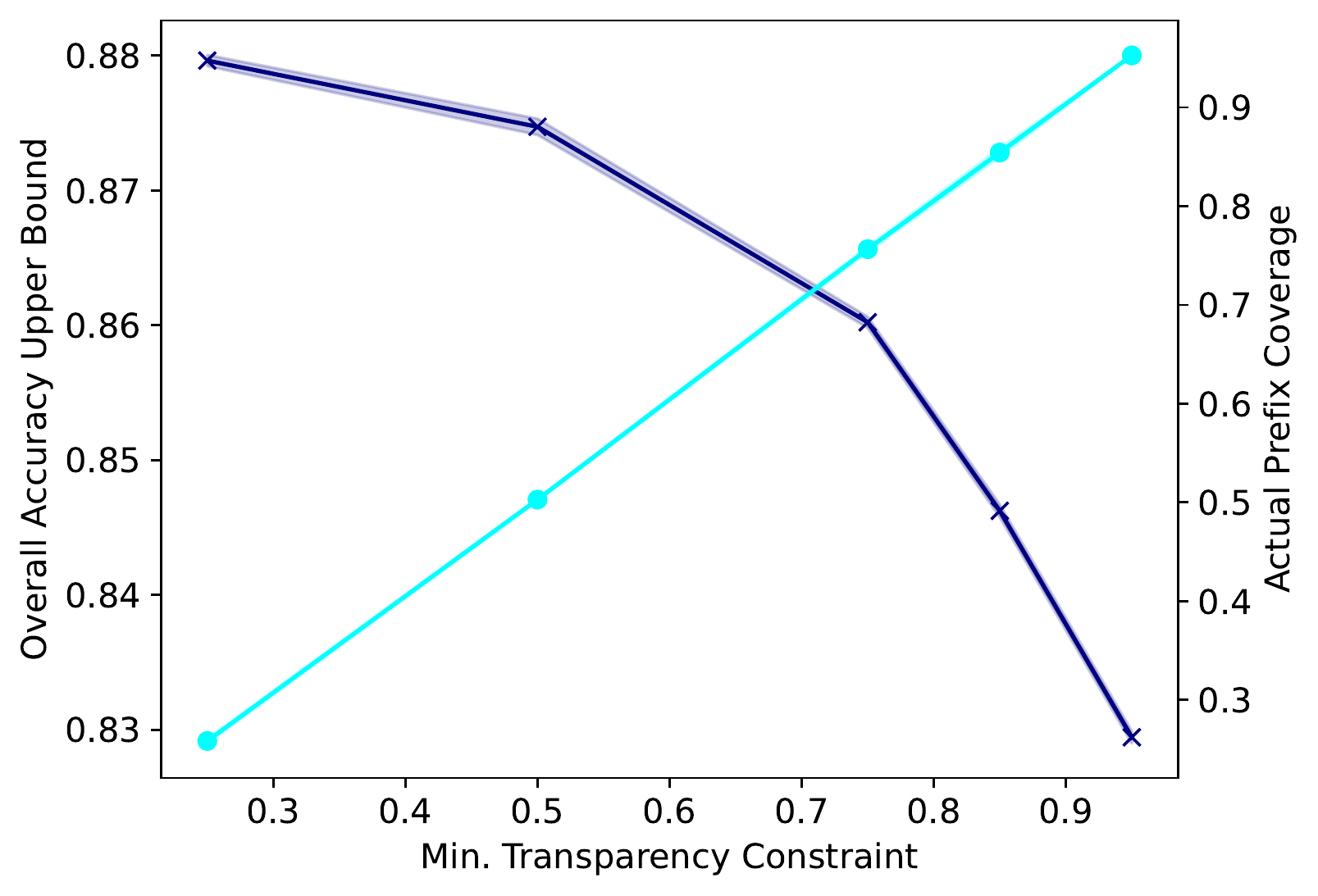}
        \includegraphics[width=0.49\textwidth]{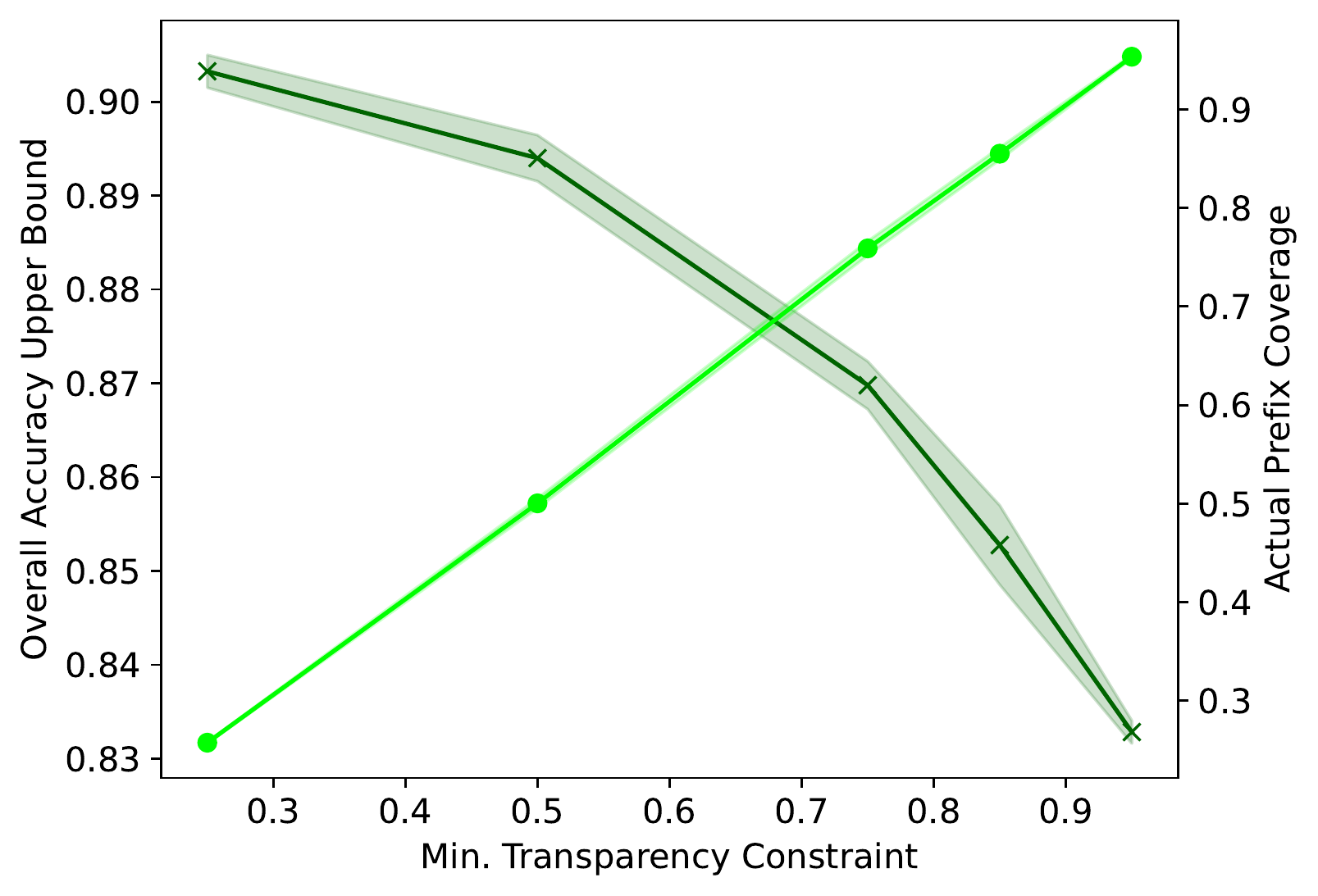}
        \caption{UCI Adult Income dataset.}
    \end{subfigure}
    
    \vskip 10 pt
    
     \begin{subfigure}{\textwidth}
        \centering
        \label{fig:results_best_prefixes_compas}
        \includegraphics[width=0.49\textwidth]{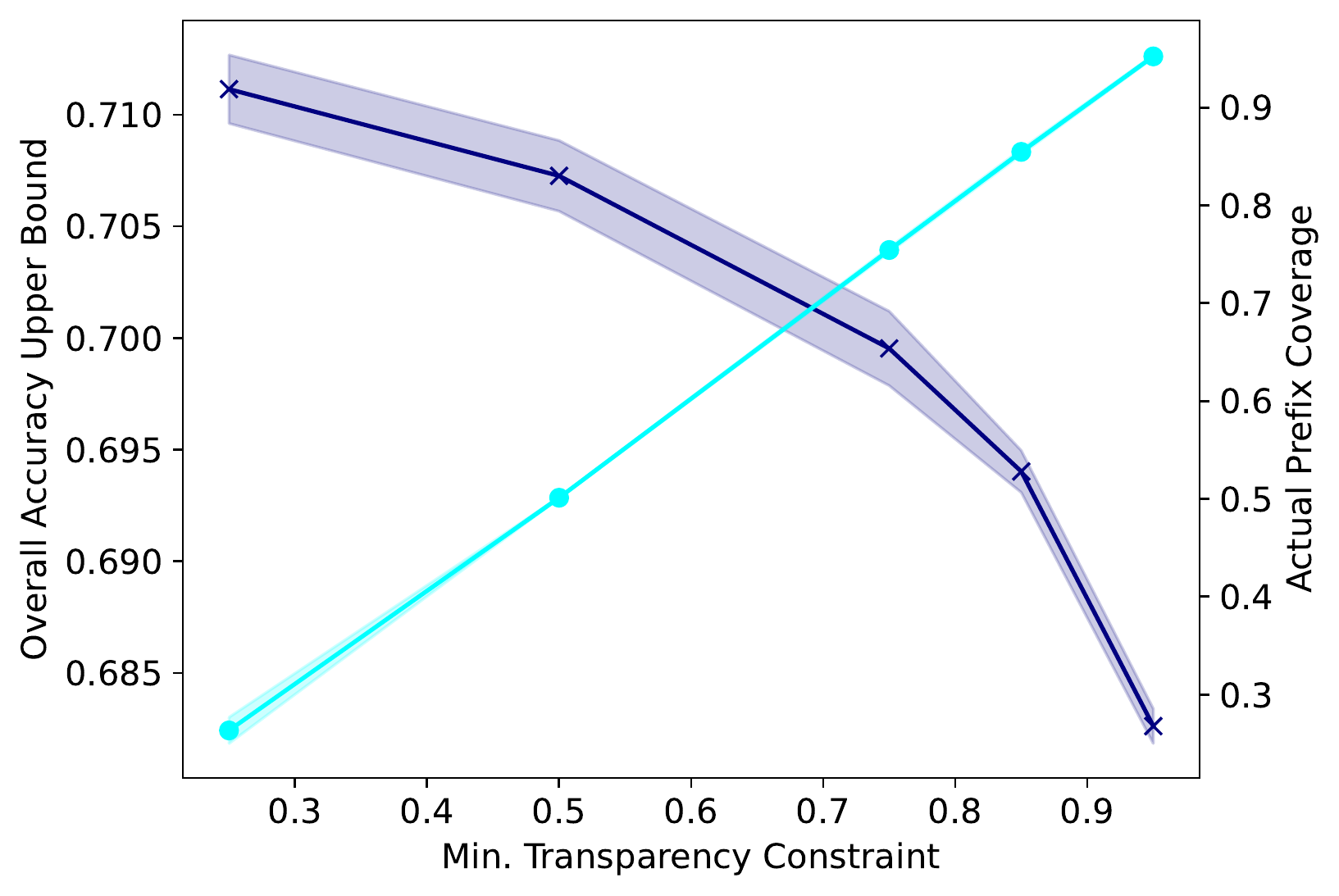}
        \includegraphics[width=0.49\textwidth]{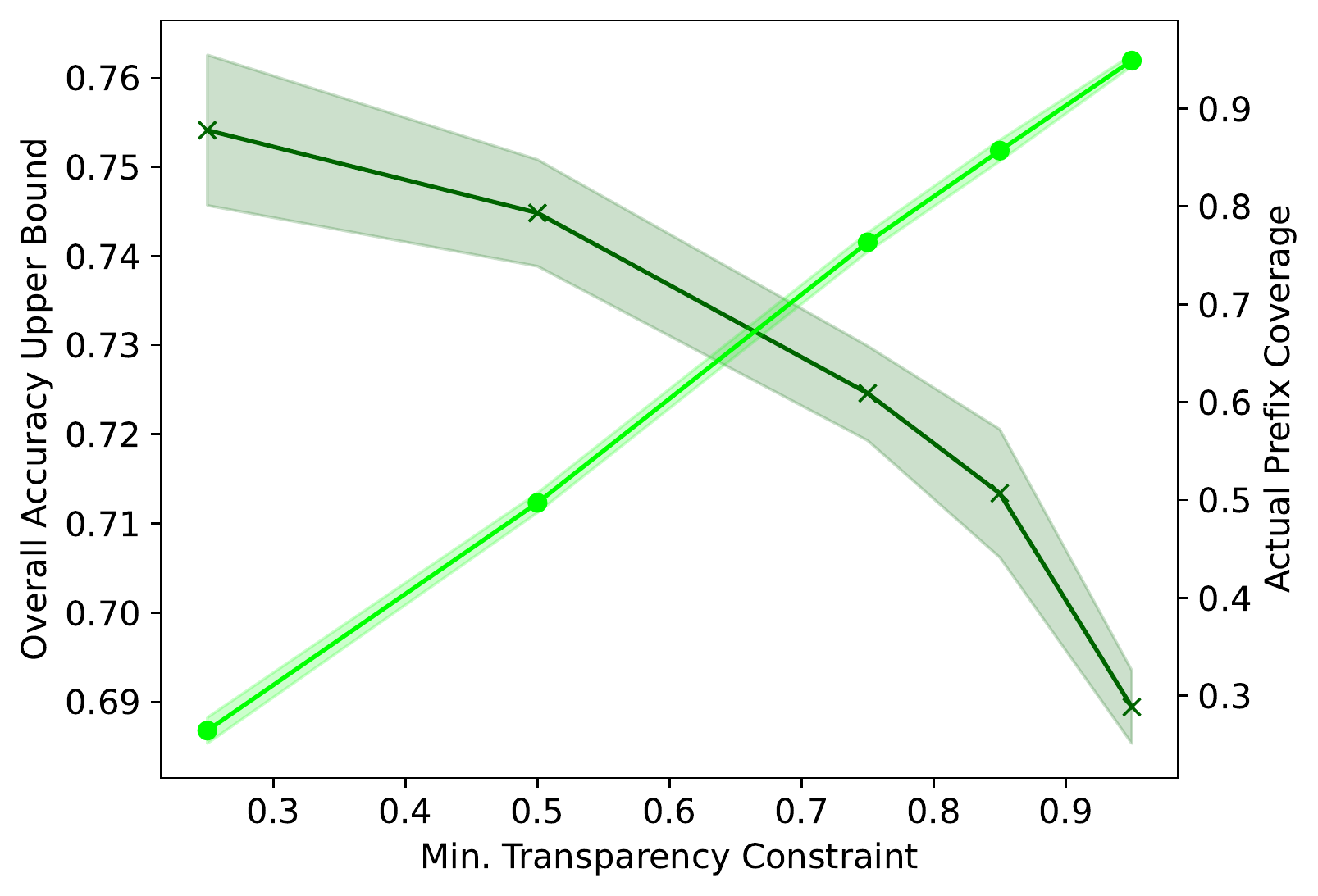}
        \caption{COMPAS dataset.}
    \end{subfigure}
    
     \vskip 10 pt
     
   \includegraphics[width=0.6\textwidth]{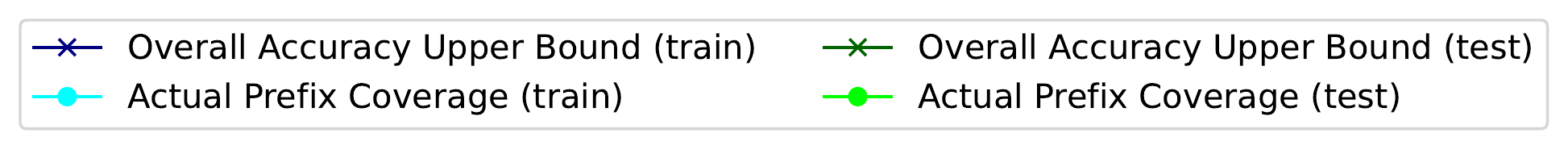}
   
    \caption{Training and test performances of the prefixes learnt using \HybridCORELSPre{}.
    We report the actual transparency (prefix coverage) and overall accuracy upper bound (considering both the prefix's errors and the remaining inconsistencies) - which corresponds to the hybrid model accuracy if the black-box classifies correctly all consistent examples.
    The plots show both average values and standard deviation.} 
    \label{fig:results_best_prefixes}
    
    \end{center}
\end{figure*}

\paragraph{Results} The train and test performances of the learned prefixes are presented in Figure~\ref{fig:results_best_prefixes}. 
As expected, when the enforced transparency level increases, the number of errors made by the interpretable part increases, 
and so does the overall hybrid model error lower bound (computed by the objective function~(\ref{eq:obj_pre})). 
We note that the actual prefix transparency on the training set is very close to the enforced constraint, with very small standard deviations.
This illustrates the conflict between accuracy and transparency. Indeed, if a prefix with very high accuracy and transparency were available, the learning algorithm would systematically select it irrespective of the transparency constraint. However, the fact that transparencies are very close to their enforced constraint 
means that increasing the coverage of the prefix hinders the performance. This empirical observation meets the theoretical discussion of \textbf{Section~\ref{sec:hybridcorels_pre}} (Objective lower bound paragraph).
We also observe that transparency generalizes well: the test set transparency levels are very close to the training set ones. 
Again, the standard deviation across dataset splits is very small.

We report results for the \texttt{AdaBoostClassifier} black-box in Figure~\ref{fig:results_pre_bb_rf_all}
for the three datasets. Results for the two other black-boxes are publicly available on our GitHub repository\footnote{\url{https://github.com/ferryjul/HybridCORELS/tree/master/paper/paper_5.2_results.zip}} and show the same trends.
As expected, higher values of the specialization coefficient $\specializationcoefficient$ lead to higher training accuracy of the black-box part. 
Indeed, the black-box component is evaluated on the subset of the data that is not captured by the interpretable part.
Hence, specializing it on this subset is expected to raise its performances \emph{on these samples}.
Note that small variations exist, which can be explained by the heuristic nature of the considered black-box training algorithms.
Overall, a \emph{reasonable} specialization is usually beneficial. 
For low transparency values, the improvements brought by specialization are relatively modest (check the y-axis scales). This is explained by the fact that, in such contexts, the black-box subset of the data already represents most of the dataset.
For very high transparency values, the black-box subset is relatively small, and an excessive specialization may not always pay off due to overfitting (as is the case with the UCI Adult Income experiment). 
For medium to high transparency values, specialization (with carefully chosen specialization coefficient $\specializationcoefficient$) is always beneficial in these experiments.
Here, specialization allows for black-box test accuracy absolute improvements up to $2.27$ pps (experiment using the ACS Employment dataset, with minimum transparency $0.95$).
Considering all the experiments run with the \texttt{AdaBoostClassifier} black-box, the improvement rate (proportion of experiments for which specialization allowed improvements of the black-box test accuracy) is the highest for $\specializationcoefficient=2$, with a $93.33$\% improvement rate.
Considering all the run experiments (including runs for the three datasets and the different transparency levels), the improvement rate values are the highest for $\specializationcoefficient \in \{1,2\}$. This confirms the usefulness of specialization but highlights the need to use reasonable specialization coefficient values $\specializationcoefficient$.
Observe that, when $\specializationcoefficient=1$, misclassifying an example belonging to the (training) black-box subset costs $\frac{e^1}{e^0} \approx 2.72$ times more than misclassifying a training example outside this set (in the optimized loss function). 
When  $\specializationcoefficient=2$, it costs $\frac{e^2}{e^0} \approx 7.39$ times more.

\begin{figure}
    \centering
    
    \begin{subfigure}{\textwidth}
        \centering
        
        \includegraphics[width=0.49\textwidth]{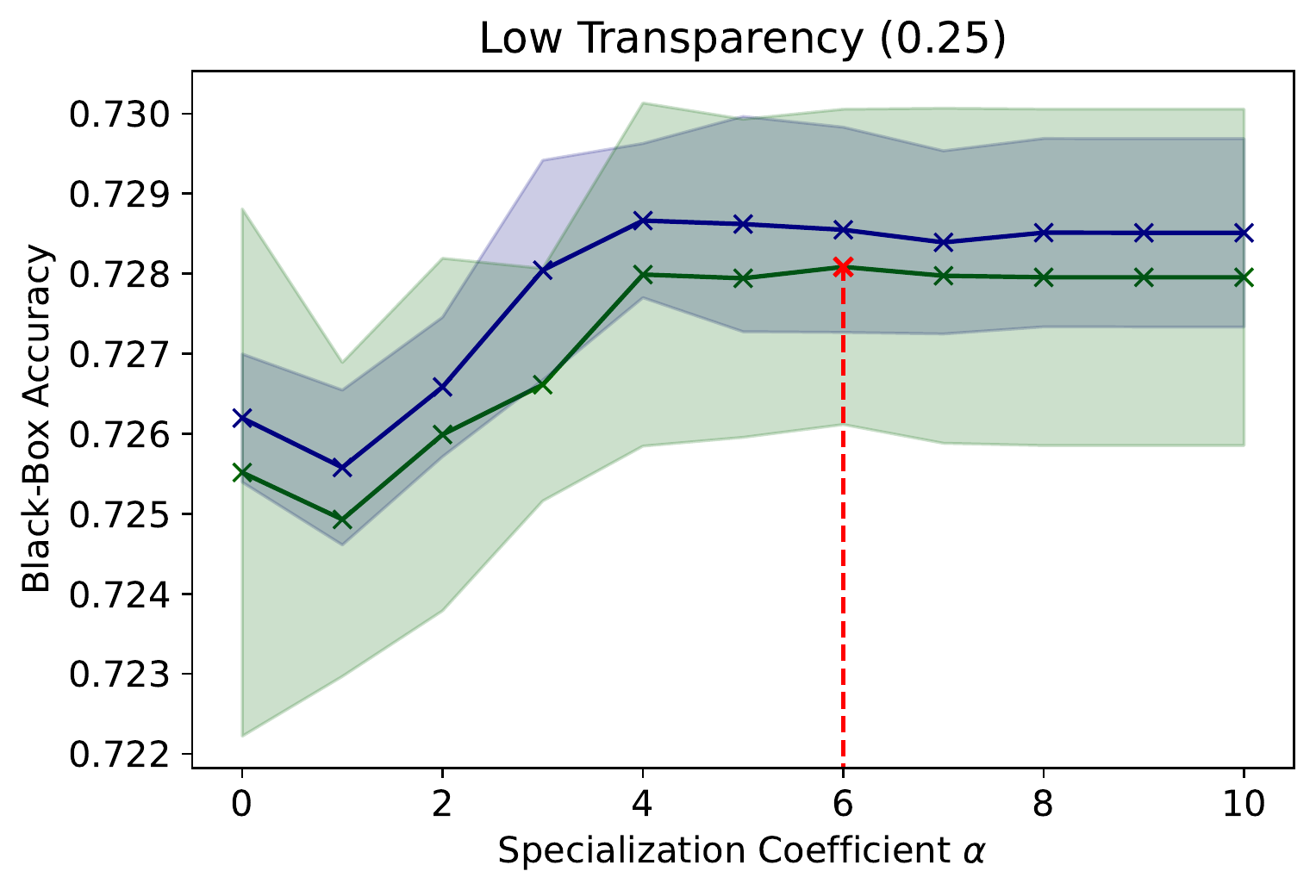}
        \includegraphics[width=0.49\textwidth]{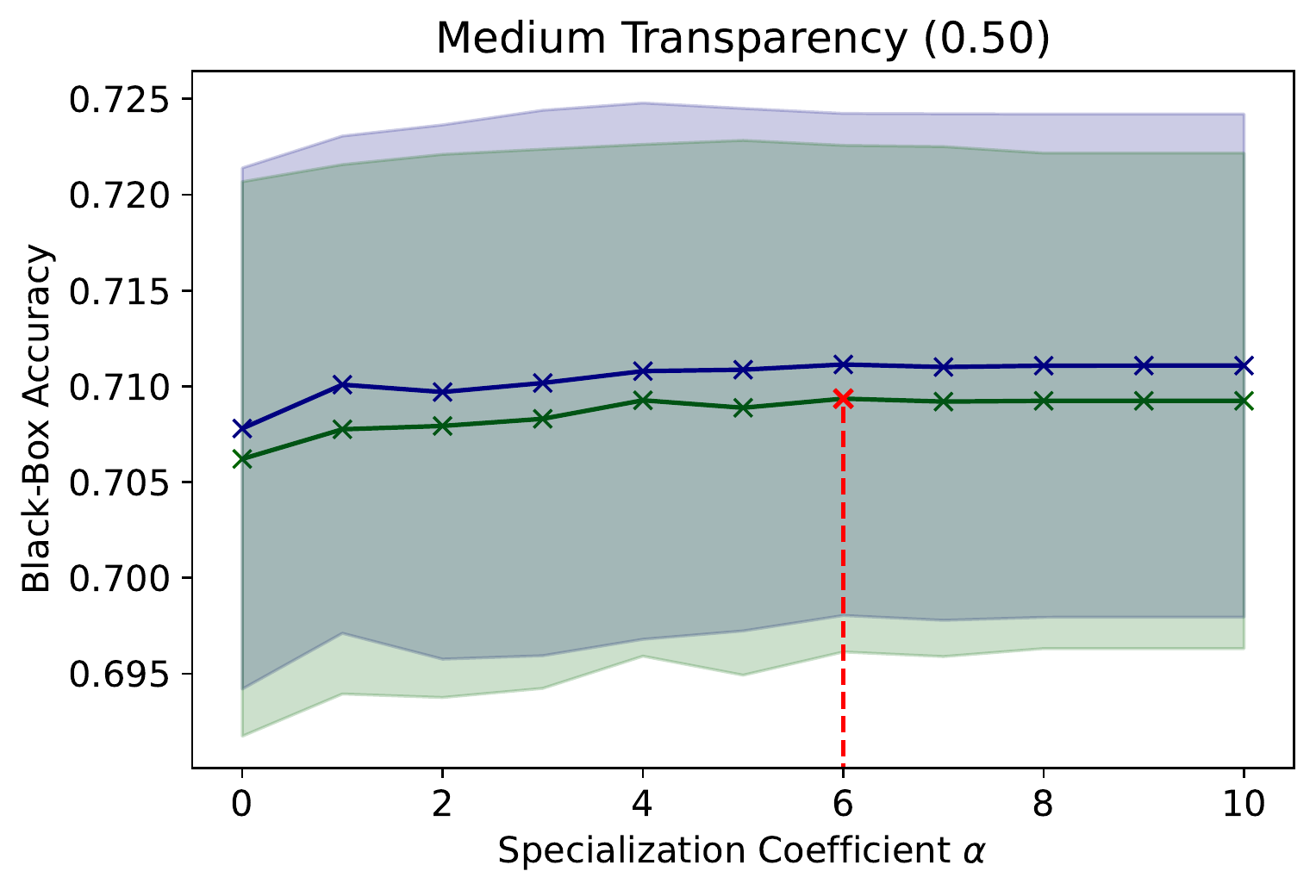}
        \includegraphics[width=0.49\textwidth]{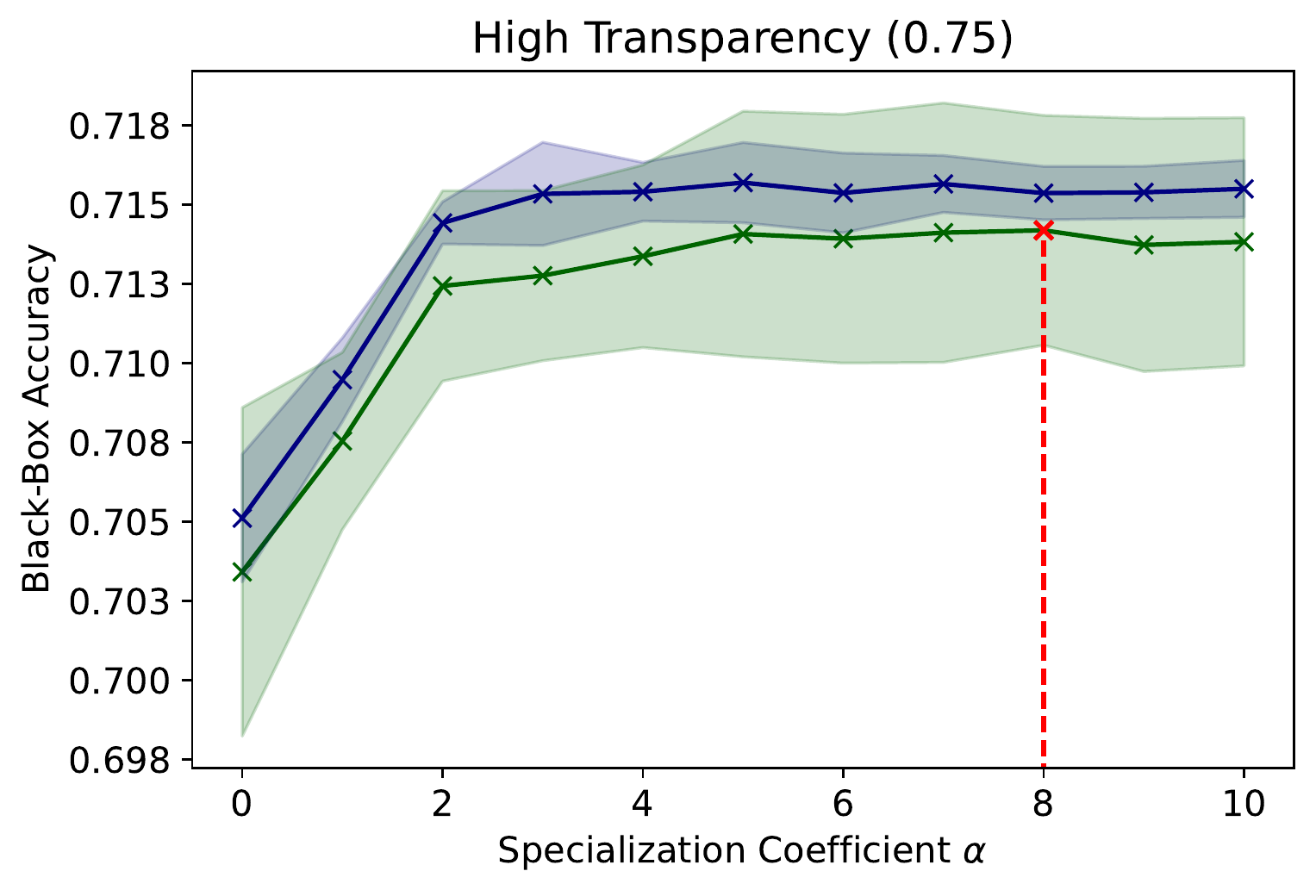}
        \includegraphics[width=0.49\textwidth]{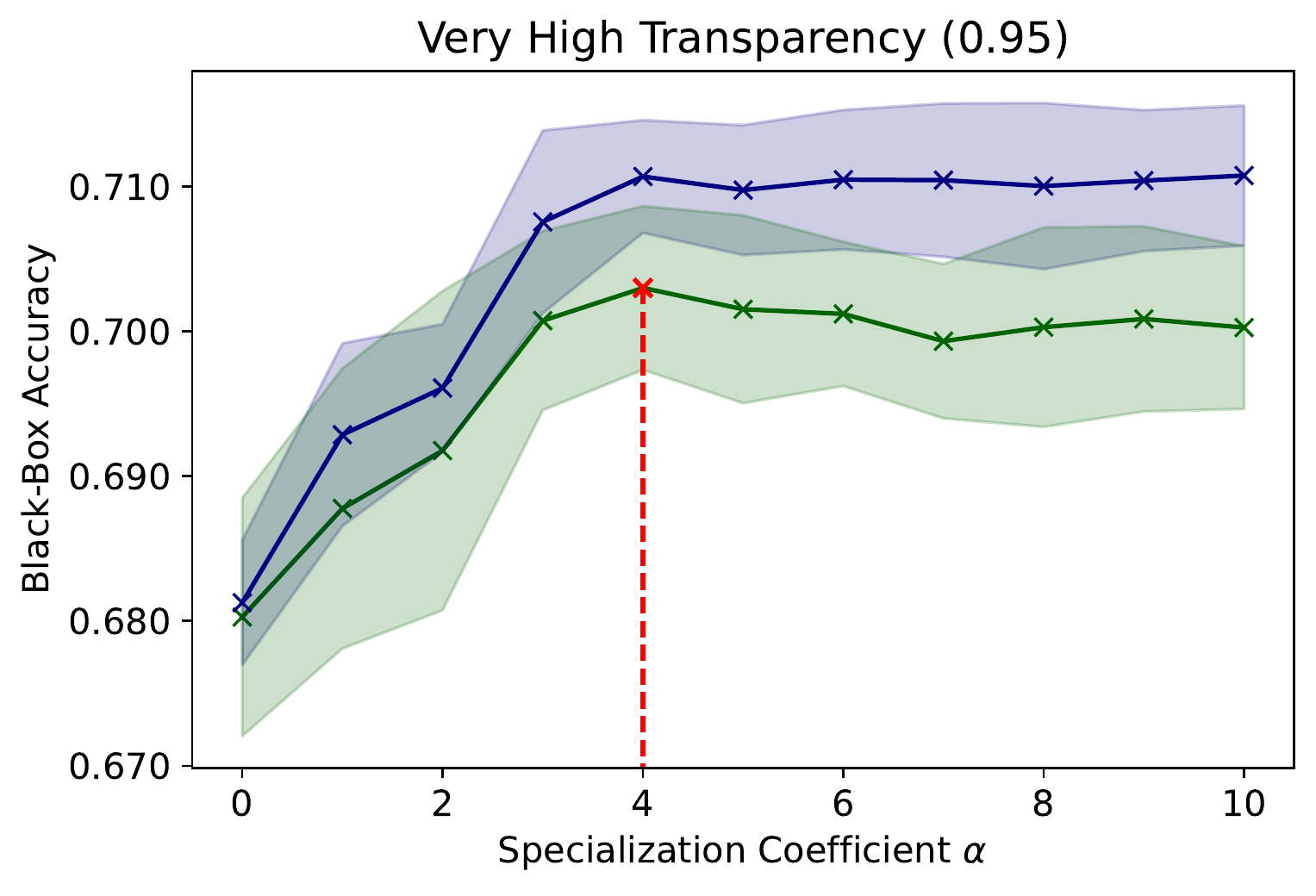}
      
        \caption{ACS Employment dataset\label{fig:results_pre_bb_rf_acs_employ}.}
        
    \end{subfigure}
    
    \par\bigskip 
    
    \begin{subfigure}{\textwidth}
        \centering
        
        \includegraphics[width=0.49\textwidth]{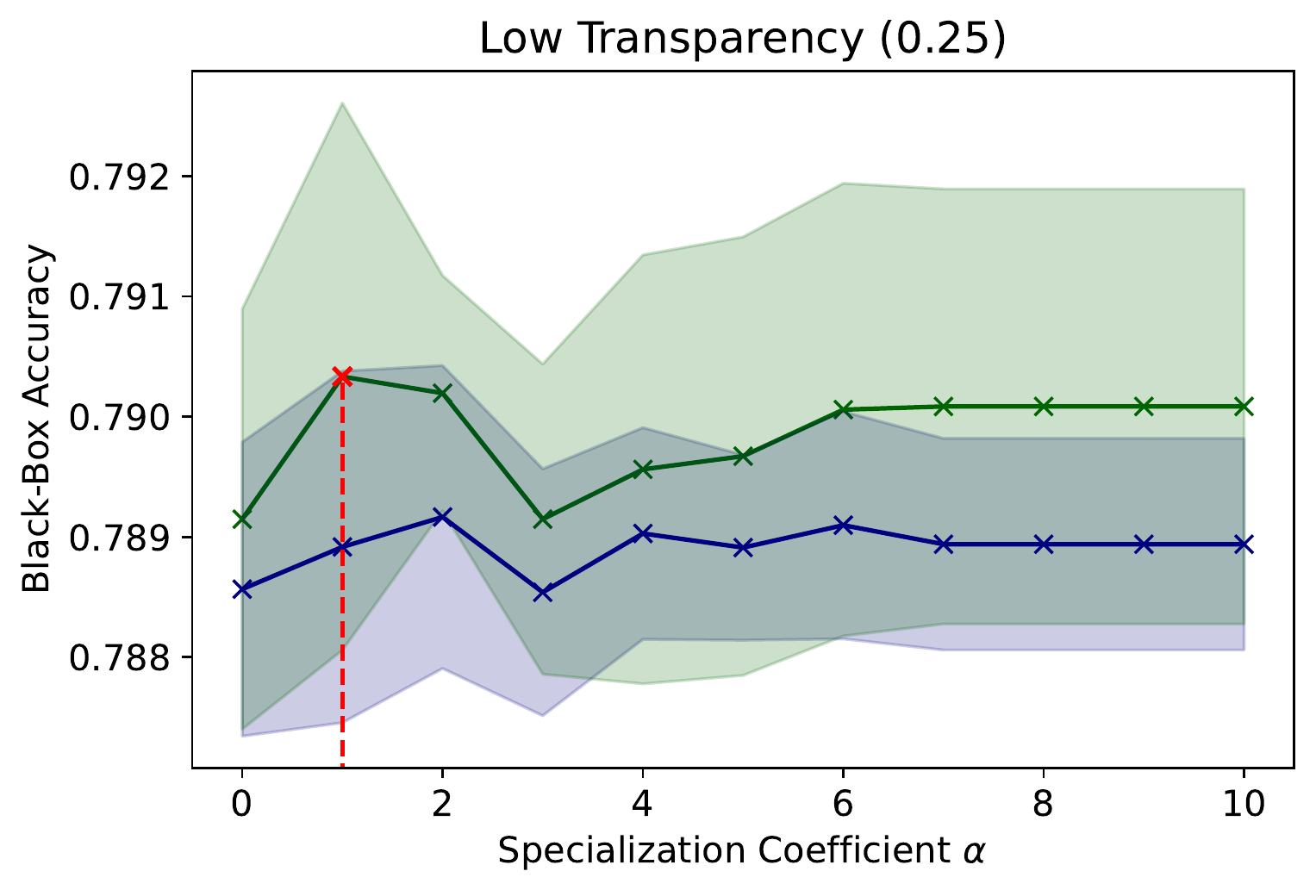}
        \includegraphics[width=0.49\textwidth]{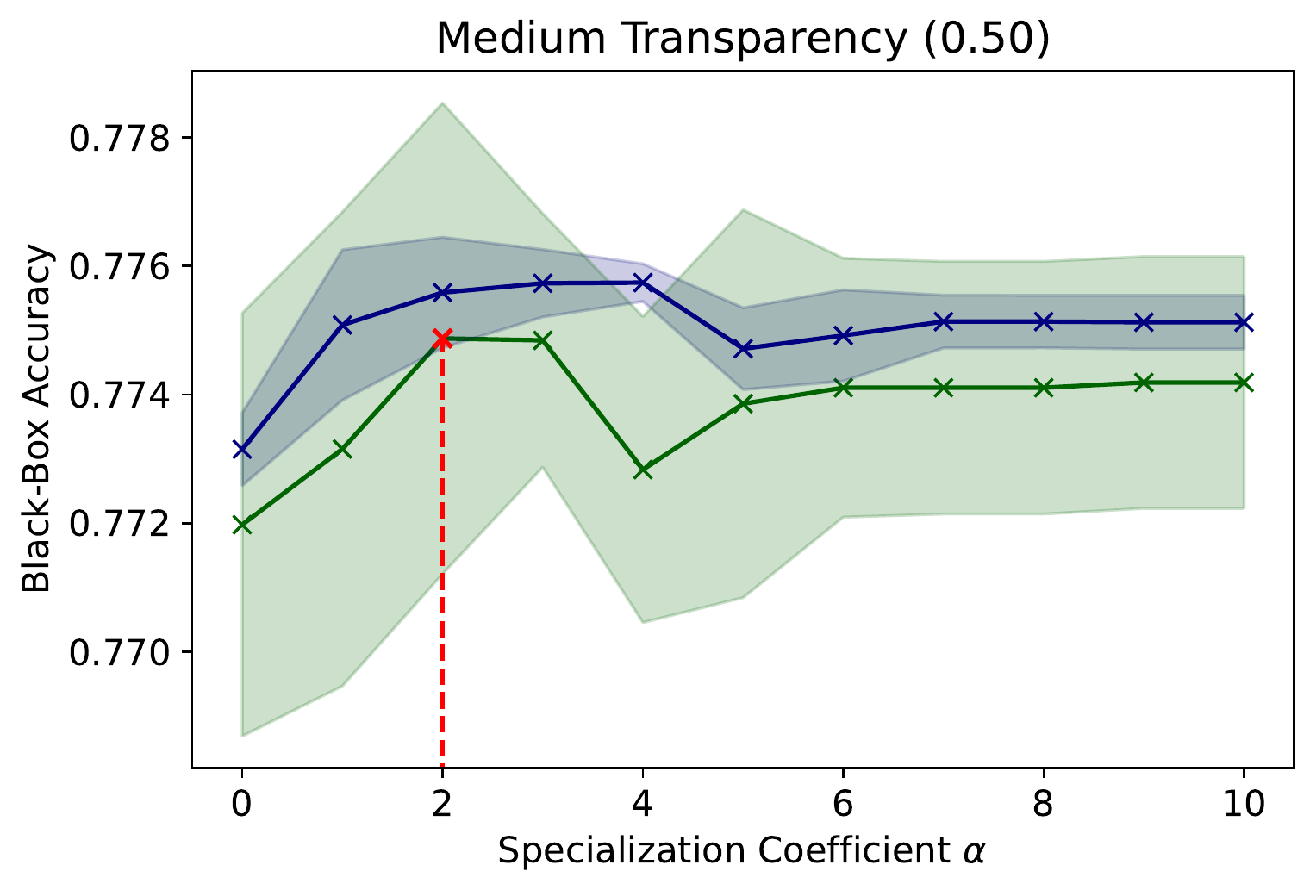}
        \includegraphics[width=0.49\textwidth]{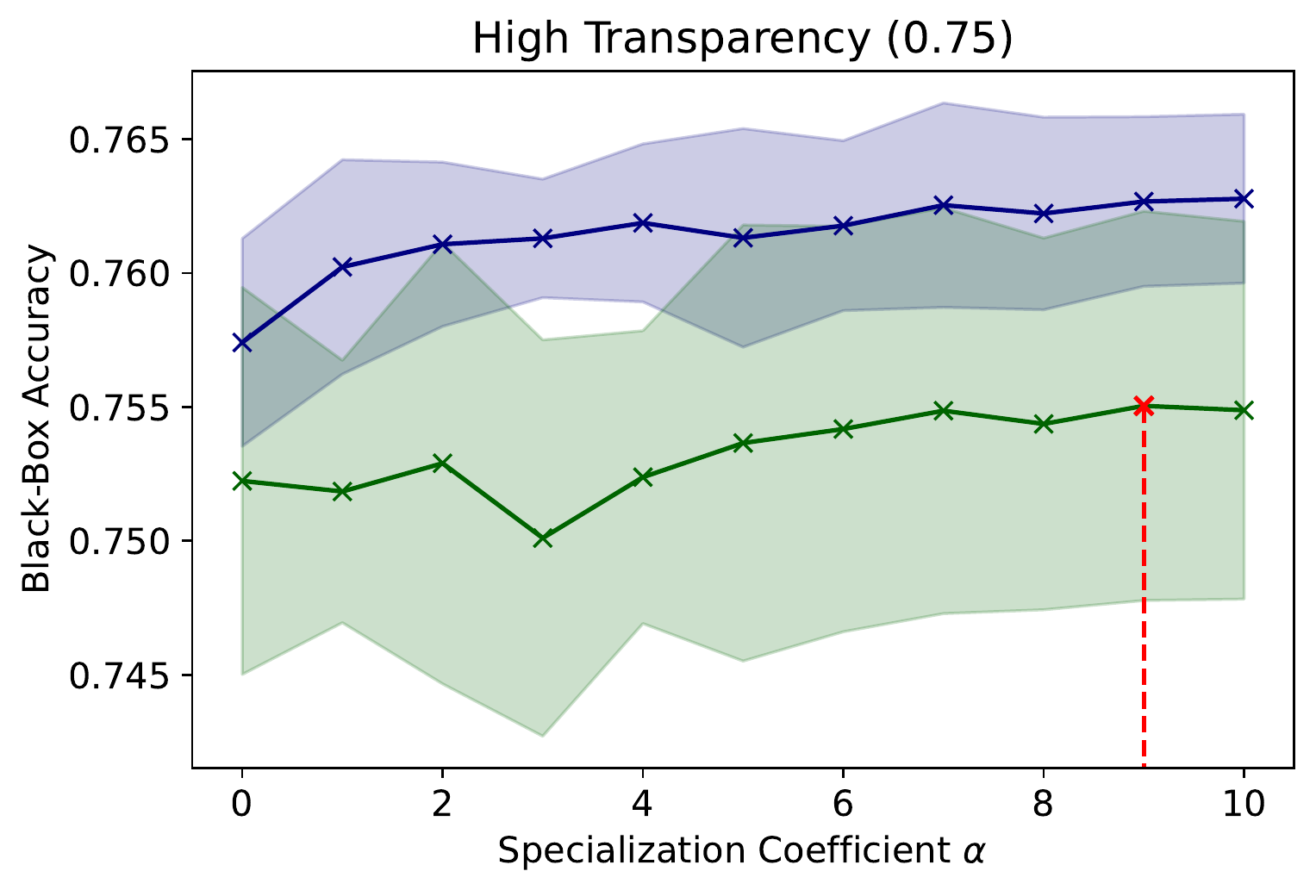}
        \includegraphics[width=0.49\textwidth]{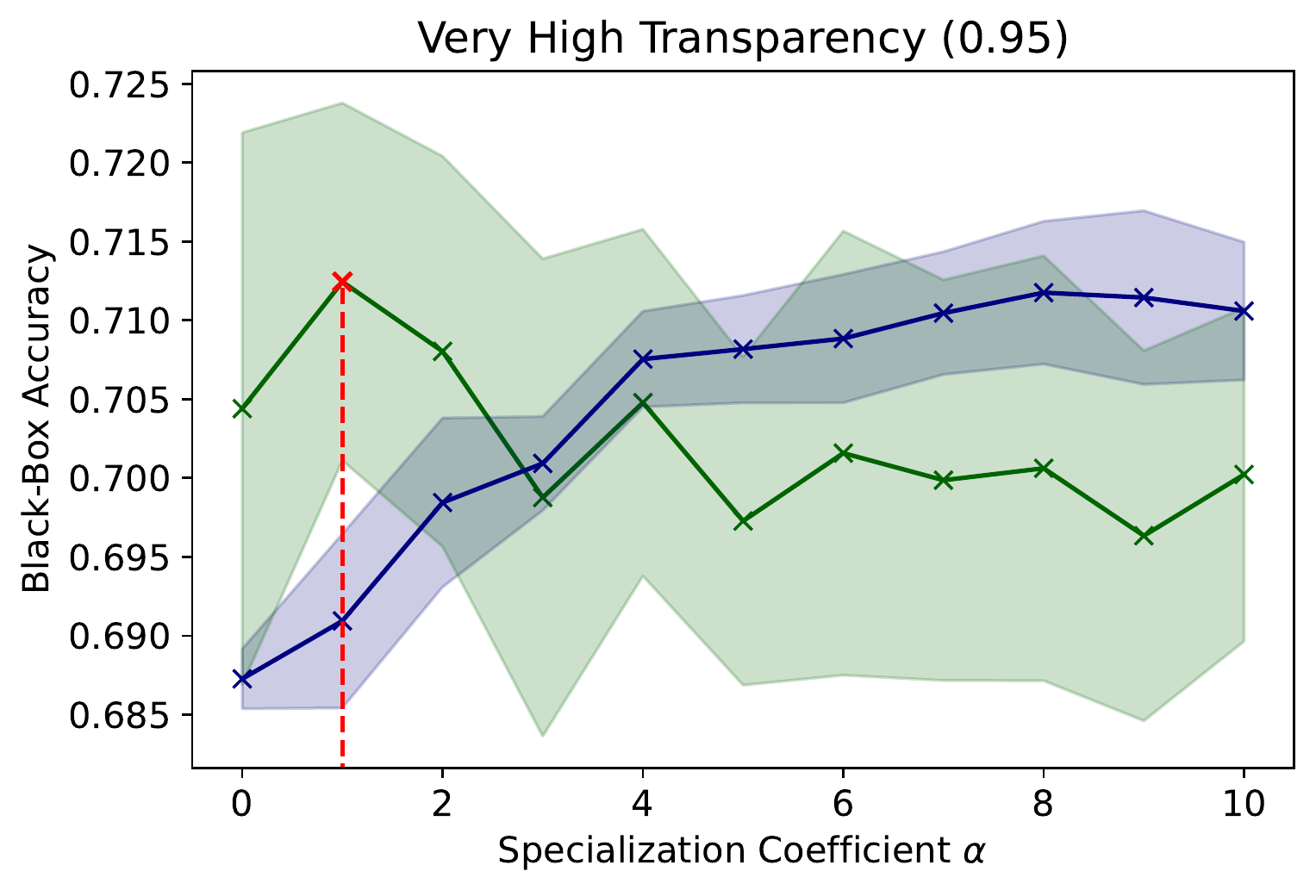}

        \caption{UCI Adult Income dataset\label{fig:results_pre_bb_rf_adult}.}
    
    \end{subfigure}
\end{figure}
\begin{figure}[t!] \ContinuedFloat
    \centering
    \begin{subfigure}{\textwidth}
        \centering

        \includegraphics[width=0.49\textwidth]{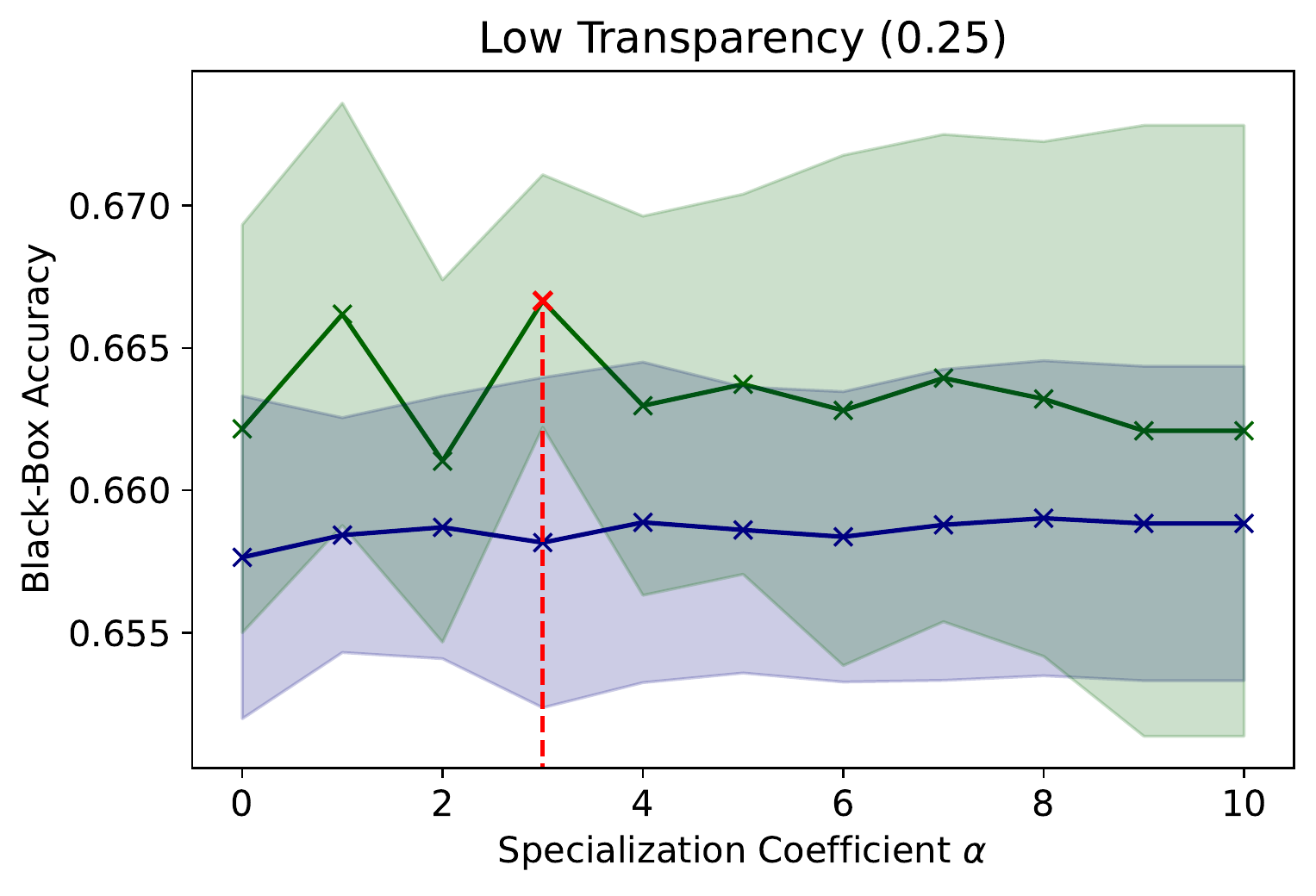}
        \includegraphics[width=0.49\textwidth]{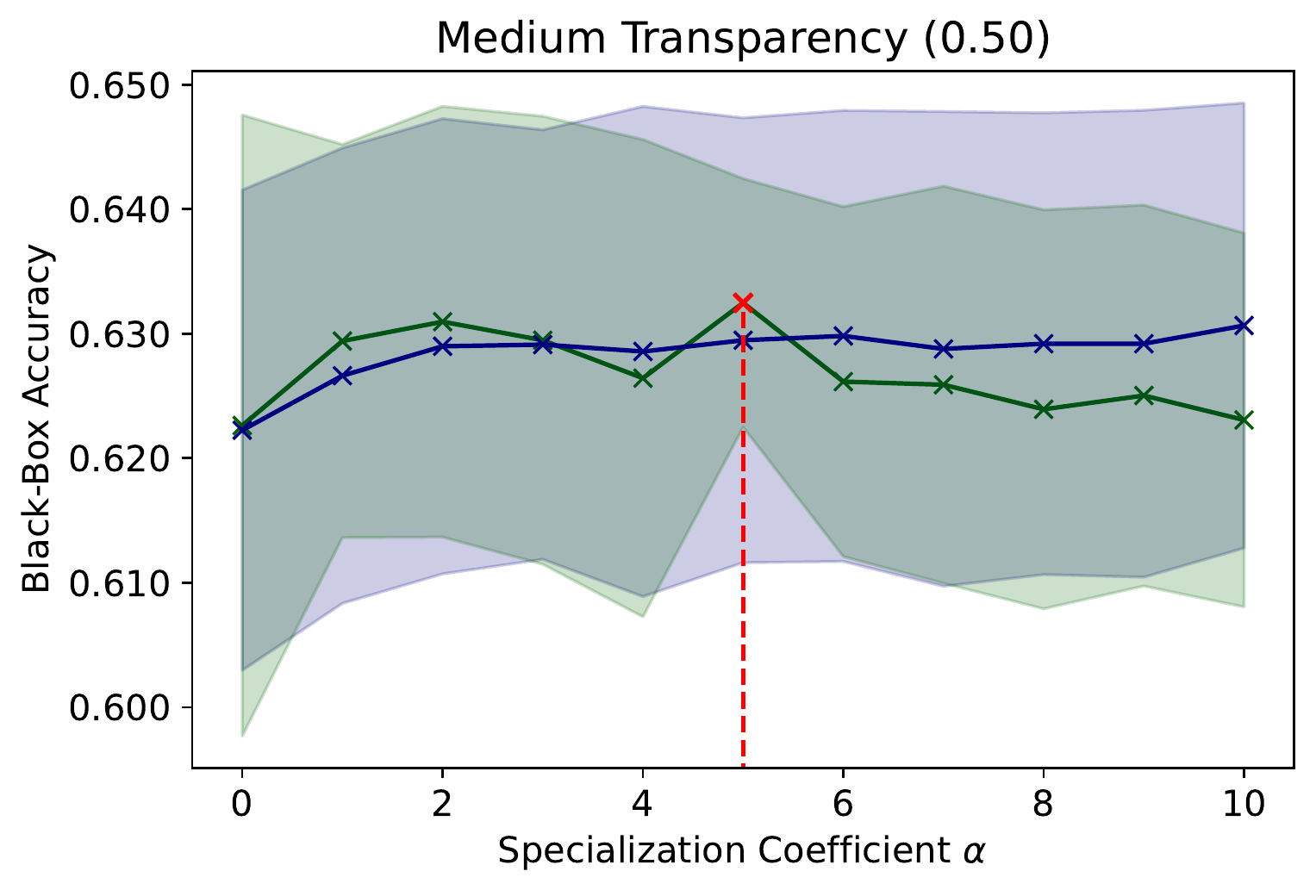}
        \includegraphics[width=0.49\textwidth]{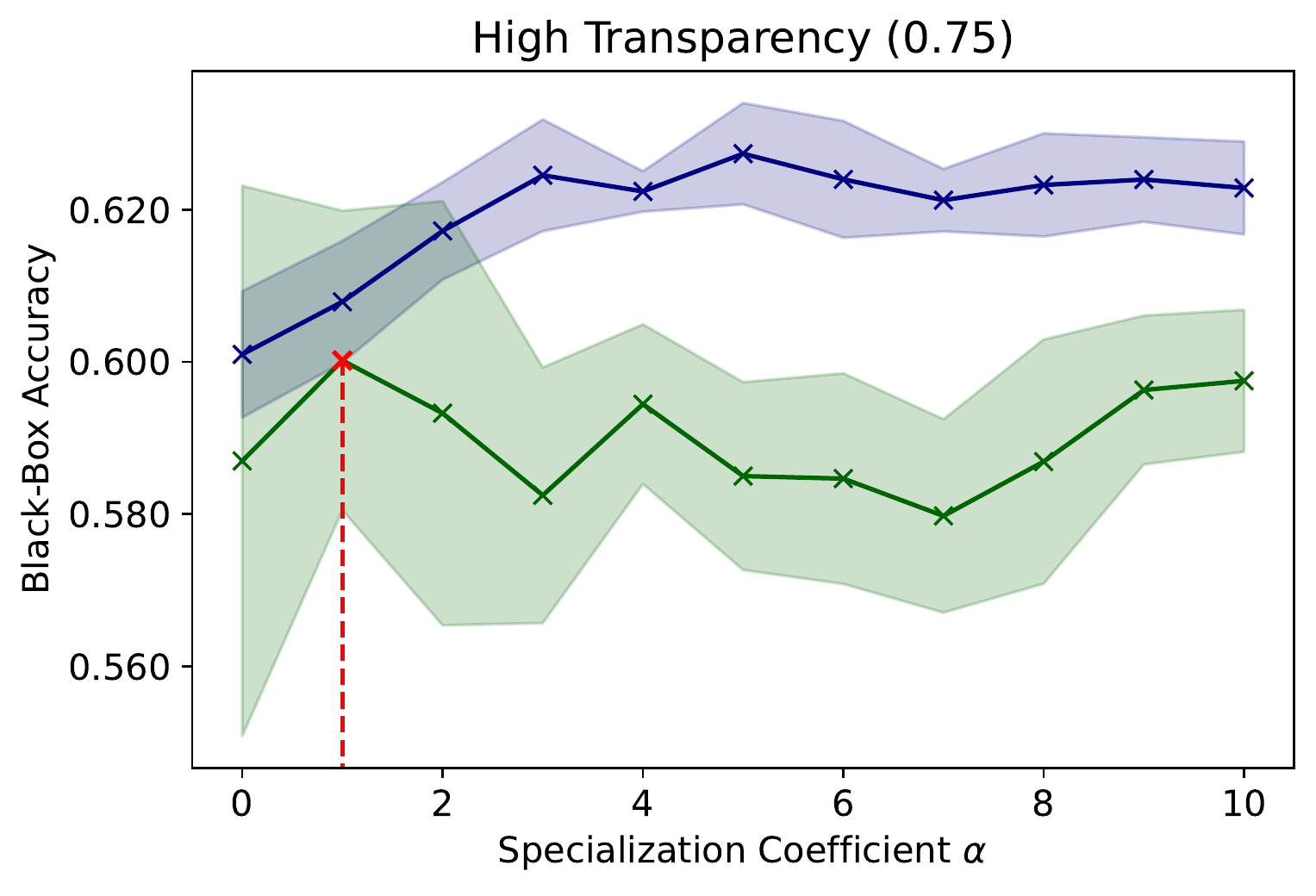}
        \includegraphics[width=0.49\textwidth]{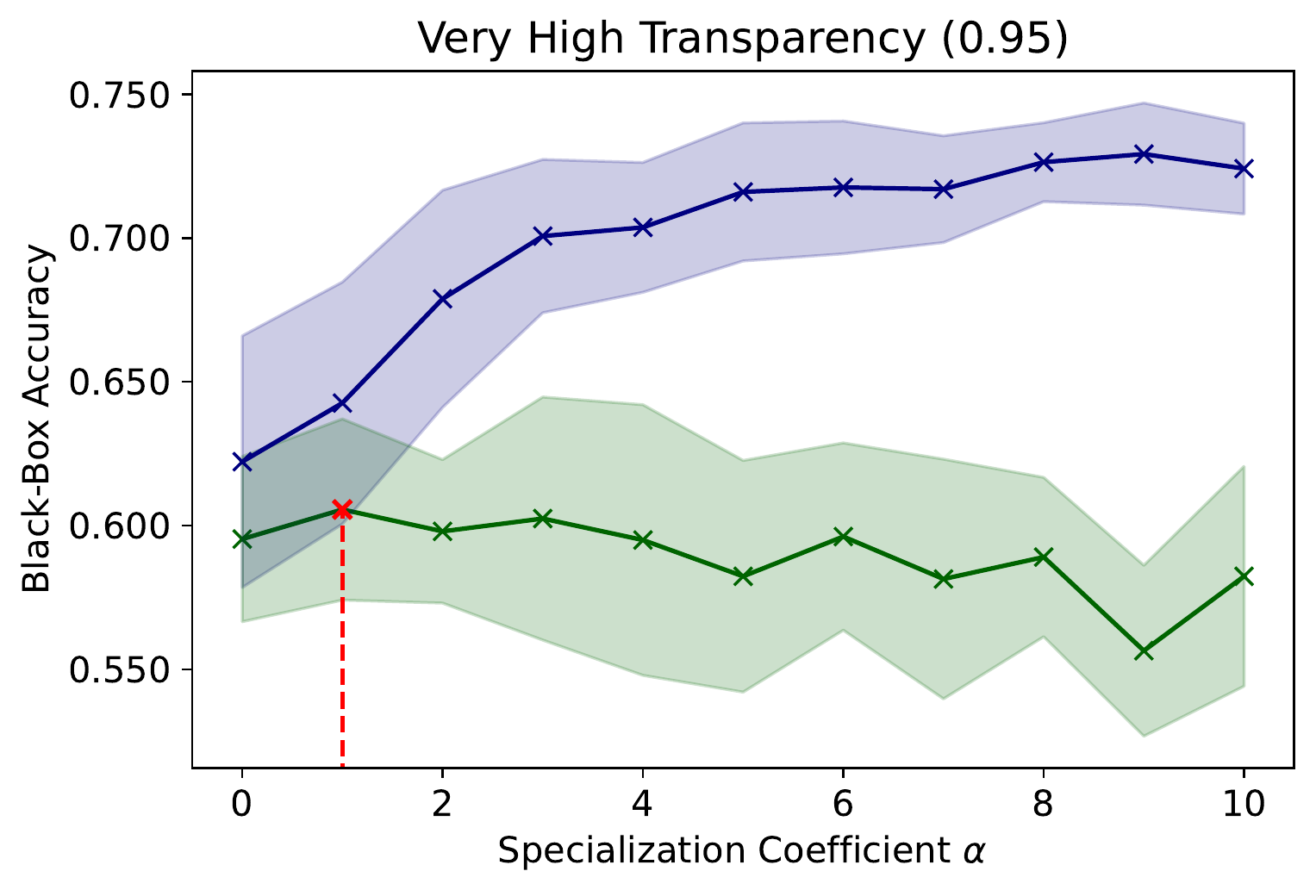}
        
        \caption{COMPAS dataset\label{fig:results_pre_bb_rf_compas}.}

    \end{subfigure}

    \vskip 5 pt
    
    \includegraphics[width=0.75\textwidth]{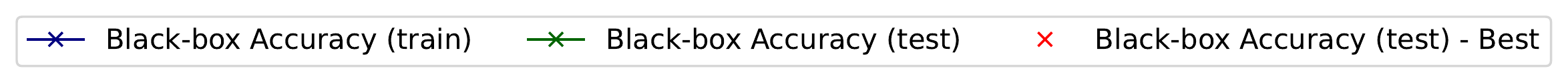}
      
    \caption{Training and test performances of the black-box parts (\texttt{AdaBoostClassifier}) of the hybrid interpretable models learnt using \HybridCORELSPre{} on different datasets, for different transparency levels. The plots show both average values and standard deviation.}    
    \label{fig:results_pre_bb_rf_all}
\end{figure}
\newpage

\subsection{Tradeoffs and Comparison with the State-of-the-Art}
\label{sec:expes_all}

\paragraph{Objective} The aim of this subsection is to explore the trade-offs between the accuracy and transparency of several hybrid interpretable models learning frameworks: the state-of-the-art \hyrs{} and \crl{} methods, as well as our proposed \HybridCORELSPost{} and \HybridCORELSPre{} algorithms. 
These experiments serve the dual purpose of quantitatively comparing the various methods, but also to advertise the considerable amounts of transparency that can be attained while maintaining high performance.

\paragraph{Setup} For these experiments, each dataset was split into training (60\%), validation (20\%), and test (20\%) sets.
We randomly generate five such splits and average the results over them.
More precisely, for each split, the training set is used to train the models (both the black-box and the interpretable parts). 
The models' hyperparameters are optimized using the (separate) validation set.
Finally, the resulting hybrid models are evaluated on the (separate) test set.
Hereafter, we detail the training and hyper-parameters optimization procedures for both the black-boxes and 
the hybrid interpretable models themselves. 


\paragraph{\prebb{} method setup} 
The experiments using the \HybridCORELSPre{} algorithm are divided into two phases. 
First, for each dataset (out of $3$) and each random split (out of $5$), we learn prefixes for $12$ different minimum transparency constraints ($0.1$, $0.2$, $0.3$, $0.4$, $0.5$, $0.6$, $0.7$, $0.8$, $0.9$, $0.925$, $0.95$, $0.975$) trying the following hyperparameters values: $\corelsreg \in \{10^{-2}, 10^{-3}, 10^{-4}\}$, $\corelsminsupport \in \{0.01, 0.05, 0.10\}$, and the \emph{objective-guided}, \emph{lower-bound-guided}, and \emph{BFS} search policies for \HybridCORELSPre{}. 
Each prefix learning is limited to a maximum CPU time of $1$ hour and a maximum memory use of $8$ GB.
For each experiment (dataset - random split - minimum transparency), the prefix yielding the best validation accuracy is retained.
In a second phase, for each retained prefix, we try three different \sklearn{}~\citep{scikit-learn} black-boxes: a \texttt{RandomForestClassifier}, an \texttt{AdaBoostClassifier}, and a \texttt{GradientBoostingClassifier}.
The black-box hyperparameters are tuned using the \hyperopt{}~\citep{bergstra2013making} Python library and its Tree of Parzen Estimators (TPE) algorithm, with $100$ iterations. 
Just like the prefixes in the first phase, the black-boxes are trained using the training split ($60$\%) and the hyperparameters are selected based on the validation split ($20$\%) performances. 
Note that, as for the training set, the validation set loss is weighted to encourage the black-box to accurately classify the examples belonging to its assigned part of the input space (which is fixed as the prefix was trained first - \emph{which allows specialization}, as previously discussed).
Based on the observations from \textbf{Section~\ref{sec:expes_pre}}, we set the specialization coefficient $\specializationcoefficient = 1$, which corresponds to a moderate black-box specialization.

\paragraph{\postbb{} methods setup}
Three methods correspond to the \postbb{} paradigm: \HybridCORELSPost{}, along with the two state-of-the-art \hyrs{}~\citep{wang2019gaining} and \crl{}~\citep{pan2020interpretable} methods.
The experiments using these methods are divided into two phases.
First, for each dataset (out of $3$) and each random split (out of $5$), we train three different \sklearn{}~\citep{scikit-learn} black-boxes: a \texttt{RandomForestClassifier}, an \texttt{AdaBoostClassifier}, and a \texttt{GradientBoostingClassifier}.
The black-box hyperparameters are tuned using the \hyperopt{}~\citep{bergstra2013making} Python library and its Tree of Parzen Estimators (TPE) algorithm, with $100$ iterations. The black-boxes are trained using the training split ($60$\%) and their hyperparameters are selected based on the validation split ($20$\%) performances.
In the second phase of the experiments, we train the interpretable parts of the hybrid models for the three compared methods, using the black-boxes learned in the previous phase. For each of the three methods, we try different hyperparameter values. Again, the training is performed on the training split ($60$\%), while the hyperparameters values are selected based on the validation split ($20$\%) performances. For \HybridCORELSPost{}, we consider $12$ different minimum transparency constraints ($0.1$, $0.2$, $0.3$, $0.4$, $0.5$, $0.6$, $0.7$, $0.8$, $0.9$, $0.925$, $0.95$, $0.975$), and the following hyperparameters values: $\corelsreg \in \{10^{-2}, 10^{-3}, 10^{-4}\}$, $\corelsminsupport \in \{0.01, 0.05, 0.10\}$, and the \emph{objective-guided}, \emph{lower-bound-guided}, and \emph{BFS} search policies. For the \hyrs{} method, similarily to what was done in \citep{wang2019gaining}, we use $10$ different values for its $\lambda$ hyperparameter (ranging logarithmically between $10^{-3}$ and $10^{-2}$) and $10$ different values for its $\beta$ hyperparameter (ranging logarithmically between $10^{-3}$ and $10^{0}$). For \crl{}, we consider $10$ different values for its \emph{temperature} hyperparameter (ranging linearly between $10^{-3}$ and $10^{-2}$) and $10$ different values for its $\lambda$ hyperparameter (ranging logarithmically between $10^{-3}$ and $10^{-1}$). For all three methods \HybridCORELSPost{}, \hyrs{}, and \crl{}, the hyperparameter grid is roughly of size 100. 
As in the \HybridCORELSPre{} experiments, the interpretable parts building is limited to a maximum CPU time of $1$ hour and a maximum memory use of $8$ GB.

\paragraph{Final results computation} After tuning the hyper-parameters, we are left with a Pareto front representing the hybrid models that are not dominated in terms of both validations set accuracy and transparency. Still, since the black box and hybrid models were fine-tuned on the validation set, we argue that this Pareto front will be an over-optimistic description of the true generalisation of our hybrid models. For this reason, we decided to take the Pareto-optimal models on validation, and compute their accuracy and transparency on the test set, which has not been used yet in this experiment. Hence, we can obtain unbiased measures of the accuracy and transparency for these models. These final measures of accuracy/transparency are used as a means to compare the different approaches and assess if increasing transparency can lead to equivalent/better generalization.

\paragraph{Results}

\def\figwidth{0.326}
\begin{figure*}
    \begin{center}
        
    

    \begin{subfigure}{\textwidth}
        \centering
        \label{fig:results_tradoffs_acs_employ_test}
        \includegraphics[width=\figwidth\textwidth]
         {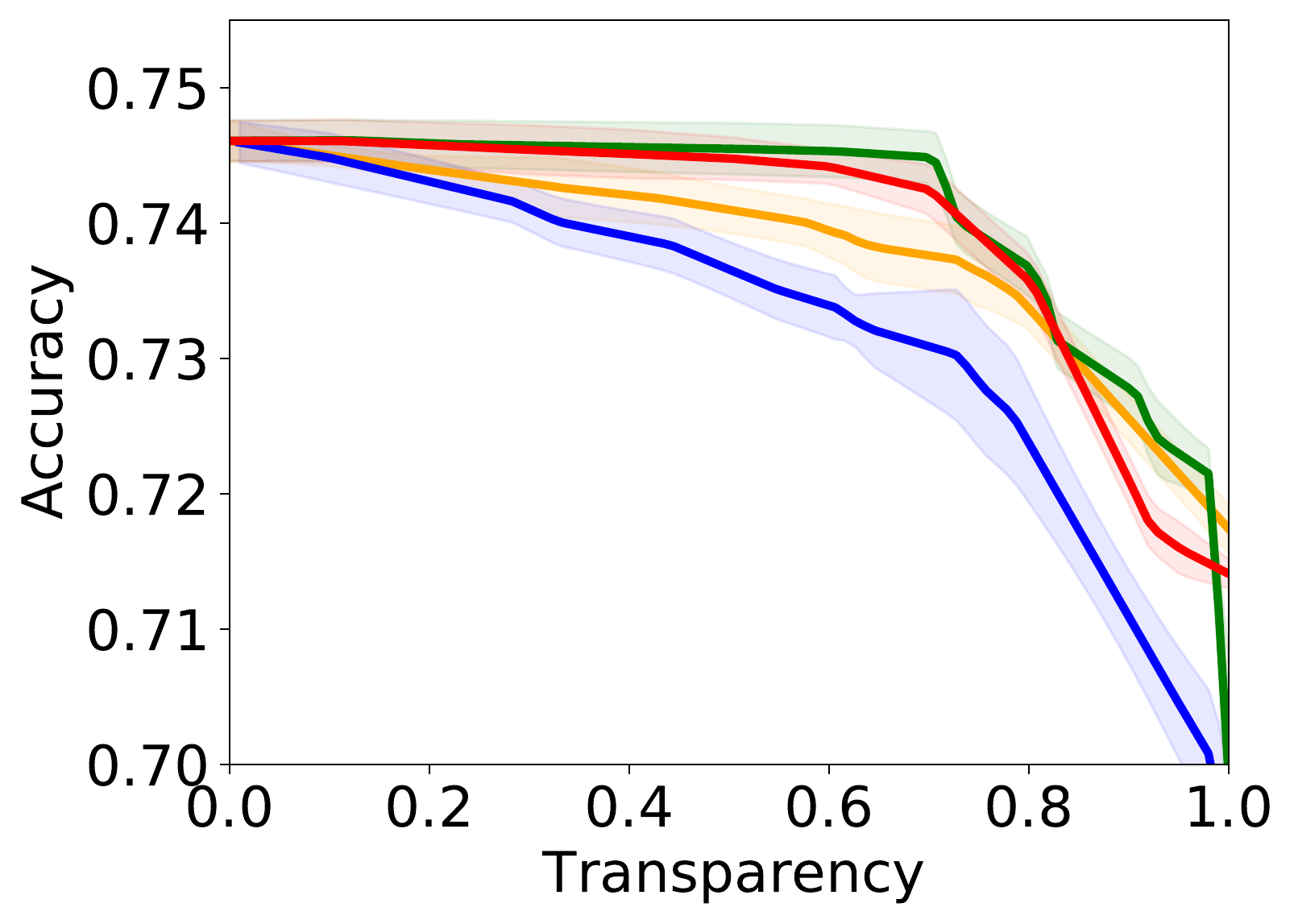}
         \includegraphics[width=\figwidth\textwidth]
         {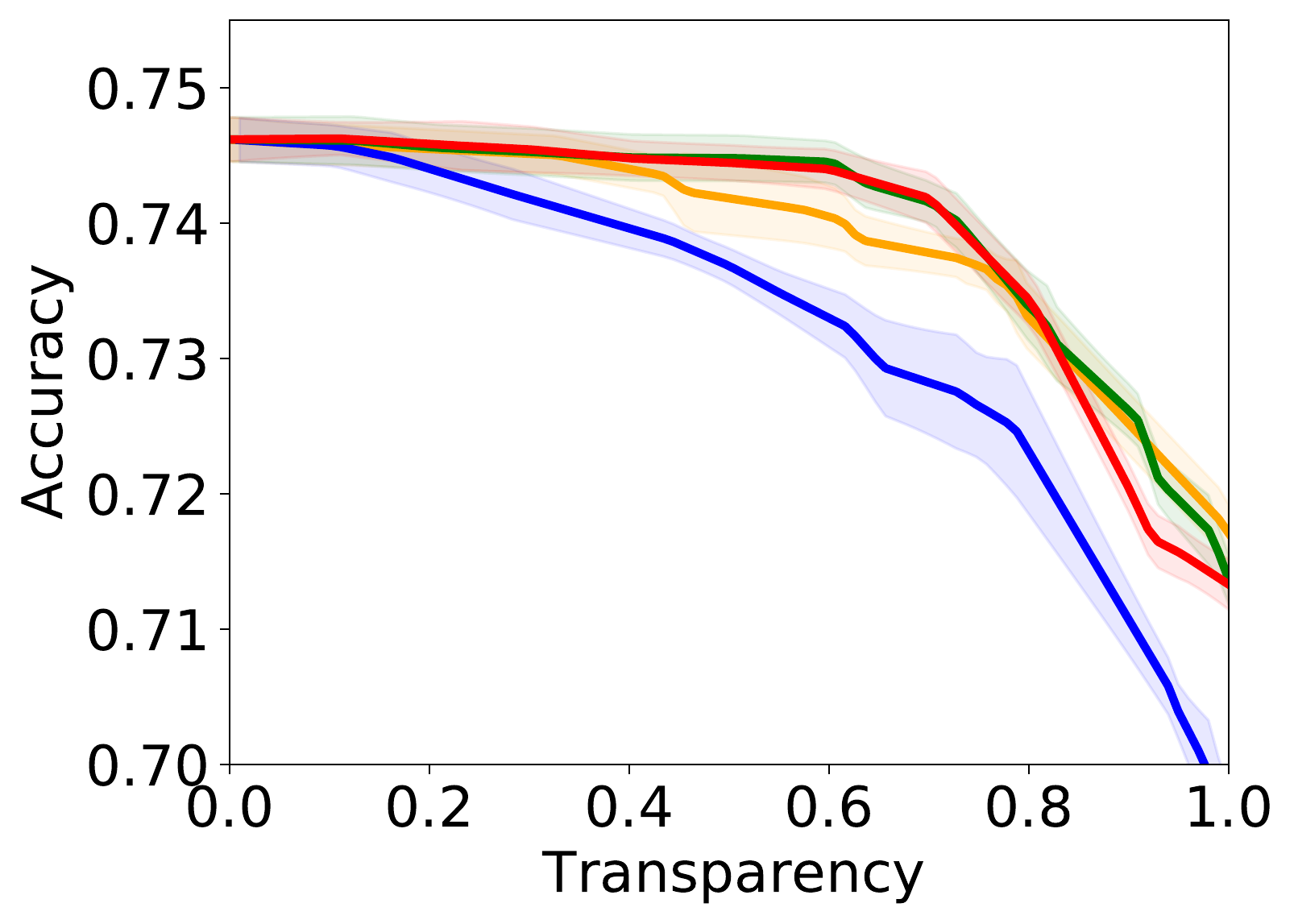}
        \includegraphics[width=\figwidth\textwidth]
         {acc_cov_test_acs_employ_gradient_boost.pdf}
        \caption{ACS Employment dataset.}
    \end{subfigure}

    \vskip 10 pt
    
    \begin{subfigure}{\textwidth}
        \centering
        \label{fig:results_tradeoffs_adult_test}
          \includegraphics[width=\figwidth\textwidth]
         {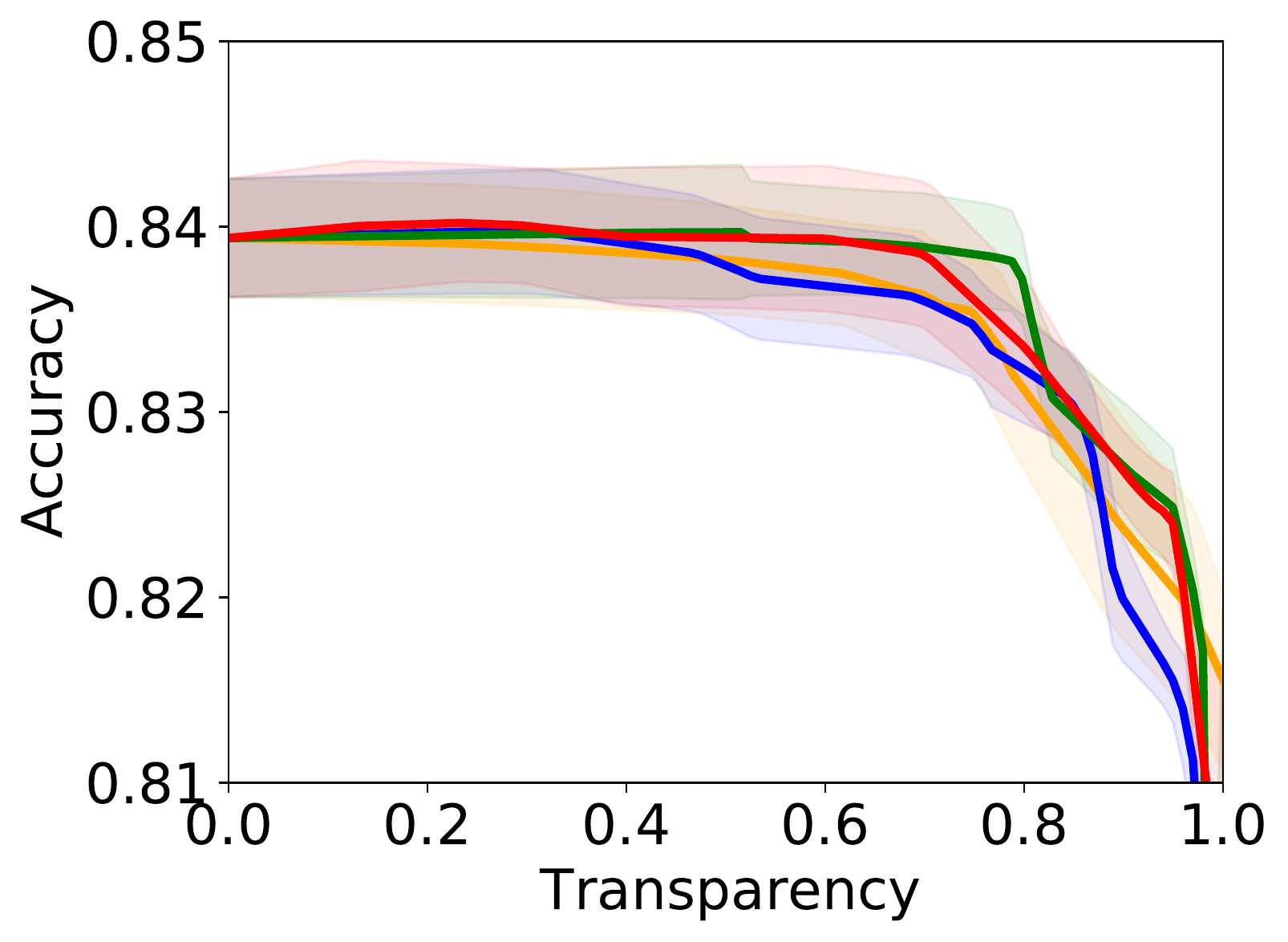}
         \includegraphics[width=\figwidth\textwidth]
         {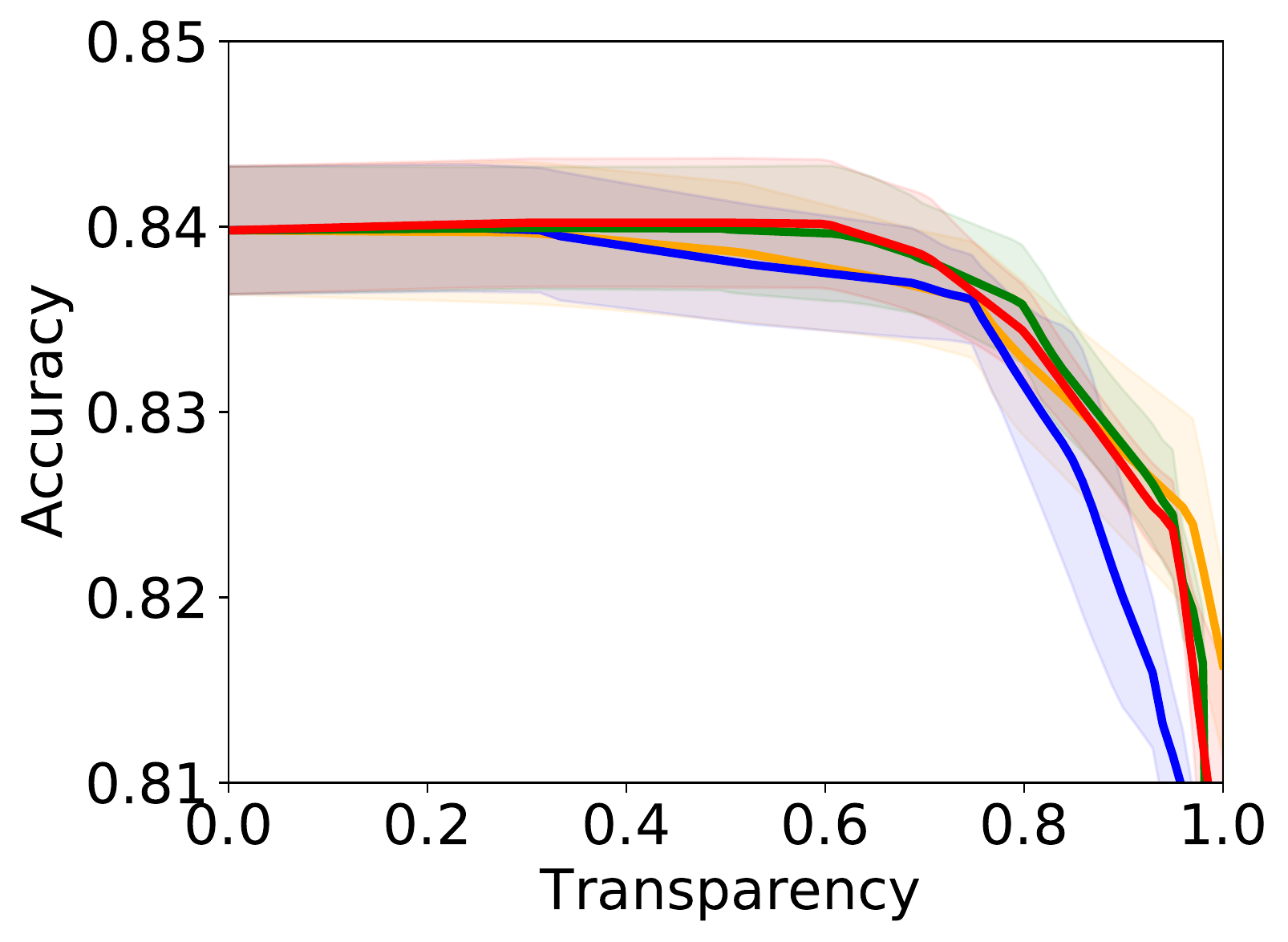}
        \includegraphics[width=\figwidth\textwidth]
         {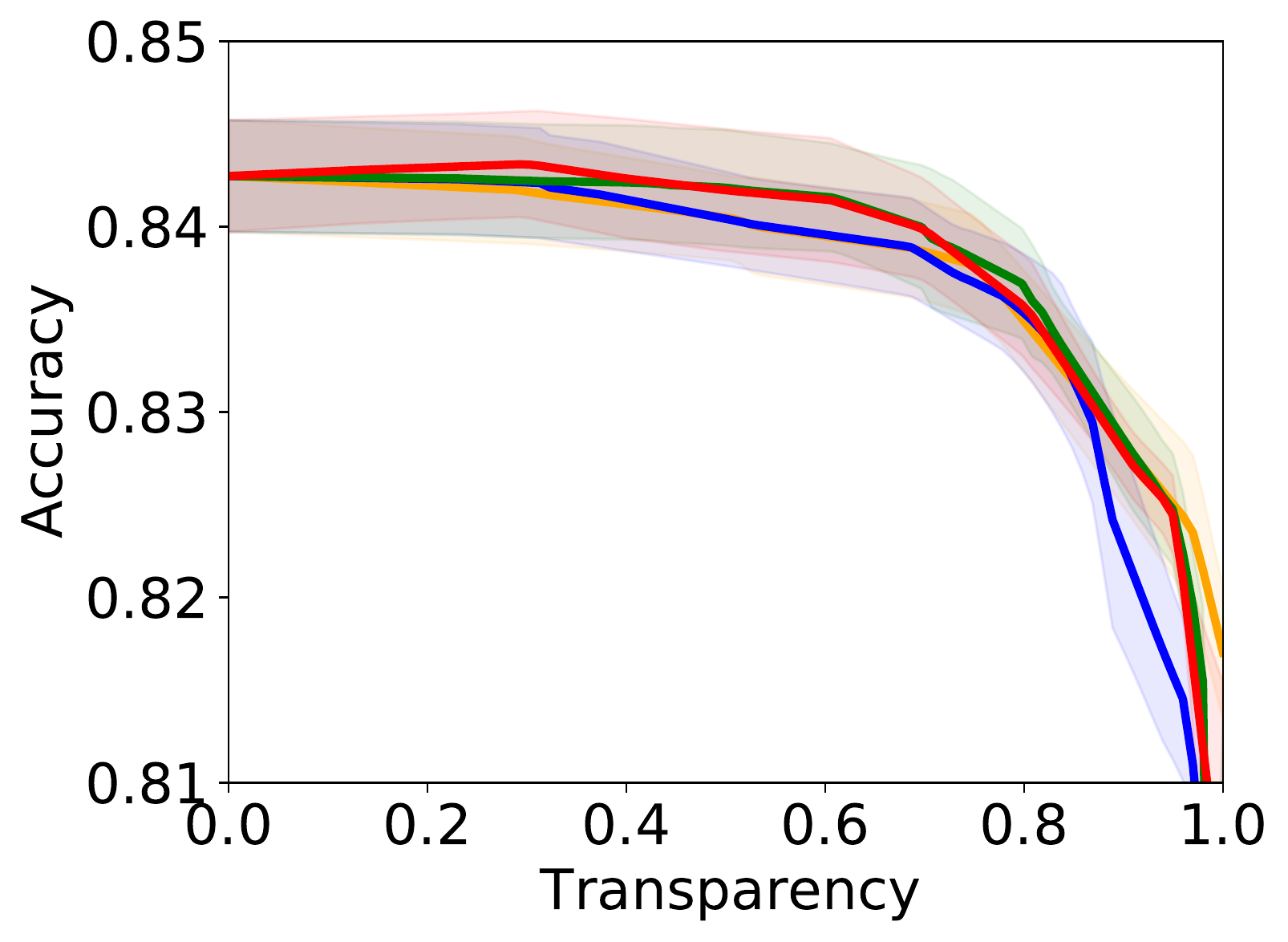}
        \caption{UCI Adult Income dataset.}
    \end{subfigure}
    
    \vskip 10 pt

    \begin{subfigure}{\textwidth}
        \centering
        \label{fig:results_tradoffs_compas_test}
        \includegraphics[width=\figwidth\textwidth]
         {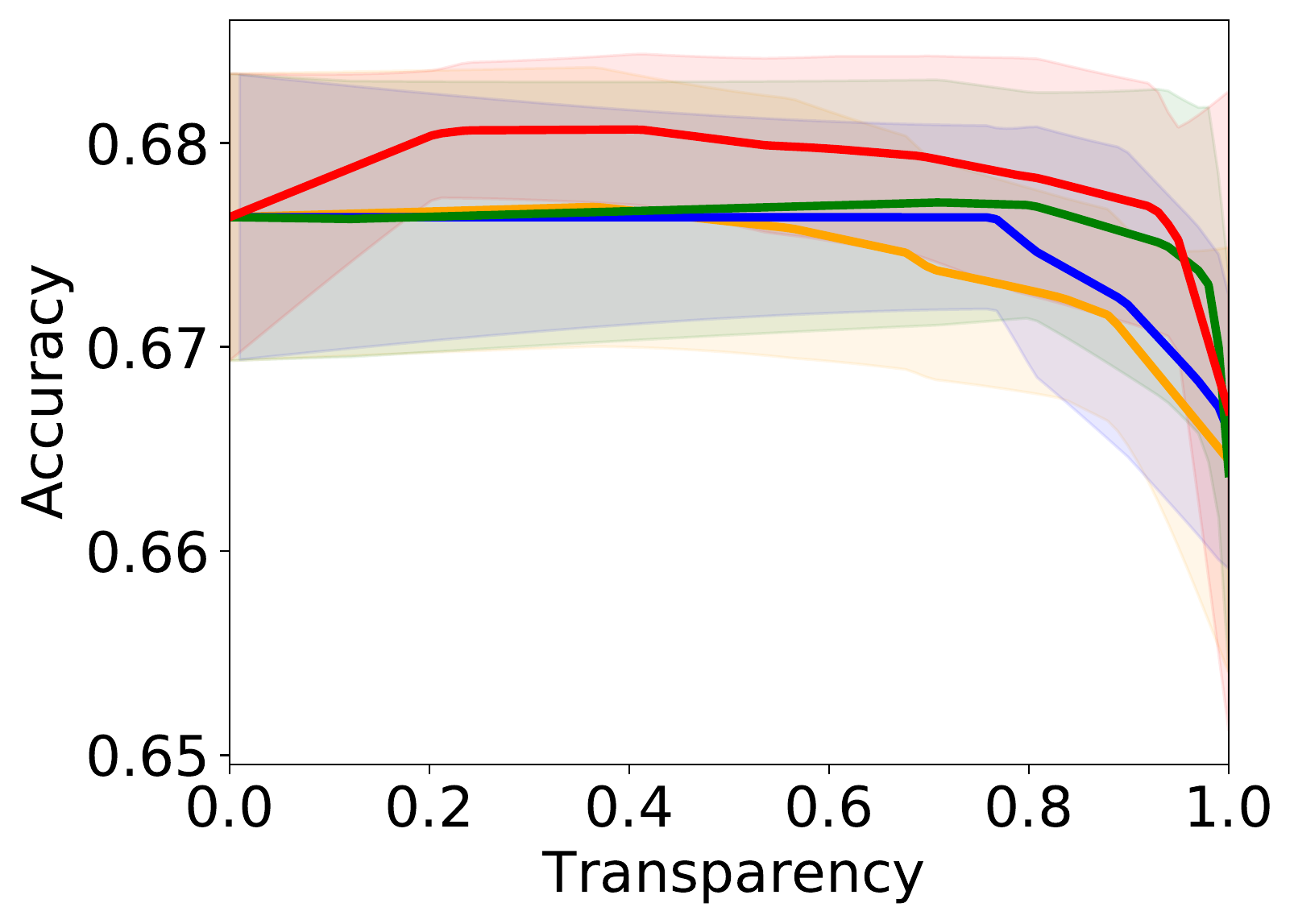}
         \includegraphics[width=\figwidth\textwidth]
         {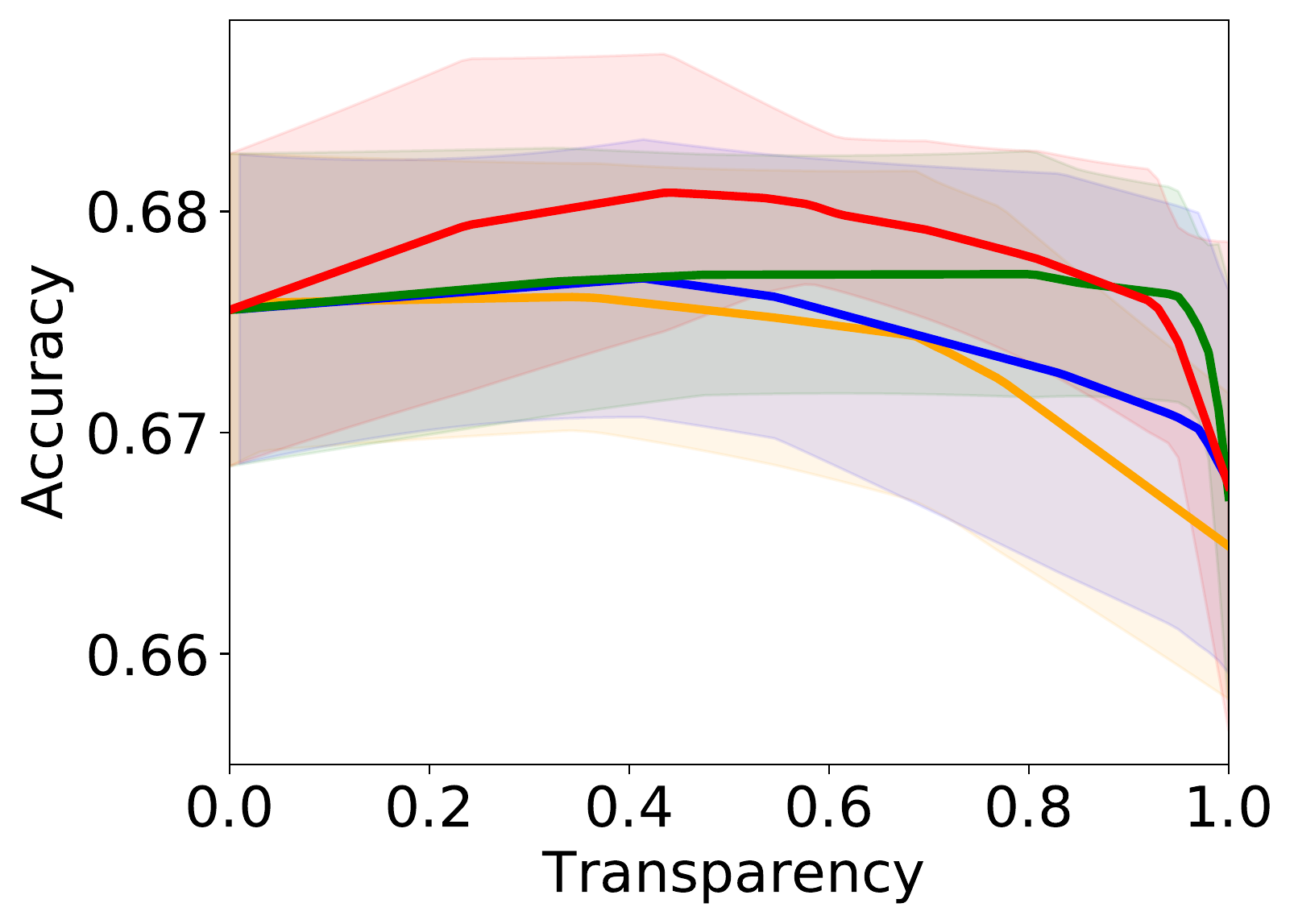}
        \includegraphics[width=\figwidth\textwidth]
         {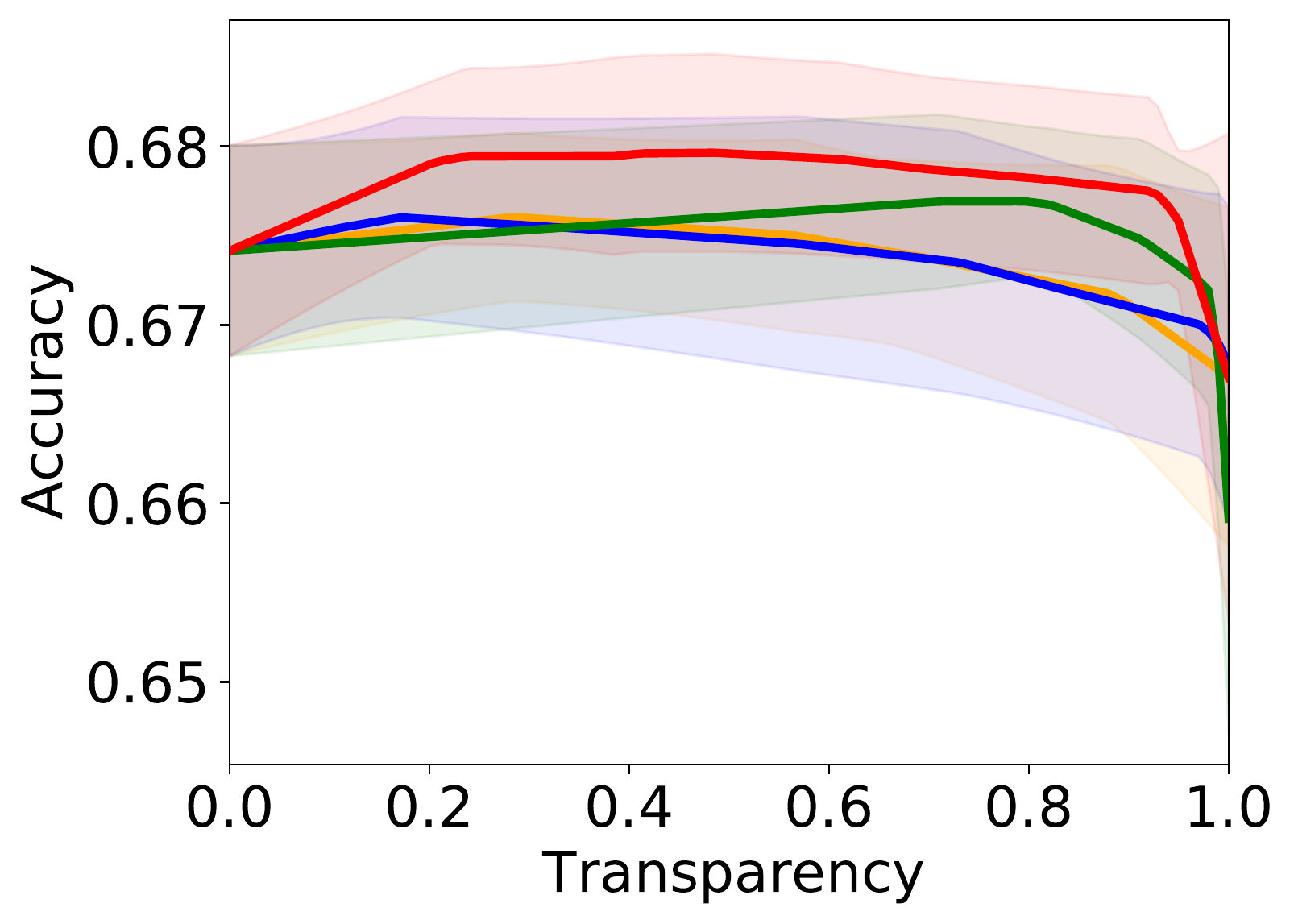}
        \caption{COMPAS dataset.}
    \end{subfigure}
    
     \vskip 10 pt
   
   \includegraphics[width=0.9\textwidth]
   {acc_cov_test_legend.pdf}
   
    \caption{Test set accuracy/transparency trade-offs for various hybrid models learning frameworks and datasets. The Pareto front for each method is represented as a line and the filled bands encode the std across the five data split reruns.
    Results are provided for several black-boxes: (Left) AdaBoost, (Middle) Random Forests, (Right) Gradient Boosted Trees.}
    \label{fig:results_tradeoff_test}
    
    \end{center}
\end{figure*}
\begin{figure}[t!]
    \centering
    \begin{subfigure}[b]{0.9\textwidth}
\begin{lstlisting}[language=RuleListsLanguage,backgroundcolor=\color{RuleListsLanguageBackgroundColor}, basicstyle=\scriptsize]
if ["age_medium" and "No Cognitive difficulty"] then 1
else if ["age_high"] then 0
else
    AdaBoost()
\end{lstlisting}
        \caption{HyRS: Test Accuracy 72.8\%, Transparency 64.3\%}
        \vspace{7pt}
    \end{subfigure}
    \vspace{7pt}
    \begin{subfigure}[b]{0.9\textwidth}
\begin{lstlisting}[language=RuleListsLanguage,backgroundcolor=\color{RuleListsLanguageBackgroundColor}, basicstyle=\scriptsize]
if  ["age_high" and "Female"] then 0
else if ["age_high" and "Native"] then 0
else if ["Reference person" and "No disability"] then 1
else if ["Husband/wife" and "No disability"] then 1
else if ["Cognitive difficulty" and "not own child of householder"] then 0
else 
    AdaBoost()
\end{lstlisting}
        \caption{CRL: Test Accuracy 73.7\%, Transparency 75.8\%.}
    \end{subfigure}
    \vspace{7pt}
    \begin{subfigure}[b]{0.9\textwidth}
\begin{lstlisting}[language=RuleListsLanguage,backgroundcolor=\color{RuleListsLanguageBackgroundColor}, basicstyle=\scriptsize]
if ["Disability" and "age_high"] then 0
else if ["Husband/wife" and "Male"] then 1
else if ["age_high" and "Native"] then 0
else if ["age_high" and "Female"] then 0
else if ["Reference person" and "No disability"] then 1
else if ["Bachelor degree"] then 1
else
    AdaBoost()
\end{lstlisting}
        \caption{\HybridCORELSPre{}: Test Accuracy 74.0\%, Transparency 70.1\%.}  
    \end{subfigure}
    \vspace{7pt}
    \begin{subfigure}[b]{0.9\textwidth}
\begin{lstlisting}[language=RuleListsLanguage,backgroundcolor=\color{RuleListsLanguageBackgroundColor}, basicstyle=\scriptsize]
if ["age_high" and "Female"] then 0
else if ["Husband/wife" and "No disability"] then 1
else if ["age_high" and "Native"] then 0
else if ["Reference person" and "No disability"] then 1
else
    AdaBoost()

\end{lstlisting}
        \caption{\HybridCORELSPost{} : Test Accuracy 73.7\%, Transparency 73.0\%.}
    \end{subfigure}
    \caption{Example hybrid interpretable models obtained by the different methods on the same data split of 
    the ACS Employment dataset with a AdaBoost black-box.}
    \label{fig:acs_adaboost_fold_1_rules}
\end{figure}

The test set accuracy/transparency trade-offs of the different hybrid models learning frameworks are shown in Figure~\ref{fig:results_tradeoff_test} 
for each dataset and black-box type. We highlight three main insights from these results.

First, on almost all datasets and black-box types, the methods \HybridCORELSPre{} and \HybridCORELSPost{} are better or
equivalent to \hyrs{} and \crl{}. The only exception is \HybridCORELSPre{} in high transparency regimes (0.85-1.0) on the ACS Employment dataset. 
The reason \HybridCORELS{} is so competitive with state-of-the-art methods is that it solves its objective to optimality, exploring the whole search space of prefixes (which methods based on local search can hardly achieve). 
Hence, given a learning paradigm and a transparency constraint, it builds the prefix that maximizes accuracy.
Furthermore, unlike other approaches, \HybridCORELS{} offers precise control over the desired level of transparency.
In Figure \ref{fig:acs_adaboost_fold_1_rules}, we show example hybrid models for each of the four methods fitted on the same data split (train/validation/test) of the ACS Employment dataset with an AdaBoost black-box. 
These models were selected on the basis of having the highest test accuracies for test transparencies restricted between 0.6 and 0.8. 
We note that \HybridCORELSPre{} and \HybridCORELSPost{} are competitive with \crl{} and even employ similar rules, for example, 
[\texttt{\textcolor{red}{"age\_high"} \textcolor{blue}{and} \textcolor{red}{"Female"}}], 
[\texttt{\textcolor{red}{"Reference person"} \textcolor{blue}{and} \textcolor{red}{"No disability"}}], 
and [\texttt{\textcolor{red}{"age\_high"} \textcolor{blue}{and} \textcolor{red}{"Native"}}]. 
\hyrs{} on the other hand, performs worst than the other three since it has a lesser accuracy and transparency.

Secondly, using \HybridCORELS{} on the ACS Employment and UCI Adult Income datasets, one can reach high transparency values (0.7) while retaining the same performance as the black-box (0.0 transparency). This observation is consistent across all black-box types, which suggests that complex models 
are often overkill in certain regions of the input space and can safely be replaced by a simpler model on those inputs. 
From the point of view of certification/maintenance of a machine learning model,
being able to assign a majority of inputs to an interpretable component is a tremendous step forward. For instance, since rule lists are interpretable, one might be able to certify that the hybrid model works properly/safely on the region $\Omega_r$ that will contain the majority of examples seen in deployment. For the minority of instances that fall outside the region, certification might require the verification of the opaque decisions by a committee of domain experts. Such verification might be time-consuming but, the higher the transparency, the fewer examples this committee would need to verify regularly. 

\begin{figure}[t!]
    \centering
    \begin{subfigure}[b]{0.9\textwidth}
\begin{lstlisting}[language=RuleListsLanguage,backgroundcolor = \color{RuleListsLanguageBackgroundColor}, basicstyle=\scriptsize]
if ["Prior-Crimes=0" and "Age>=30"] then 0
else if ["Prior-Crimes>5" and "Age=24-30"] then 1
else if ["Prior-Crimes=1-3" and "Age>=30"] then 0
else
    RandomForest()
\end{lstlisting} 
    \caption{\HybridCORELSPre{}: Test Accuracy 68.1\%, Transparency 42.7\%}
    \end{subfigure}
    \vspace{2pt}
    \caption{Example hybrid interpretable model obtained by \HybridCORELSPre{} on the COMPAS dataset with a Random Forest black-box. 
    Consistent with Figure~\ref{fig:results_tradeoff_test}, this model generalizes better than the black-box alone.}
    \label{fig:compas_rf_example_pre_fold_3}
\end{figure}

Thirdly, when studying \HybridCORELSPre{} fitted on COMPAS,
one can consistently observe a ``sweet spot'' for transparency where the generalization is maximal and even better than the standalone black-box. The existence of such a ``sweet spot'' is predicted by our theory of \textbf{Section \ref{subsec:theory_sweet_spot}} and highlights the regularization effect of the hybrid modeling. Although retaining the same level of performance while increasing the transparency is enough to argue in favor of hybrid modeling 
(as was the case with the ACS Employment and UCI Adult Income datasets), it is interesting to see that hybrid models can also improve the 
generalization performance. This generalization improvement is mainly observed with the \HybridCORELSPre{} method, which constitutes an argument in favor 
of the \prebb{} paradigm.  We report in Figure~\ref{fig:compas_rf_example_pre_fold_3} an example model learned with \HybridCORELSPre{} on COMPAS,
which generalizes better than a standalone black-box. As we observe, just adding three simple rules before the 
black-box model allows for test accuracy improvements.

\section{Conclusion}

In this paper, we laid the foundations for a promising line of work that was initiated some years ago: hybridizing interpretable and black-box models to ``take the best of both worlds". 
More precisely, we first provided theoretical evidence that such models have generalization advantages, while also being easier to certify and understand.
We then proposed a taxonomy of learning algorithms aimed at producing such models, along with a generic framework implementing the (new) \prebb{} paradigm. 
We introduced algorithms belonging to two identified paradigms, namely \prebb{} and \postbb{}.
Compared to state-of-the-art methods, our proposed approaches, coined \HybridCORELSPre{} and \HybridCORELSPost{}, certify the optimality of the learned models 
and provide direct control over the desired transparency level.
Our thorough experiments demonstrated the ability of the proposed \prebb{} paradigm and the high competitivity of our algorithms with the state-of-the-art.
Furthermore, empirical findings suggest that this new paradigm may lead to better-generalizing models. Investigating the reasons for this observation is an interesting future work.
Adapting other optimal search-based learning algorithms (as was done with \corels{} in this work) - for instance those producing optimal sparse decision trees - to produce new forms of hybrid interpretable models is also a promising research avenue.
Finally, designing fully end-to-end and certifiably optimal hybrid interpretable models' learning algorithms is an open challenge, whose main difficulty consists in encoding both parts of the model within a unified framework to train them jointly.



\acks{This work is partially supported by the Canada Research Chairs (Privacy-preserving and ethical analysis of Big Data chair), the LabEx CIMI (ANR-11-LABX-0040), and the NSERC Discovery Grants program (2022-04006).\\ 
The authors also wish to thank the DEEL project CRDPJ 537462-18 funded by the National Science and Engineering Research Council of Canada 
(NSERC) and the Consortium for Research and Innovation in Aerospace in Québec (CRIAQ), together with its industrial partners Thales Canada inc, 
Bell Textron Canada Limited, CAE inc and Bombardier inc.} 


\newpage 

\vskip 0.2in
\bibliography{sample}

\newpage
\appendix

\section{Proof of Theorem~\ref{theorem:generalization_bound}}
\label{app:proofs}
\setcounter{theorem}{0}
\begin{theorem}
Given a finite hybrid model space $(|\Hyb|< \infty$) and some
$\epsilon > 0$,
letting $C_\Omega:= 
\Prob_{\bm{x}\sim \mathcal{D}}[\bm{x}\in \Omega]$ be the transparency of $\Omega$, 
then for any distribution $\D$ where there exists a triplet $\tripletstar$
with zero generalization error (as defined in~(\ref{eq:optimal_model})), 
the following holds for a training set of size $M$:
 \begin{equation*}
      \failure \leq \sum_{\Omega\in \mathcal{P}}
        \mathcal{B}(\epsilon, C_\Omega, \Hc, \Hs, M),
 \end{equation*}
 where 
 \begin{equation*}
      \mathcal{B}(\epsilon, C_\Omega, \Hc, \Hs, M) := (1-|\Hc|)C_\Omega^M + (1-|\Hs|)C_{\overline{\Omega}}^M +
    |\Hc|(C_{\overline{\Omega}}e^{-\epsilon} + C_{\Omega})^M +
    |\Hs|(C_{\Omega}e^{-\epsilon} + C_{\overline{\Omega}})^M.
 \end{equation*}
 If we assume that the optimal subset $\Omega \equiv \Omega^\star$ is known in advance, then the bound tightens
\begin{equation*}
      \failure \leq \mathcal{B}(\epsilon, C_\Omega, \Hc, \Hs, M).
 \end{equation*}

\begin{proof}
The distribution $\D$ is fixed apriori and our only assumption is that it can be perfectly solved by a hybrid model in $\Hyb$.
Since we assume a perfect model exists in
$\Hyb$, we must have $\emploss{\S}(\triplet_\S)=0$.
Given $\epsilon>0$, our main objective is to upper bound the probability $\failure$ which corresponds to the probability of ``failure'' by the model.
Letting $\Hyb_\epsilon := \{\triplet \in \Hyb : 
\poploss(\triplet) > \epsilon\}$ be the set of all ``failing'' hybrid models, we have that
\begin{equation}
    \begin{aligned}
    \failure &\leq \Prob_{\S\sim \D^M}[\exists \triplet
    \in \Hyb_\epsilon \,\,\text{with}\,\,  \emploss{\S}(\triplet)=0]\\
    &\leq \sum_{\Omega \in \mathcal{P}}\Prob_{\S\sim \D^M}[\exists \triplet 
    \in \Hyb_\epsilon \,\,\text{with}\,\,  \emploss{\S}(\triplet)=0],
    \end{aligned}
    \label{eq:union_bound}
\end{equation}
where we have used the union bound over all $\Omega\in \mathcal{P}$.
From this point on, we will assume that the domain $\Omega$ is fixed. Letting
$C_\Omega:= \Prob_{\bm{x}\sim \mathcal{D}}[\bm{x}\in \Omega]$ and $\overline{\Omega}:=\X \setminus \Omega$, we
can see the distribution $\D$ as a mixture of two distributions $\D_c,\D_s$ with disjoint supports $\overline{\Omega}$ and $\Omega$. 
Formally, we have $\D = C_{\overline{\Omega}} \D_c + C_\Omega \D_s$. The edge cases $\text{supp}(\mathcal{D})\subset \Omega$ and 
$\text{supp}(\mathcal{D})\subset \overline{\Omega}$ are covered by setting $C_{\Omega}\!=\!1,C_{\overline{\Omega}}\!=\!0$ and 
$C_{\Omega}\!=\!0,C_{\overline{\Omega}}\!=\!1$ respectivelly. Sampling from such a mixture distribution $\D$ can be seen as a two-step process. 
First, we choose a number of instances $m\sim \text{Bin}(C_{\Omega}, M)$ from a binomial law of $M$ trials and probability 
$C_{\Omega}$ of success. Then we sample $m$ simple examples $\S_s\sim \D_s^m$, and sample $M-m$ hard examples $\S_c\sim \D_c^{M-m}$. We get
\begin{equation}
\begin{aligned}
    \Prob_{\S\sim \D^M}[\exists &\triplet 
    \in \,\Hyb_\epsilon \,\,\text{with}\,\,  \emploss{\S}(\triplet)=0]\\
    &= \ProbSBin[\exists \triplet 
    \in \Hyb_\epsilon \,\,\text{with}\,\,  \emploss{\S_c\cup \S_s}(\triplet)=0]\\
    &= \sum_{m=0}^M b(m;C_\Omega, M) 
    \ProbS[\exists \triplet 
    \in \Hyb_\epsilon \,\,\text{with}\,\,  \emploss{\S_c\cup \S_s}(\triplet)=0].
    \label{eq:bin_expansion}
\end{aligned}
\end{equation}

Where we have introduced $b(m;C_\Omega, M) := {M \choose m}C_\Omega^m(1- C_{\Omega})^{M-m}$ as the binomial coefficients. In this formula, there
are two extreme edges cases $m=0$ and $m=M$ which occur with probability
$C_{\overline{\Omega}}^M$ and $C_\Omega^M$ respectively. The issue with both of these extreme cases is that we are
meant to bound the population loss of the whole hybrid model while only one of its sub-models is evaluated on empirical data. 
We decide to employ trivial bounds which will become less and less dominant when the probability of 
these extreme cases goes to zero as $M\rightarrow \infty$, assuming $C_{\Omega}\in ]0,1[$.
\newline
\newline
\noindent
\textbf{Case} $\mathbf{m=0}$
\begin{equation*}
    \Prob_{\S_c\sim \D_c^M}[\exists \triplet 
    \in \Hyb_\epsilon \,\,\text{with}\,\,\emploss{\S_c}(h_c)=0]\leq 1
\end{equation*}

\noindent
\textbf{Case} $\mathbf{m=M}$
\begin{equation*}
    \Prob_{\S_s\sim \D_s^M}[\exists \triplet 
    \in \Hyb_\epsilon \,\,\text{with}\,\,\emploss{\S_s}(h_s)=0]\leq 1
\end{equation*}

\noindent
\textbf{Case} $\mathbf{0<m<M}$
Since the expected loss can be rewritten
\begin{equation*}
    \poploss(\triplet) = C_{\overline{\Omega}}
    \mathcal{L}_{\D_c}(h_c) + 
    C_{\Omega}
    \mathcal{L}_{\D_s}(h_s),
\end{equation*}
we have that
\begin{equation*}
    \mathcal{L}_{\D_c}(h_c) \leq \epsilon \,\,\text{and}\,\, \mathcal{L}_{\D_s}(h_s) \leq \epsilon \Rightarrow \poploss(\triplet) \leq \epsilon,
\end{equation*}
which implies
\begin{equation} \label{eq:failure_implies_one_of_two}
    \triplet \in \Hyb_\epsilon \Rightarrow h_c \in \mathcal{H}_{c,\epsilon} \,\,\,\text{or}\,\,\, h_s \in \mathcal{H}_{s,\epsilon},
\end{equation}
where $\mathcal{H}_{c,\epsilon}:=\{h_c\in \Hc : 
\mathcal{L}_{\D_c}(h_c) > \epsilon\}$ and
$\mathcal{H}_{s,\epsilon}:=\{h_s\in \Hs : 
\mathcal{L}_{\D_s}(h_s) > \epsilon\}$ are the sets of
complex and simple models ``failing'' on the distributions $\D_c$ and $\D_s$. Note that the ``or" in~(\ref{eq:failure_implies_one_of_two}) is not exclusive and both parts of the model may fail simultaneously, although it is not necessary.
Therefore the following holds
\begin{equation*}
\begin{aligned}
    \ProbS[\exists &\triplet 
    \in \Hyb_\epsilon \,\,\text{with}\,\,\emploss{\S_c\cup \S_s}(\triplet)=0]\\
    &\leq \ProbS[\{\exists h_c\in \mathcal{H}_{c,\epsilon} \,\,\text{s.t.} \,\,\emploss{\S_c}(h_c)=0\}\,\, \text{or}\,\,\{\exists h_s \in \mathcal{H}_{s,\epsilon} \,\,\text{s.t.}\,\,\emploss{\S_s}(h_s)=0\}]\\
    &\leq \Prob_{\S\sim \D_c^{M-m}}[\exists h_c\in \mathcal{H}_{c,\epsilon} \,\,\text{s.t.} \,\,\emploss{S}(h_c)=0] + \,\, \Prob_{S\sim \D_s^{m}}[\exists h_s \in \mathcal{H}_{s,\epsilon} \,\,\text{s.t.}\,\,\emploss{\S}(h_s)=0]\\
    &\leq |\Hc| e^{-\epsilon(M - m)} + |\Hs|e^{-\epsilon m},
\end{aligned}
\end{equation*}
where we have used the inequality 
$\Prob_{\S\sim \D_s^{m}}[\exists h_s \in \mathcal{H}_{s,\epsilon} \,\,\text{s.t.}\,\,\emploss{\S}(h_s)=0]
\leq |\Hs|e^{-\epsilon m}$ (Equation 2.9 of \cite{shalev2014understanding}),
and a similar one for $\Hc$. Going back to Equation (\ref{eq:bin_expansion}),
we get

\begin{equation*}
\begin{aligned}
    \Prob_{\S\sim \D^M}[\exists &\triplet 
    \in \Hyb_\epsilon \,\,\text{with}\,\,\emploss{S}(\triplet)=0]\\
    &= \sum_{m=0}^M b(m;C_\Omega, M) 
    \ProbS[\exists \triplet 
    \in \Hyb_\epsilon \,\,\text{with}\,\,  \emploss{\S_c\cup \S_s}(\triplet)=0]\\
    &\leq C_{\overline{\Omega}}^M + C_\Omega^M+
    \sum_{m=1}^{M-1} b(m;C_\Omega, M) \big(|\Hc| e^{-\epsilon(M - m)} + |\Hs|e^{-\epsilon m}\big)\\
    &=   C_{\overline{\Omega}}^M + C_\Omega^M+ 
    |\Hc|\sum_{m=1}^{M-1} b(m;C_\Omega, M) e^{-\epsilon(M - m)} + |\Hs|\sum_{m=1}^{M-1}b(m;C_\Omega, M) e^{-\epsilon m}\\
    &=   C_{\overline{\Omega}}^M + C_\Omega^M+
    |\Hc|\sum_{m=1}^{M-1} b(m;C_{\overline{\Omega}}, M) e^{-\epsilon m} + |\Hs|\sum_{m=1}^{M-1}b(m;C_\Omega, M) e^{-\epsilon m}\\
    &\leq   C_{\overline{\Omega}}^M + C_\Omega^M+
    |\Hc|\sum_{m=1}^M b(m;C_{\overline{\Omega}}, M) e^{-\epsilon m} + |\Hs|\sum_{m=1}^M b(m;C_\Omega, M) e^{-\epsilon m}\\
    &= (1-|\Hc|)C_\Omega^M + (1-|\Hs|)C_{\overline{\Omega}}^M +
    |\Hc|\sum_{m=0}^M b(m;C_{\overline{\Omega}}, M) e^{-\epsilon m} + |\Hs|\sum_{m=0}^M b(m;C_\Omega, M) e^{-\epsilon m}\\
    &= (1-|\Hc|)C_\Omega^M + (1-|\Hs|)C_{\overline{\Omega}}^M +
    |\Hc|(C_{\overline{\Omega}}e^{-\epsilon} + C_{\Omega})^M +
    |\Hs|(C_{\Omega}e^{-\epsilon} + C_{\overline{\Omega}})^M\\
    &:= \mathcal{B}(\epsilon, C_\Omega, \Hc, \Hs, M)
\end{aligned}
\end{equation*}
where for the second-to-last step we have used the
identity
\begin{equation*}
    \sum_{m=0}^M b(m;C_{\Omega}, M) e^{-\epsilon m}=(C_\Omega e^{-\epsilon} + C_{\overline{\Omega}})^M.
\end{equation*}

Finally, combining this with Equation (\ref{eq:union_bound}),
\begin{equation*}
    \failure \leq
    \sum_{\Omega\in \mathcal{P}}
    \mathcal{B}(\epsilon, C_\Omega, \Hc, \Hs, M),
\end{equation*}
which is the first desired result.

Now assuming that the optimal region
$\Omega\equiv \Omega^\star$ is known in
advance, then the logic of the proof
is identical except that we do not employ
a union bound over all $\Omega\in\mathcal{P}$ as in Equation
(\ref{eq:union_bound}).
\end{proof}
\end{theorem}

\section{Pseudo-Codes of the \HybridCORELS{} algorithms}

While the \corels{} algorithm and our proposed \HybridCORELS{} variants were already introduced in \textbf{Section~\ref{sec:hybridcorels}}, we describe them in more detail in this appendix section.
We first introduce some necessary notation that we later use to provide a detailed pseudo-code and description of the \corels{} algorithm.
We then depict our proposed variants \HybridCORELSPost{} and \HybridCORELSPre{} for learning hybrid interpretable models.

\subsection{Notations}

To formally describe the pseudo-code of the \corels{} algorithm and those of our modified \HybridCORELS{} variants, we first need to introduce some more detailed notation.
As mentioned in \textbf{Section~\ref{sec:hybridcorels_corels}}, a rule list $\rulelist$ consists in an ordered set of rules $\prefix$, called a prefix, followed by a default decision $\defconseq$. Then, we note: $\rulelist = (\prefix, \defconseq)$. 
Each individual rule $\singlerule_i$ involved within prefix $\prefix$ consists of an \emph{antecedent} $\oneant_i$ (``if" part of the rule, consisting in a Boolean assertion over the features' values) and a consequent $\oneconseq_i$ (``then" part of the rule, consisting in a prediction). We note: $\singlerule_i = \oneant_i \to \oneconseq_i$, and $\prefix = (\singlerule_1, \singlerule_2, \ldots, \singlerule_{\lenrulelist[\prefix]})$ with $\lenrulelist[\prefix]$ the length of prefix $\prefix$.

\subsection{\corels{}}

The pseudo-code of the \corels{} algorithm is provided within Algorithm~\ref{alg:corels}.
As mentioned in \textbf{Section~\ref{sec:hybridcorels_corels}}, \corels{} is a branch-and-bound algorithm exploring a prefix tree, in which each node corresponds to a prefix $\prefix$ and its children are prefixes formed by extending $\prefix$.
At each step of the exploration, the nodes belonging to the exploration frontier are sorted within a priority queue $\priorityqueue$, ordered according to a given search policy. \corels{} implements several such policies, including Breadth First Search, Depth First Search, and several Best First Searches. 
While these policies define the order in which the nodes of the prefix tree are ordered (and may affect the convergence speed), note that they do not affect optimality, and must all lead to the same optimal objective function value given sufficient time and memory.
At each step of the exploration, the most promising prefix $\prefix$ is popped from the priority queue $\priorityqueue$ (line~\ref{line:pop}). 
If its lower bound is greater than the best objective found so far (\emph{i.e.,} $\prefix$ can not lead to a rule list improving the current best objective function), it is discarded. 
Otherwise, it is used to build a rule list by appending a default prediction $\defconseq$ (line~\ref{line:rulelistbuilding}).
If the resulting rule list $\rulelist$ has a better objective function than the best one reached so far, the current best solution is updated at line~\ref{algl:bestUpdated}.
Finally, each possible extension of $\prefix$ formed by adding a new rule at the end of $\prefix$ gives a new node which is pushed into the priority queue at line~\ref{line:prefixPushed}.
The exploration is completed (and optimality is proved) once the priority queue is empty. Note that efficient data structures are used to cut the prefix tree symmetries: for instance, a prefix permutation map ensures that only the best permutation of every set of rules is kept.

\label{appendix:pseudo_codes_corels}

\begin{algorithm}[ht!]
\caption{\corels{}}\label{alg:corels}

\noindent \textbf{Input}: Training data $\dataset{}$ with
set of pre-mined antecedents $\minedrules$; 
initial best known rule list $\rulelist^0$ such that $\obj(\rulelist^0,\dataset) = \oneobj{}^0$
\hspace*{\algorithmicindent} 

\textbf{Output}: 
$(\rulelist^*, \oneobj{}^*)$ in which $\rulelist^*$ is a rule list with the minimum objective function value $\oneobj{}^*$

\begin{algorithmic}[1]
    \State $(\rulelist^c, \oneobj{}^c) \gets  (\rulelist^0, \oneobj{}^0)$
    \State $\priorityqueue \gets queue(())$     \Comment{Initially the queue contains the empty prefix $()$}
    \While{$\priorityqueue$ not empty} \Comment{Stop when the queue is empty}
        \State $\prefix{} \gets \priorityqueue.pop()$ \label{line:pop}
        \If{ \label{line:lb} $\lb(\prefix,\dataset) < \oneobj{}^c$}
            \State $\rulelist \gets (\prefix{}, \defconseq)$\label{line:rulelistbuilding} \Comment{Set default prediction $\defconseq$ to minimize training error}
            \State $\oneobj{} \gets \obj(\rulelist,\dataset) = \frac{\errs[\rulelist,\dataset]}{\lvert \dataset \rvert} + \corelsreg \cdot \lenrulelist[\prefix]$ \Comment{Compute objective $\obj{}(\rulelist, \dataset{})$} 
           \ \If{$\oneobj{} < \oneobj{}^c$}
                \State $(\rulelist^c, \oneobj{}^c) \gets (\rulelist, \oneobj{})$  \label{algl:bestUpdated}\Comment{Update best rule list and objective}
            \EndIf

        \For {$\oneant{}$ in $\minedrules \setminus \{ a_i \mid \exists \singlerule_i \in \prefix{}, \singlerule_i = \oneant_i \to \oneconseq_i \}$} \Comment{Antecedent $\oneant{}$ not involved in $\prefix{}$}
            \State $\singlerule_{new} \gets (\oneant{} \to \oneconseq{})$ \Comment{Set $\oneant{}$'s consequent $\oneconseq{}$ to minimize training error}
             \State $\priorityqueue.push(\prefix{} \cup \singlerule_{new})$ \label{line:prefixPushed}  \Comment{Enqueue extension of $\prefix{}$ with new rule $\singlerule_{new}$}
        \EndFor
        
    \EndIf
\EndWhile
\State $(\rulelist^*, \oneobj{}^*) \gets (\rulelist^c, \oneobj{}^c)$
\end{algorithmic}
\end{algorithm}


\subsection{\HybridCORELS{}}

A key difference between our proposed \HybridCORELS{} algorithms and the original \corels{} is that our methods aim at learning prefixes (expressing partial classification functions) while \corels{}' purpose is to learn rule lists (classification functions). Both \HybridCORELSPost{} and \HybridCORELSPre{} return prefixes (and not rule lists) and take as input an initial best known prefix $\prefix^0$ satisfying the transparency constraint (while the original \corels{} takes as input an initial rule list $\rulelist^0$).
A simple choice for the initial prefix $\prefix^0$ satisfying the transparency constraint is a constant majority prediction: $\prefix^0 \gets [(True \to \oneconseq{}_0)]$ (whose transparency is $1.0$).
In practice, we use such trivial initial solution for all our experiments.

\label{appendix:pseudo_codes_hybridcorels_post}
\begin{algorithm}[ht!]
\caption{\HybridCORELSPost{}}\label{alg:hybridcorels_post}

\noindent \textbf{Input}: Training data $\dataset{}$ with
set of pre-mined antecedents $\minedrules$; 
minimum transparency value $\mintransp$;
initial prefix $\prefix^0$ such that $\frac{\lvert \dataset_{\prefix^0} \rvert}{\lvert \dataset \rvert} \geq \mintransp$;
pre-trained black-box model $\blackbox$
\hspace*{\algorithmicindent} 

\textbf{Output}: 
$(\prefix^*, \oneobj{}^*)$ in which $\prefix^*$ is a prefix with the minimum objective function value $\oneobj{}^*$

\begin{algorithmic}[1]
    \State $(\prefix^c, \oneobj{}^c) \gets  (\prefix^0, \oneobj{}^0)$
    \State $\priorityqueue \gets queue(())$     \Comment{Initially the queue contains the empty prefix $()$}
    \While{$\priorityqueue$ not empty} \Comment{Stop when the queue is empty}
        \State $\prefix{} \gets \priorityqueue.pop()$
        \If{ \label{line:lb_post} $\lb(\prefix,\dataset) < \oneobj{}^c$}
            \State $\oneobj{} \gets \frac{\errs[\prefix, \dataset_\prefix]+\errs[h_c, \dataset \setminus \dataset_\prefix]}{\lvert \dataset \rvert} + \corelsreg \cdot \lenrulelist[\prefix] + \regcoeffhycorels \cdot \frac{\lvert \dataset \setminus \dataset_\prefix \rvert}{\lvert \dataset \rvert}$\label{line:objective_post_eval}\Comment{Compute objective $\obj_\text{post}(\prefix,\dataset)$}
           \ \If{$\oneobj{} < \oneobj{}^c$ and $\frac{\lvert \dataset_{\prefix} \rvert}{\lvert \dataset \rvert} \geq \mintransp$}\label{line:check_constraint_post}
                \State $(\prefix^c, \oneobj{}^c) \gets (\prefix, \oneobj{})$  \label{algl:bestUpdated_post}\Comment{Update best prefix and objective}
            \EndIf

        \For {$\oneant{}$ in $\minedrules \setminus \{ a_i \mid \exists \singlerule_i \in \prefix{}, \singlerule_i = \oneant_i \to \oneconseq_i \}$} \Comment{Antecedent $\oneant{}$ not involved in $\prefix{}$}
            \State $\singlerule_{new} \gets (\oneant{} \to \oneconseq{})$ \Comment{Set $\oneant{}$'s consequent $\oneconseq{}$ to minimize training error}
             \State $\priorityqueue.push(\prefix{} \cup \singlerule_{new})$ \Comment{Enqueue extension of $\prefix{}$ with new rule $\singlerule_{new}$}
        \EndFor
        
    \EndIf
\EndWhile
\State $(\prefix^*, \oneobj{}^*) \gets (\prefix^c, \oneobj{}^c)$
\end{algorithmic}
\end{algorithm}

The pseudo-code of \HybridCORELSPost{} is provided in Algorithm~\ref{alg:hybridcorels_post}. Key modifications include the use of a different objective function~(\ref{eq:obj_post}) at line~\ref{line:objective_post_eval}, aimed at evaluating the overall hybrid interpretable model's performances. One can note that the computation of the new objective function $\obj_\text{post}(\prefix,\dataset)$ requires access to the pre-trained black-box $\blackbox$, which is part of the algorithm's inputs. 
The original \corels{}' lower bound is valid and tight for our new objective (as discussed in \textbf{Section~\ref{sec:hybridcorels_post}}) so we keep this computation unchanged at line~\ref{line:lb_post}.
Finally, to ensure that the built prefix satisfies a given transparency constraint~(\ref{eq:transparency_constraint}), this condition is verified at line~\ref{line:check_constraint_post} before updating the current best solution at line~\ref{algl:bestUpdated_post}.

\label{appendix:pseudo_codes_hybridcorels_pre}
\begin{algorithm}[ht!]
\caption{\HybridCORELSPre{}}\label{alg:hybridcorels_pre}

\noindent \textbf{Input}: Training data $\dataset{}$ with
set of pre-mined antecedents $\minedrules$; 
minimum transparency value $\mintransp$;
initial prefix $\prefix^0$ such that $\frac{\lvert \dataset_{\prefix^0} \rvert}{\lvert \dataset \rvert} \geq \mintransp$
\hspace*{\algorithmicindent} 

\textbf{Output}: 
$(\prefix^*, \oneobj{}^*)$ in which $\prefix^*$ is a prefix with the minimum objective function value $\oneobj{}^*$

\begin{algorithmic}[1]
    \State $(\prefix^c, \oneobj{}^c) \gets  (\prefix^0, \oneobj{}^0)$
    \State $\priorityqueue \gets queue(())$     \Comment{Initially the queue contains the empty prefix $()$}
    \While{$\priorityqueue$ not empty} \Comment{Stop when the queue is empty}
        \State $\prefix{} \gets \priorityqueue.pop()$
        \If{ \label{line:lb_pre} $\lb(\prefix,\dataset) < \oneobj{}^c$}
            \State $\oneobj{} \gets \frac{\errs[\prefix,\dataset_r] + \incons(\dataset \setminus \dataset_\prefix)}{\lvert \dataset \rvert} + \corelsreg \cdot \lenrulelist[\prefix] + \beta \cdot \frac{\lvert \dataset \setminus \dataset_r \rvert}{\lvert \dataset \rvert}$\label{line:objective_pre_eval}\Comment{Compute objective $\obj_\text{pre}(\prefix,\dataset)$}
           \ \If{$\oneobj{} < \oneobj{}^c$ and $\frac{\lvert \dataset_{\prefix} \rvert}{\lvert \dataset \rvert} \geq \mintransp$}\label{line:check_constraint_pre}
                \State $(\prefix^c, \oneobj{}^c) \gets (\prefix, \oneobj{})$  \label{algl:bestUpdated_pre}\Comment{Update best prefix and objective}
            \EndIf

        \For {$\oneant{}$ in $\minedrules \setminus \{ a_i \mid \exists \singlerule_i \in \prefix{}, \singlerule_i = \oneant_i \to \oneconseq_i \}$} \Comment{Antecedent $\oneant{}$ not involved in $\prefix{}$}
            \State $\singlerule_{new} \gets (\oneant{} \to \oneconseq{})$ \Comment{Set $\oneant{}$'s consequent $\oneconseq{}$ to minimize training error}
             \State $\priorityqueue.push(\prefix{} \cup \singlerule_{new})$ \Comment{Enqueue extension of $\prefix{}$ with new rule $\singlerule_{new}$}
        \EndFor
        
    \EndIf
\EndWhile
\State $(\prefix^*, \oneobj{}^*) \gets (\prefix^c, \oneobj{}^c)$
\end{algorithmic}
\end{algorithm}

The pseudo-code of \HybridCORELSPre{} is provided in Algorithm~\ref{alg:hybridcorels_pre}. Again, the objective function computation is modified at line~\ref{line:objective_pre_eval} to use our proposed $\obj_\text{pre}(\prefix,\dataset)$ objective~(\ref{eq:obj_pre}). As before, the original lower bound is still valid (as discussed in \textbf{Section~\ref{sec:hybridcorels_pre}}) so we leave it unchanged at line~\ref{line:lb_pre}. Just like for \HybridCORELSPost{}, the transparency constraint~(\ref{eq:transparency_constraint}) is checked line~\ref{line:check_constraint_pre}, right before the current best solution update (line~\ref{algl:bestUpdated_pre}).
Once the optimal prefix $\prefix^*$ is known, the black-box part can be trained (which is not represented in the pseudo-code) using our proposed specialization scheme as described in \textbf{Section~\ref{sec:weighting_scheme}}.

Finally, both our proposed approaches are \emph{anytime}: the user can specify any desired running time and memory limits, and the algorithm returns the current best solution and objective value $(\prefix^c, \oneobj^c)$ if one of the limits is hit and the priority queue is not empty. Even if optimality is not guaranteed in such case, the ability to precisely bound running times and memory footprints is a very practical feature for real-life applications.

\section{Another \prebb{} Implementation for \HybridCORELS{}}
\label{appendix:hybridcorelsprenocollab}

In this appendix section, we describe another possible implementation of the \prebb{} paradigm based on the \corels{} algorithm but optimizing a different objective function.
We discuss the theoretical differences with the \HybridCORELSPre{} algorithm introduced in \textbf{Section~\ref{sec:hybridcorels_pre}} and empirically compare the two methods.

\subsection{\HybridCORELSPreNoCollab{}: Theoretical Presentation}
\label{appendix:hybridcorelsprenocollab_theory}
We now introduce another possible variant of \corels{} implementing the \prebb{} paradigm.
We coin it \HybridCORELSPreNoCollab{}, because contrary to the \HybridCORELSPre{} algorithm introduced in \textbf{Section~\ref{sec:hybridcorels_pre}}, the prefix learning phase of \HybridCORELSPreNoCollab{} does not account for the task left to the black-box part.
Instead, the prefix is learned to maximize its own accuracy, which results in the remaining examples (that will be handled by the black-box model) being the hardest ones to classify. 
While black-box specialization could be helpful to deal with such difficult tasks, we observe that, in practice, it has to deal with many inconsistent examples, which considerably lowers its performances.

\paragraph{Objective} \HybridCORELSPreNoCollab{} builds a prefix $\prefix$ capturing at least a proportion of $\mintransp$ of the training data (transparency constraint~(\ref{eq:transparency_constraint})), and minimizing the weighted sum of $\prefix$'s classification error and sparsity: 
\begin{align}
\obj_\text{pre,nocollab}(\prefix,\dataset) = \frac{\errs[\prefix,\dataset_\prefix]}{\lvert \dataset_\prefix \rvert} + \corelsreg \cdot \lenrulelist[\prefix] + \regcoeffhycorels \cdot \frac{\lvert \dataset \setminus \dataset_\prefix \rvert}{\lvert \dataset \rvert}\label{eq:obj_pre_no_collab}
\end{align} 

\paragraph{Objective lower bound} \corels{}' original lower bound~(\ref{eq:lb_original}) does not hold for objective function~(\ref{eq:obj_pre_no_collab}). Indeed, the difficulty here is that $\obj_\text{pre,nocollab}$ quantifies a prefix's error only on the subset of examples that it classifies ($\dataset_\prefix$), hence it is not possible to directly consider the inconsistent examples $\incons(\dataset \setminus \dataset_\prefix)$ as in $\lb$~(\ref{eq:lb_original}): an extension of $\prefix$ may not capture them at all. 
To obtain a tight lower bound $\lb_\text{pre,nocollab}$, one needs to consider simultaneously the support $\dataset_\prefix$ and errors $\errs[\prefix, \dataset_\prefix]$ of prefix $\prefix$, as well as the labels cardinalities among each group of inconsistent examples (also called \emph{set of equivalent points} in the context of \corels{}~\citep{DBLP:journals/jmlr/AngelinoLASR17}). 
A pre-processing step computes a list $\mathcal{G}$ of \emph{inconsistent groups of examples}. 
Each group $g \in \mathcal{G}, g \subset \dataset{}$ is defined by its number of minority examples $min_g$ (those with the least frequent label among group $g$), and its number of majority examples $maj_g$. 
In fact, our previously introduced count of unavoidable errors uses such groups for its computation: $\incons(\dataset) = \sum_{g \in \mathcal{G}}{(min_g)}$.
For each group $g \in \mathcal{G}$ not captured by prefix $\prefix{}$ ($g \not\subset \dataset{}_{\prefix}$), we verify whether capturing its examples could lower the current prefix's error rate: $c_{p,g} = \indicator{}\left[\frac{min_g}{min_g+maj_g} \leq \frac{\errs[\prefix, \dataset_\prefix]}{\lvert \dataset_\prefix \rvert}\right]$. Then:
\begin{align}
     \lb_\text{pre,nocollab}(\prefix,\dataset) =& \frac{\errs[\prefix, \dataset_\prefix] + \sum_{g\in \mathcal{G}, g \not\subset \dataset{}_{\prefix}}{c_{p,g}  \cdot min_g}}{\lvert \dataset_\prefix \rvert + \sum_{g\in \mathcal{G}, g \not\subset \dataset{}_{\prefix}}{c_{p,g}  \cdot (min_g + maj_g)}}\nonumber\\
     &+ (K_\prefix+1) \cdot \corelsreg \label{eq:lb_pre_tight_no_collab}
\end{align}
Finally, $\lb_\text{pre,nocollab}$ precisely quantifies the best objective function that can be reached based on prefix $\prefix$, by only capturing inconsistent groups of examples that improve the objective function (lowering the error rate). The definition of $c_{p,g} $ uses a less or equal operator because in case the error rate is unchanged after capturing an additional group of inconsistent examples, the operation should be performed as it would increase the coverage (and the associated regularisation term). There exists a (partial) classification function whose error rate is exactly the one computed in $\lb_\text{pre,nocollab}$, so this bound is tight.

Finally, \HybridCORELSPreNoCollab{} is an exact method: it provably returns a prefix $\prefix$ for which $\obj_\text{pre,nocollab}(\prefix,\dataset)$~(\ref{eq:obj_pre_no_collab}) is the smallest among those satisfying the transparency constraint~(\ref{eq:transparency_constraint}). 
This means that, given desired transparency level, it produces an optimal prefix (interpretable part of the final hybrid model) in terms of accuracy/sparsity. The pseudo-code of \HybridCORELSPreNoCollab{} is similar to that of \HybridCORELSPre{} presented in Algorithm~\ref{alg:hybridcorels_pre}, except that the objective function $\obj_\text{pre}(\prefix,\dataset)$ and lower bound $\lb(\prefix,\dataset)$ on lines~\ref{line:objective_pre_eval} and~\ref{line:lb_pre} are replaced by $\obj_\text{pre,nocollab}(\prefix,\dataset)$ and $\lb_\text{pre,nocollab}(\prefix,\dataset)$, as introduced in equations~(\ref{eq:obj_pre_no_collab}) and~(\ref{eq:lb_pre_tight_no_collab}).

Again, note that within this proposed implementation, the prefix learning phase does not consider the difficulty of the task let to the black-box learning part. 
For datasets containing inconsistent examples, this could result in sub-optimal overall accuracy in regimes of medium to high transparency, when collaboration between both parts of the hybrid interpretable model is required. 

\subsection{\HybridCORELSPreNoCollab{}: Empirical Evaluation}
\label{appendix:hybridcorelsprenocollab_expes}

We ran the experiments of \textbf{Section~\ref{sec:expes_all}} using \HybridCORELSPreNoCollab{} (with the same setup as~\HybridCORELSPre{}), and provide a comparison with \HybridCORELSPre{} within Figure~\ref{fig:results_tradeoff_test_pre_no_collab}.
The results show that for very low transparency values, \HybridCORELSPreNoCollab{} and \HybridCORELSPre{} have very close performances.
Indeed, in such regimes, most of the classification task is handled by the black-box part of the model and the absence of collaboration with the 
interpretable part does not really matter.
We observe the same phenomenon in regimes of very high transparency, where most of the examples are classified by the interpretable part.
However, in regimes of medium to high transparency, we observe a significant drop of \HybridCORELSPreNoCollab{}'s performances.
This trend is particularly visible with the ACS Employment dataset.
It can be explained by the absence of collaboration between both parts of the model: the prefix learning sacrifices the black-box performances 
(sending it most of the inconsistent examples) to obtain the most accurate prefix possible. 
While this policy leads to slightly more accurate interpretable parts compared to the prefixes learned by \HybridCORELSPre{}, it also harms the overall model accuracy considerably, and the obtained trade-offs are not competitive with those produced by \HybridCORELSPre{}.
As observed in \textbf{Section~\ref{sec:expes_all}} with~\HybridCORELSPre{}, on the COMPAS dataset, hybrid models with intermediate transparency 
values exhibit better test accuracies than the standalone black-box, due to better generalization. Again, this constitutes an argument 
in favor of the \prebb{} paradigm, as this trend was not observed with the other \postbb{} methods.

We provide in Figure~\ref{fig:acs_adaboost_compare_pre_fold_0} examples of hybrid models found with \HybridCORELSPre{} and 
\HybridCORELSPreNoCollab{} on the same data splits of the ACS Employment dataset and transparencies roughly $80\%$.
We observe, as aforementioned, that the black-boxes trained after the~\HybridCORELSPreNoCollab{} prefixes exhibit considerably lower performances.
On the other hand, the prefix and black-box parts of the models trained using~\HybridCORELSPre{} have comparable classification performances, as the former was trained while accounting for the inconsistent samples that would be left for the later to classify.

\def\figwidth{0.326}
\begin{figure*}[htbp]
    \begin{center}

    \begin{subfigure}{\textwidth}
        \centering
        \label{fig:results_tradoffs_acs_employ_test_pre_no_collab}
        \includegraphics[width=\figwidth\textwidth]
         {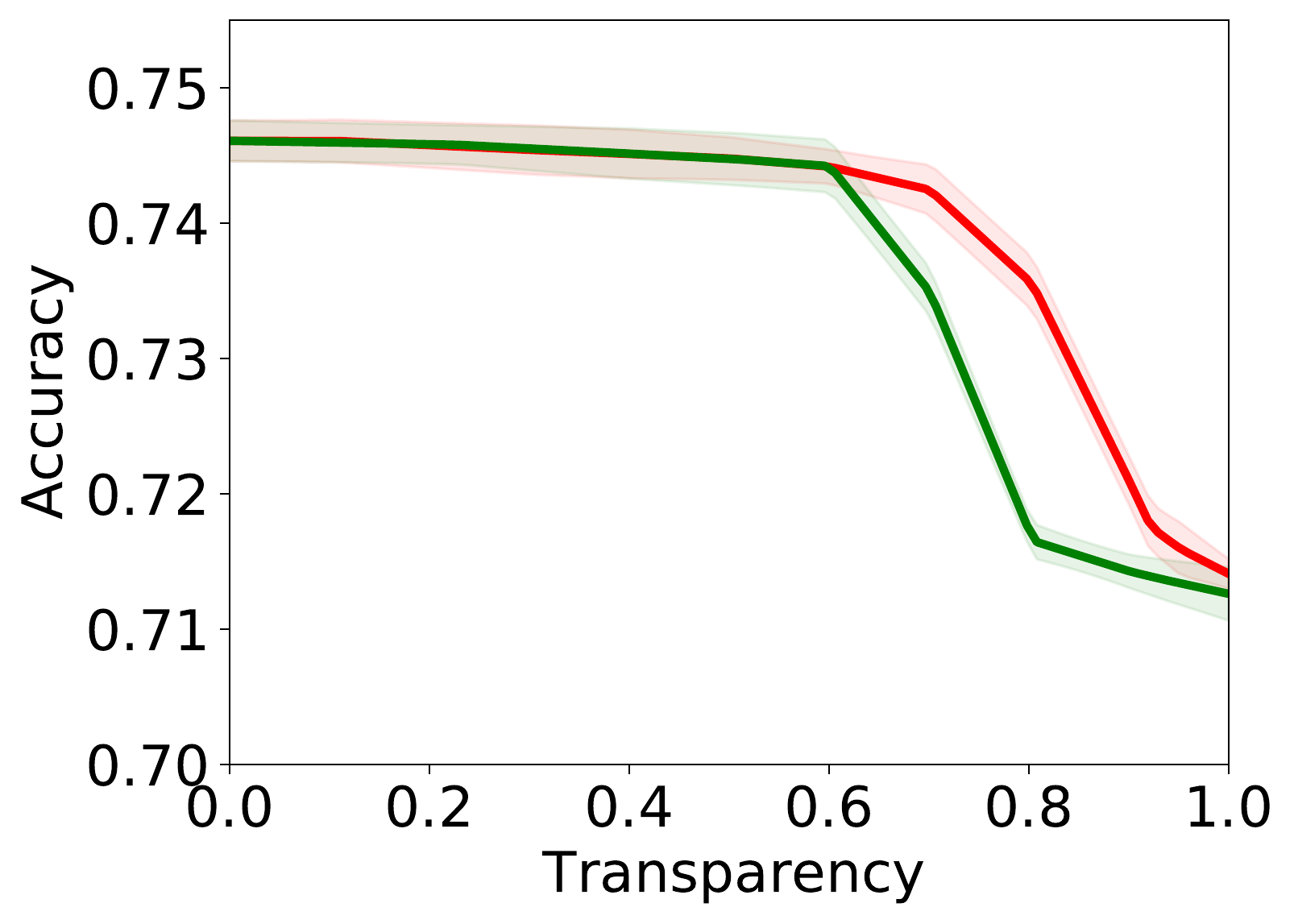}
         \includegraphics[width=\figwidth\textwidth]
         {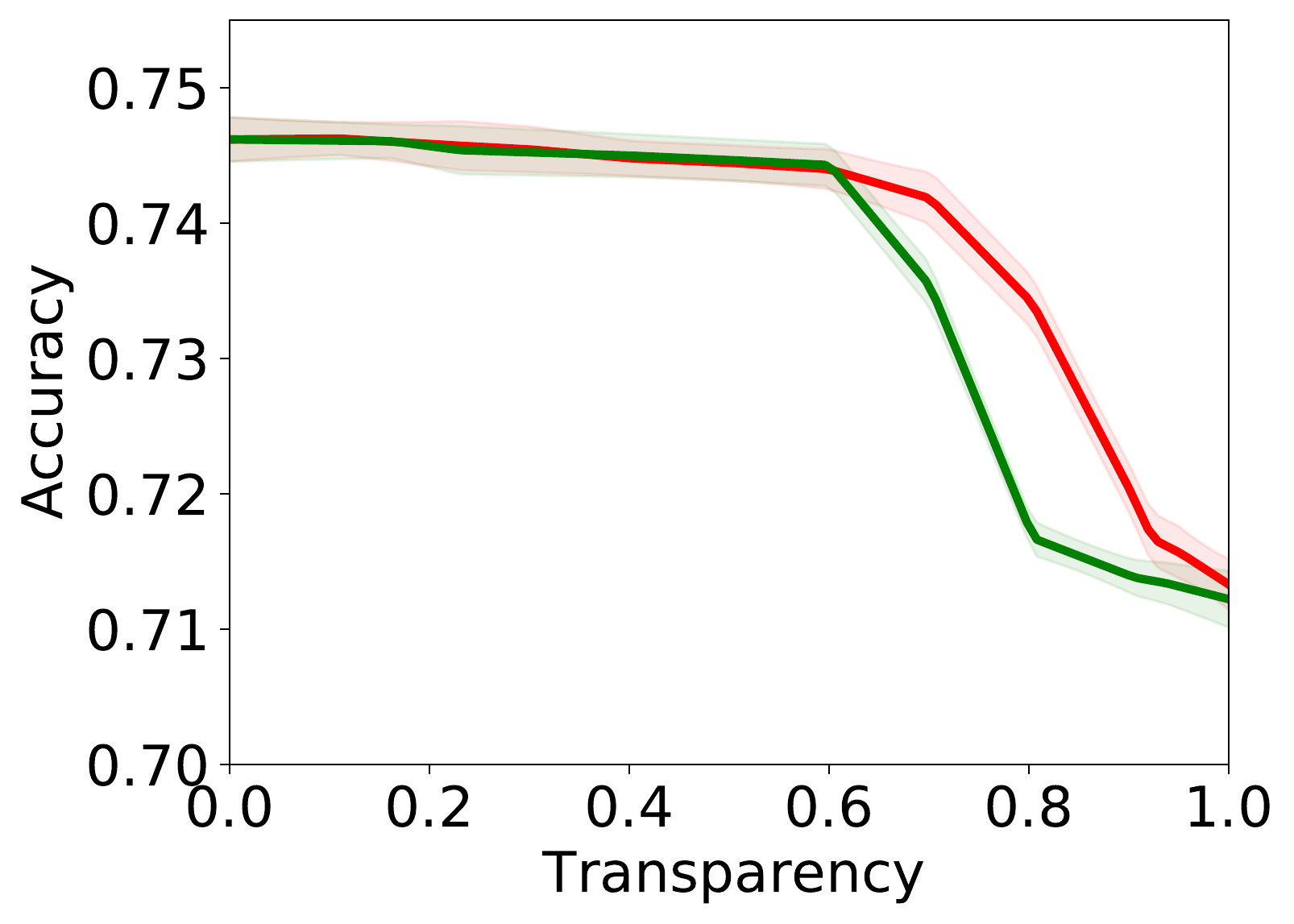}
        \includegraphics[width=\figwidth\textwidth]
         {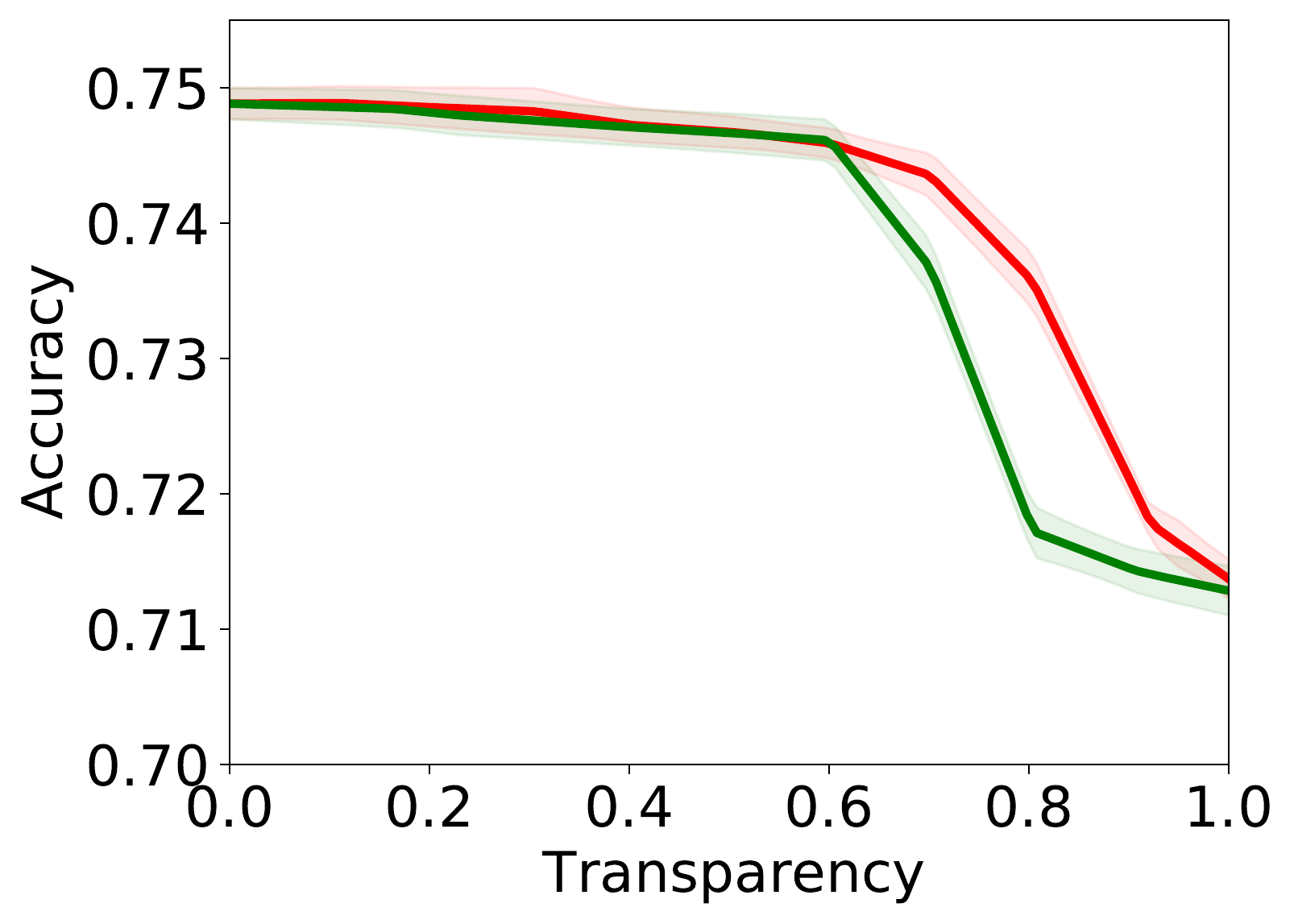}
        \caption{ACS Employment dataset.}
    \end{subfigure}

    \vskip 10 pt
    
    \begin{subfigure}{\textwidth}
        \centering
        \label{fig:results_tradeoffs_adult_test_pre_no_collab}
          \includegraphics[width=\figwidth\textwidth]
         {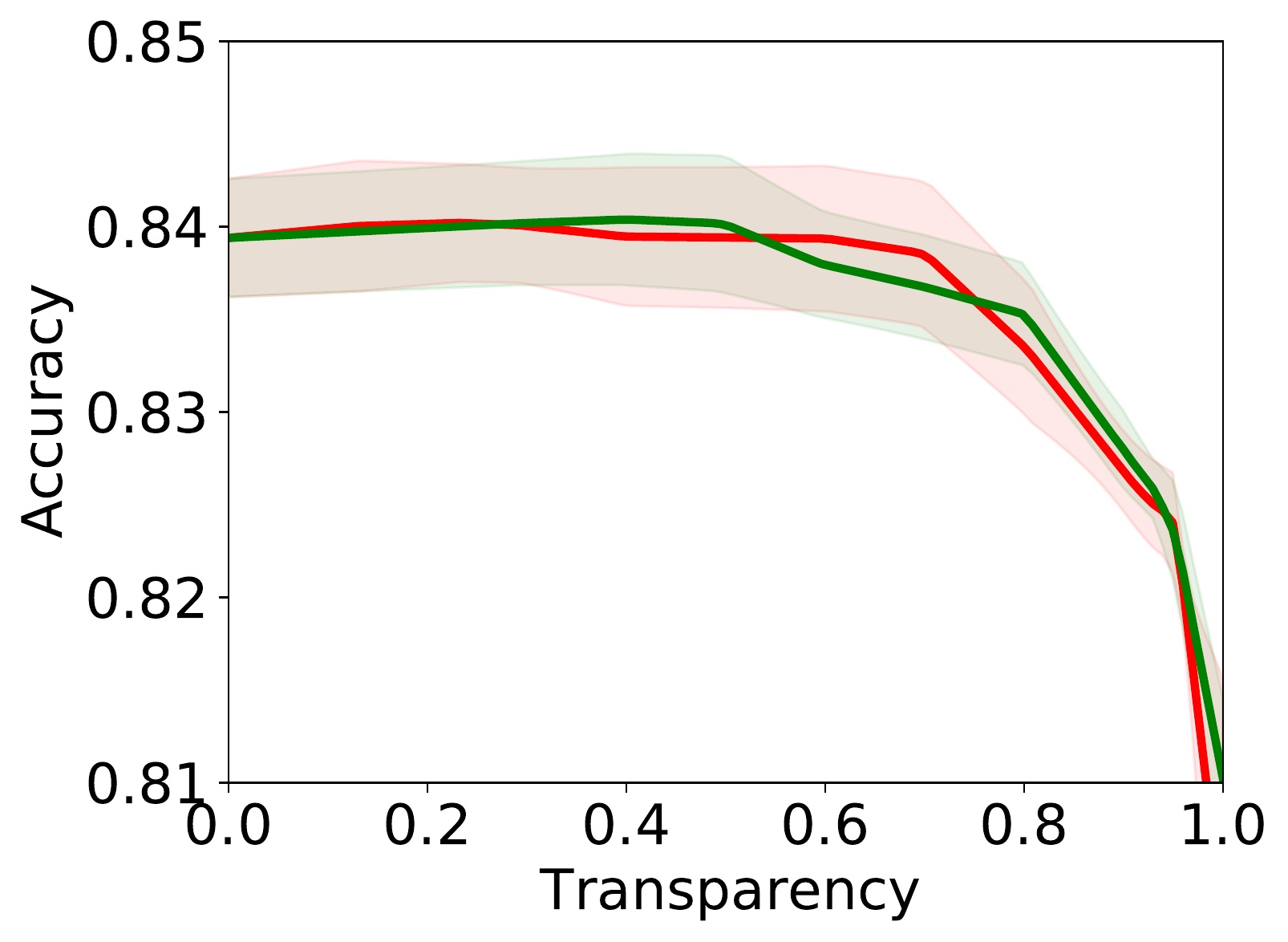}
         \includegraphics[width=\figwidth\textwidth]
         {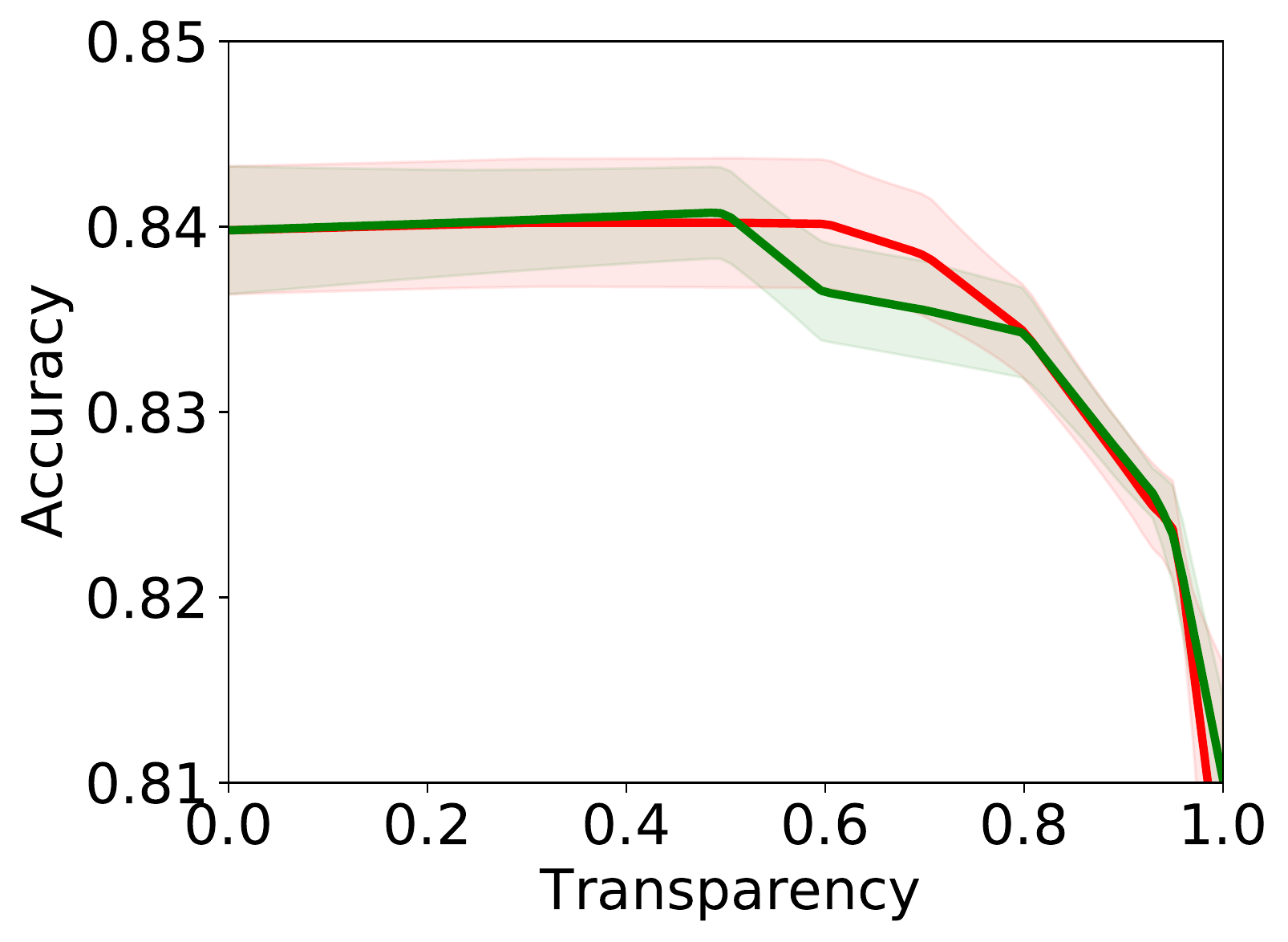}
        \includegraphics[width=\figwidth\textwidth]
         {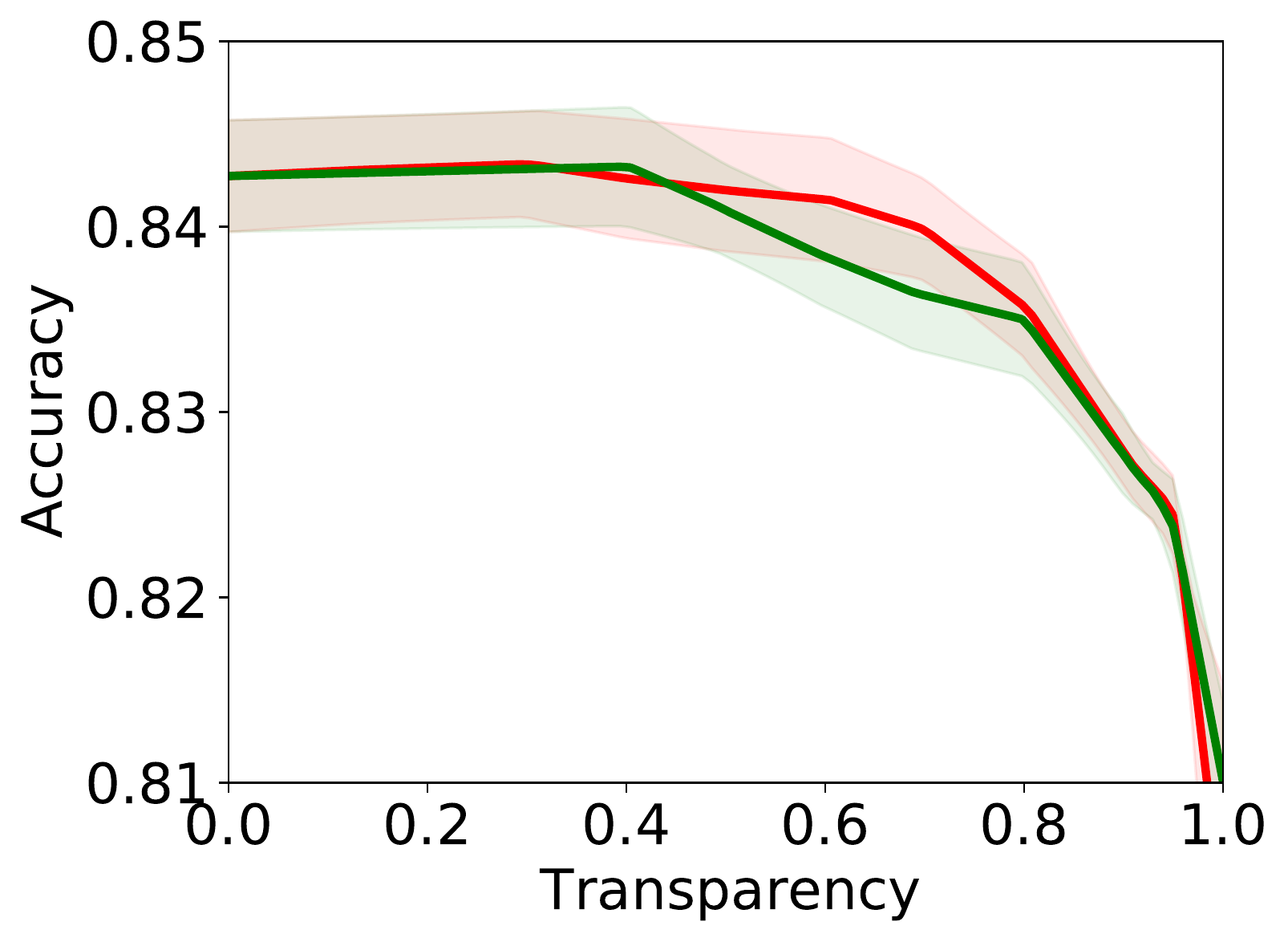}
        \caption{UCI Adult Income dataset.}
    \end{subfigure}
    
    \vskip 10 pt

    \begin{subfigure}{\textwidth}
        \centering
        \label{fig:results_tradoffs_compas_test_pre_no_collab}
        \includegraphics[width=\figwidth\textwidth]
         {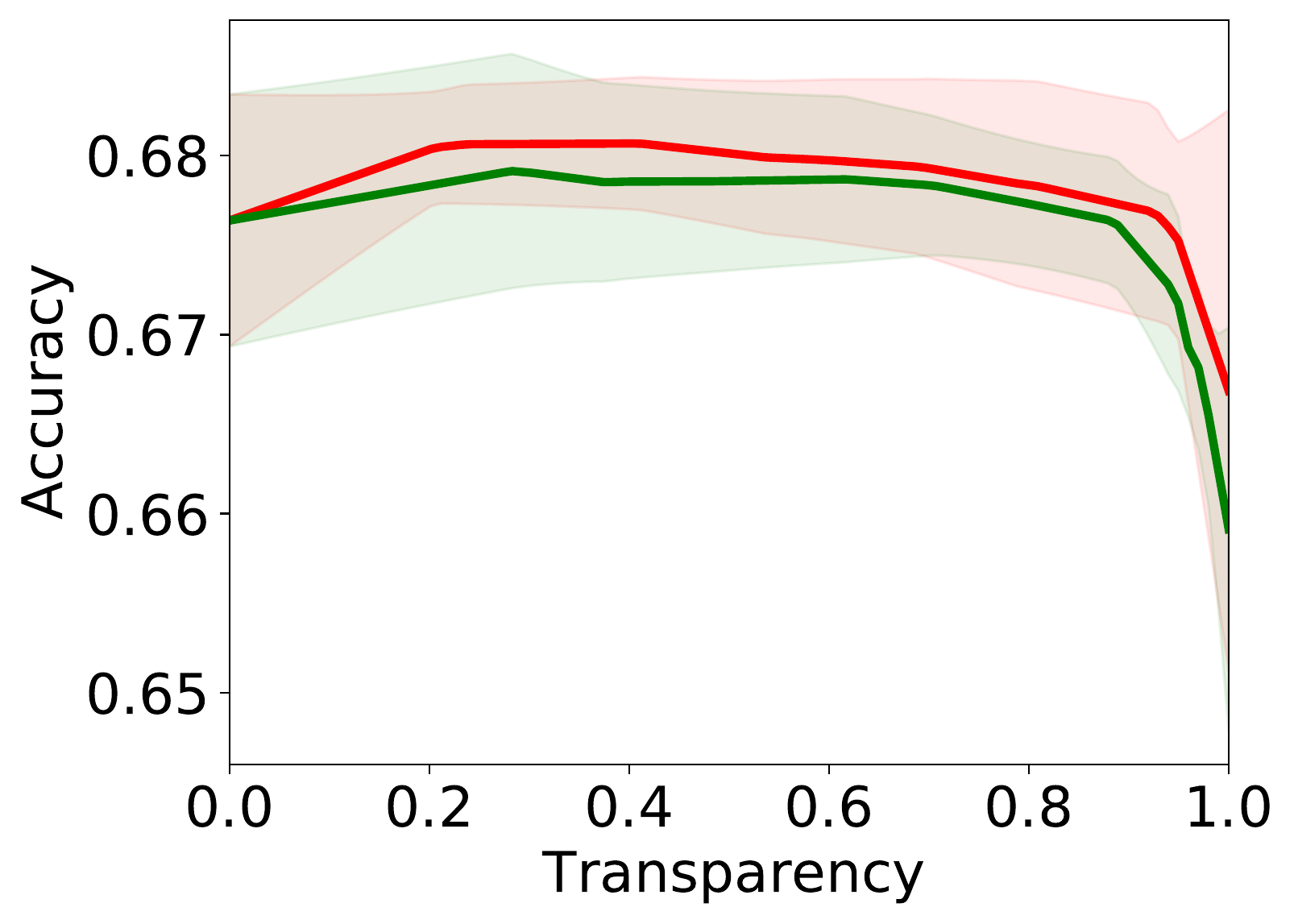}
         \includegraphics[width=\figwidth\textwidth]
         {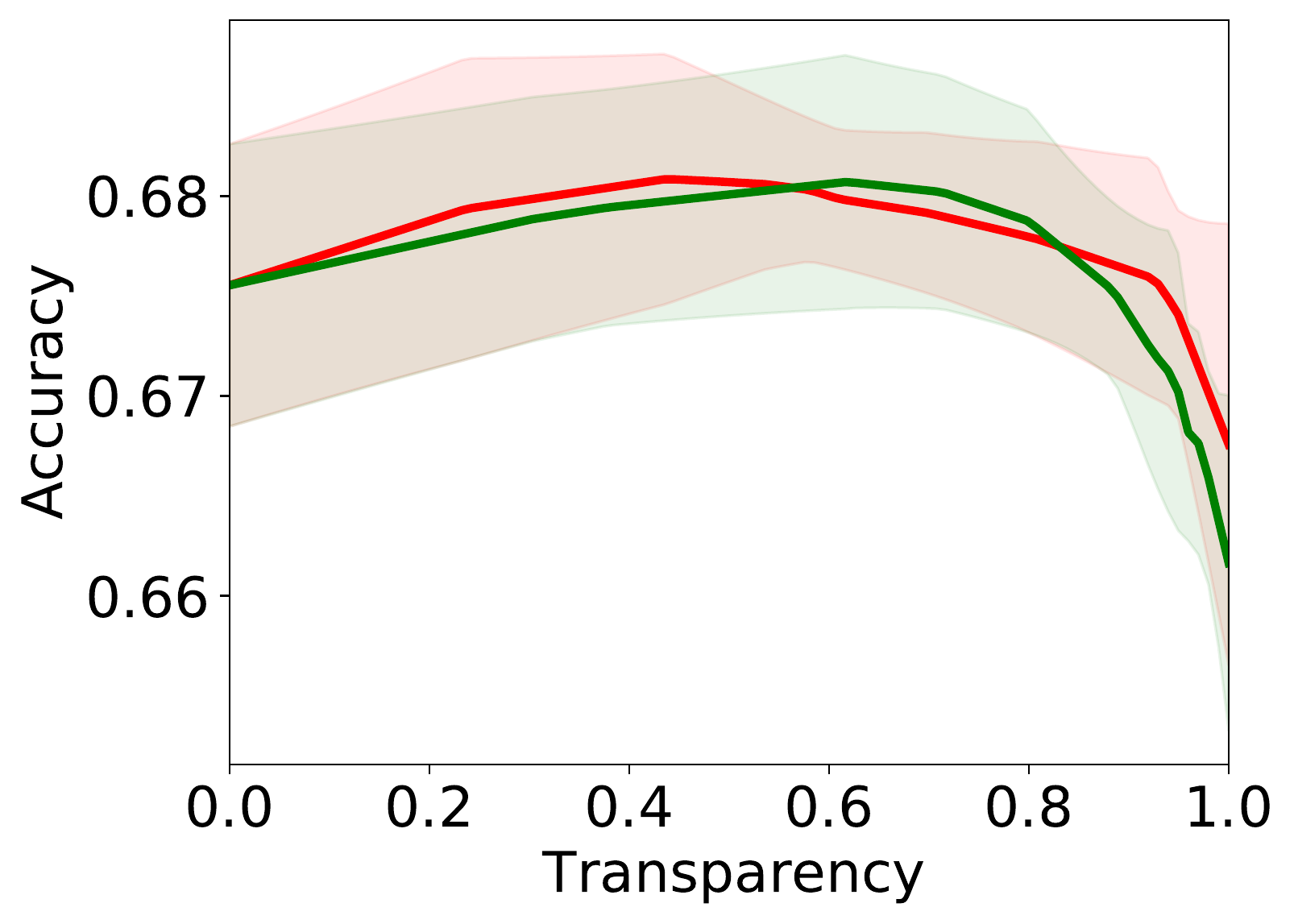}
        \includegraphics[width=\figwidth\textwidth]
         {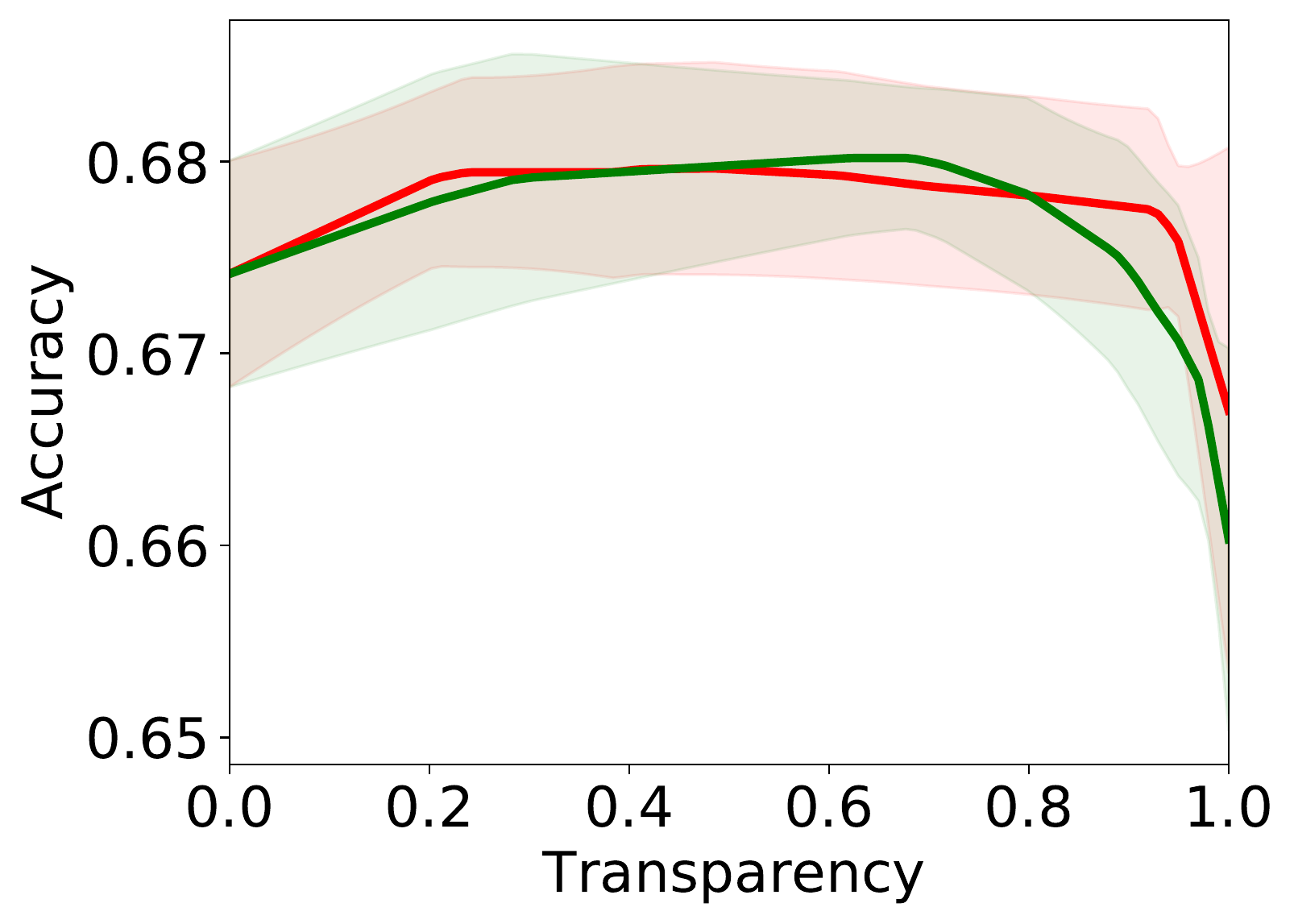}
        \caption{COMPAS dataset.}
    \end{subfigure}
    
     \vskip 10 pt
   
   \includegraphics[width=0.6\textwidth]
   {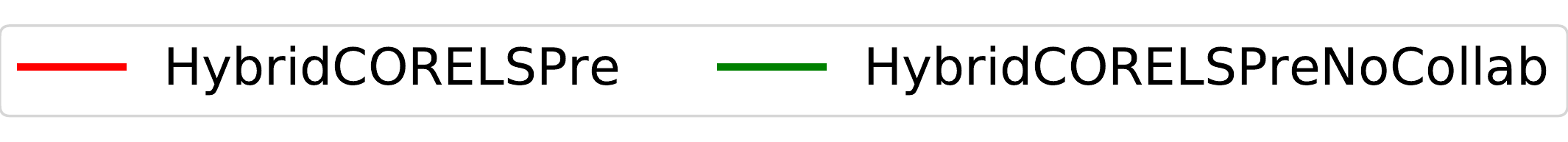}
   
    \caption{Test set accuracy/transparency trade-offs for our two \prebb{} variants of \HybridCORELS{}. 
    The Pareto front for each method is represented as a line and the filled bands encode the std across the five data split reruns.
    Results are provided for several black-boxes: (Left) AdaBoost, (Middle) Random Forests, (Right) Gradient Boosted Trees.}
    \label{fig:results_tradeoff_test_pre_no_collab}
    
    \end{center}
\end{figure*}
\begin{figure}[t!]
    \centering
     \begin{subfigure}[b]{0.9\textwidth}
\begin{lstlisting}[language=RuleListsLanguage,backgroundcolor=\color{RuleListsLanguageBackgroundColor}, basicstyle=\scriptsize]
if ["age_high" and "Female"] then 0 (acc 71.3%)
else if ["Husband/wife" and "No disability"] then 1 (acc 71.9%)
else if ["age_high" and "Native"] then 0 (acc 64.5%)
else if ["Bachelor's degree" and "No disability"] then 1 (acc 84.8%)
else if ["Reference person" and "No disability"] then 1 (acc 82.3%)
else if ["high school diploma" and "No disability"] then (acc 60.0%)
else
    AdaBoost() (acc 71.9%)
\end{lstlisting}
        \caption{\HybridCORELSPre{}: Test Accuracy 73.7\%, Transparency 80.2\%.} 
        \vspace{7pt}
    \end{subfigure}
    \vspace{7pt}
    \begin{subfigure}[b]{0.9\textwidth}
\begin{lstlisting}[language=RuleListsLanguage,backgroundcolor=\color{RuleListsLanguageBackgroundColor}, basicstyle=\scriptsize]
if ["age_low" and "Reference person"] then 1 (acc 82.2%)
else if ["Disability"] then 0 (acc 77.7%)
else if ["age_medium"] then 1 (acc 79.2%)
else if ["age_low" and "Married"] then 1 (acc 69.1%)
else if ["Husband/wife" and "Female"] then 0 (acc 68.6%)
else if ["age_low" and "not own child of householder"] then (acc 59.0%)
else
    AdaBoost() (acc 60.4%)
\end{lstlisting}
        \caption{\HybridCORELSPreNoCollab{} : Test Accuracy 71.7\%, Transparency 80.3\%.}
    \end{subfigure}
    \caption{Examples of hybrid interpretable models obtained on
    the ACS Employment dataset with AdaBoost black-boxes and the same train/validation/test split. The models with transparency closest to 
    80\% were selected. We note that the black-box has worst performance in \HybridCORELSPreNoCollab{} than \HybridCORELSPre{}
    seeing as the prefix sent it the inconsistent examples.}
    \label{fig:acs_adaboost_compare_pre_fold_0}
\end{figure}

\end{document}